\documentclass[10pt]{article}
\usepackage[margin=1in]{geometry}

\usepackage[affil-it]{authblk} 
\usepackage{etoolbox}
\usepackage{lmodern}
\usepackage{rotating}
\makeatletter
\patchcmd{\@maketitle}{\LARGE \@title}{\fontsize{14}{17.2}\selectfont\@title}{}{}
\makeatother

\usepackage{sectsty}

\sectionfont{\fontsize{13}{15}\selectfont}
\subsectionfont{\fontsize{12}{15}\selectfont}

\usepackage[font=footnotesize]{caption}
\usepackage{authblk}
\usepackage{microtype}
\usepackage{graphicx}
\usepackage{subfigure}
\usepackage{booktabs} 
\usepackage{amsmath,amssymb, amsthm, bm}
\usepackage{color,soul}

\usepackage{setspace}
\usepackage{gensymb}
\usepackage{bbm}
\usepackage[colorlinks=true,allcolors=blue]{hyperref}
\usepackage{natbib}
\usepackage{url}
\usepackage{wrapfig}
\usepackage{comment}
\usepackage{amsmath,amsfonts,amssymb}

\usepackage{mathtools}

\usepackage[ruled,vlined]{algorithm2e}
\usepackage{amsmath}
\usepackage[makeroom]{cancel}

\DeclarePairedDelimiterX{\infdivx}[2]{(}{)}{%
  #1\;\delimsize\|\;#2%
}

\newcommand{\norm}[1]{\left\lVert#1\right\rVert}

\newtheorem{theorem}{Theorem}
\newtheorem{assumption}{Assumption}

\newtheorem{subassumption}{Assumption\normalfont}[assumption]
\newenvironment{assumption*}
 {\ifnum\value{subassumption}=0 \stepcounter{assumption}\fi\subassumption}
 {\endsubassumption}
\newenvironment{assumption+}[1]
 {\renewcommand{\thesubassumption}{#1}\subassumption}
 {\endsubassumption}

\newtheorem{lemma}{Lemma}

\newtheorem{remark}{Remark}
\usepackage{colortbl}

\usepackage{amsmath}
\DeclareMathOperator{\Tr}{Tr}
\DeclareMathOperator*{\argmin}{arg\,min}

\DeclareMathOperator{\btheta}{\bm{\theta}}
\DeclareMathOperator{\bbtheta}{\bm{\bar{\theta}}}
\DeclareMathOperator{\Ker}{\text{Ker}}

\usepackage{fontawesome5}
\usepackage{hyperref}
\makeatletter
\newcommand{\github}[1]{%
   \href{#1}{\faGithubSquare}%
}
\makeatother

\title{\textbf{Federated Gaussian Process: \\Convergence, Automatic Personalization and Multi-fidelity Modeling}}
\date{\vspace{-7ex}}

\author[1]{Xubo Yue, Raed Al Kontar\\Industrial and Operations Engineering\\University of Michigan, Ann Arbor}

\begin{document}
\maketitle

\begin{abstract}
In this paper, we propose \texttt{FGPR}: a Federated Gaussian process ($\mathcal{GP}$) regression framework that uses an averaging strategy for model aggregation and stochastic gradient descent for local client computations. Notably, the resulting global model excels in personalization as \texttt{FGPR} jointly learns a global $\mathcal{GP}$ prior across all clients. The predictive posterior then is obtained by exploiting this prior and conditioning on local data which encodes personalized features from a specific client. Theoretically, we show that \texttt{FGPR} converges to a critical point of the full log-likelihood function, subject to statistical error. Through extensive case studies we show that \texttt{FGPR} excels in a wide range of applications and is a promising approach for privacy-preserving multi-fidelity data modeling.
\end{abstract}

\section{Introduction}
\label{sec:intro}
The modern era of computing is gradually shifting from a centralized regime where data is stored in a centralized location, often a cloud or central server, to a decentralized paradigm that allows clients to collaboratively learn models while keeping their data stored locally \citep{kontar2021internet}. This paradigm shift was set forth by the massive increase in compute resources at the edge device and is based on one simple idea: instead of learning models on a central server, edge devices execute small computations locally and only share the minimum information needed to learn a model. This modern paradigm is often coined as federated learning (FL). Though the prototypical idea of FL dates back decades ago, to the early work of \citet{mangasarian1994backpropagation}, it was only brought to the forefront of deep learning after the seminal paper by \citet{mcmahan2017communication}. In their work,  \citet{mcmahan2017communication} propose Federated Averaging (\texttt{FedAvg}) for decentralized learning of a deep learning model. In \texttt{FedAvg}, a central server broadcasts the network architecture and a global model (e.g., initial weights) to select clients, clients perform local computations (using stochastic gradient descent - SGD) to update the global model based on their local data and the central server then takes an average of the resulting local models to update the global model. This process is iterated until an accuracy criterion is met.   

Despite the simplicity of taking averages of local estimators in deep learning, \texttt{FedAvg} \citep{mcmahan2017communication} has seen immense success and has since generated an explosive interest in FL. To date, \texttt{FedAvg} for decentralized learning of deep neural networks (NN) was tailored to image classification, text prediction, wireless network analysis and condition monitoring \& failure detection \citep{smith2018cocoa, brisimi2018federated, tran2019federated, kim2019blockchained,yue2019renyi, li2020federated}. Besides that, building upon \texttt{FedAvg}'s success, literature has been proposed to: (i) tackle adversarial attacks in FL \citep{bhagoji2019analyzing, wang2020attack}; (ii) allow personalization whereby each device retains its own individualized model \citep{li2021ditto}; (iii) ensure fairness in performance and participation across clients \citep{li2019fair, mohri2019agnostic, yue2021gifair, zeng2021improving}; (iv) develop more complex aggregation strategies that accommodate deep convolution network \citep{wang2020federated}; (v) accelerate FL algorithms to improve convergence rate or reduce communication cost \citep{karimireddy2020scaffold, yuan2020federated}; (vi) improve generalization through model ensembling \citep{shi2021fed}.

Despite the aforementioned ubiquitous application of FL, most if not all, FL literature lies within an empirical risk minimization (ERM) framework - a direct consequence of the focus on deep learning.  To date, very few papers study FL beyond ERM, specifically when correlation exists. In this paper we go beyond ERM and focus on the Gaussian process ($\mathcal{GP}$). We investigate both theoretically and empirically the (i) plausibility of federating model/parameter estimation in $\mathcal{GP}$s and (ii) applications where federated $\mathcal{GP}$s can be of immense value. Needless to say, the inherent capability to encode correlation, quantify uncertainty and incorporate highly flexible model priors has rendered $\mathcal{GP}$s a key inference tool in various domains such as Bayesian calibration \citep{plumlee2014building, plumlee2017bayesian}, multi-fidelity modeling, experimental design \citep{rana2017high, yue2020non, jiang2020binoculars, krishna2021robust}, manufacturing \citep{tapia2016prediction, peng2017bayesian}, healthcare \citep{imani2018nested, ketu2021enhanced}, autonomous vehicles \citep{goli2018vehicle} and robotics \citep{deisenroth2013gaussian, jang2020multi}. Therefore, the success of FL within $\mathcal{GP}$s may help pave the way for FL to infiltrate many new applications and domains.

The central challenge is that, unlike empirical risk minimization, $\mathcal{GP}$s feature correlations across all data points such that any finite collection of which has a joint Gaussian distribution \citep{sacks1989design, currin1991bayesian}. As a result, the empirical risk does not simply sum over the loss of individual data points. Adding to that, mini-batch gradients become biased estimators when correlation exists. The performance of FL in such a setting is yet to be understood and explored. 

To this end, we propose \texttt{FGPR}: a \textbf{F}ederated $\bm{\mathcal{GP}}$ \textbf{R}egression framework that uses \texttt{FedAvg} (i.e. averaging strategy) for model aggregation and SGD for local clients computations. First, we show that, under some conditions, \texttt{FGPR} converges to a critical point of the full log-likelihood loss function and recovers true parameters (or minimizes averaged loss) up to statistical errors that depend on the client's mini-batch size. Our results hold for kernel functions that exhibit exponential or polynomial eigendecay which is satisfied by a wide range of kernels commonly used in $\mathcal{GP}$s such as the Mat\'{e}rn and radial basis function (RBF) kernel. Our proof offers standalone value as it is the first to extend theoretical results of FL beyond ERM and to a correlated paradigm. In turn, this may help researchers further investigate FL within alternative stochastic processes built upon correlations, such as L\'evy processes. Second, we explore \texttt{FGPR} within various applications to validate our results. Most notably, we propose \texttt{FGPR} as a privacy-preserving approach for multi-fidelity data modeling and show its advantageous properties compared to the state-of-the-art. In addition, we find an interesting yet unsurprising observation. The global model in \texttt{FGPR} excels in personalization. This feature is due to the fact that ultimately \texttt{FGPR} learns a global prior across all clients. The predictive posterior then is obtained by exploiting this prior and conditioning on local data which encodes personalized features from a specific client. This notion of automatic personalization is closely related to meta-learning where the goal is to learn a model that can achieve fast personalization. 

\subsection{Summary of Contributions \& Findings} \label{sec:contributions} 

We briefly summarize our contributions below:

\begin{itemize}
    \item \textbf{Convergence:} We explore two data-generating scenarios. (1) Homogeneous setting where local data is generated from the same underlying distribution or stochastic process across all devices; (2) Heterogeneous setting where devices have distributional differences. Under both scenarios and for a large enough batch size $M$, we prove that \texttt{FGPR} converges to a critical point of the full-likelihood loss function (from all data) for kernels that exhibit an exponential or polynomial eigendecay. We also provide uniform error bounds on parameter estimation errors and highlight the ability of \texttt{FGPR} to recover the underlying noise variance.
    \begin{itemize}
    \item Interestingly, our derived bounds not only depend on iteration $T$, but also explicitly depend on batch size $M$ which is a direct consequence of correlation. %shedding additional light on the sample complexity. 
 Our results do not assume any specific functional structure such as convexity, Lipschitz continuity or bounded variance.
    \end{itemize}
    \item \textbf{Automatic Personalization Capability:} We demonstrate that \texttt{FGPR} can automatically personalize the shared global model to each local device. Learning a global model by \texttt{FGPR} can be viewed as jointly learning a global $\mathcal{GP}$ prior. On the other hand, the posterior predictive distribution of a $\mathcal{GP}$ depends both on this shared prior and the local training data. The latter one can be viewed as a personalized feature encoded in the $\mathcal{GP}$ model.  This important personalization feature allows \texttt{FGPR} to excel in the scenario where data among each local device is heterogeneous (Sec. \ref{sec:exp_fidelity} and Sec. \ref{sec:exp_robot}).
    \begin{itemize}
    \item In addition to the personalization capability, we find that the prior class learned from \texttt{FGPR} excels in transfer learning to new devices (Sec. \ref{sec:exp_nasa}). This idea is similar to meta-learning, where one tries to learn a global model that can quickly adapt to a new task.
    %and helps local devices with bad parameter initialization fit an accurate surrogate model (Sec. \ref{sec:exp_sim})
    \end{itemize}
    \item \textbf{Multi-fidelity modeling and other applications:} We propose \texttt{FGPR} as a privacy-preserving approach for multi-fidelity data modeling which combines datasets of varying fidelities into one unified model. We find that in such settings, not only does \texttt{FGPR} preserve privacy but also can improve generalization power across various existing state-of-the-art multi-fidelity approaches. We also validate \texttt{FGPR} on various simulated datasets and real-world datasets to highlight its advantageous properties.
\end{itemize}

The remainder of this paper is organized as follows. In Sec. \ref{sec:algorithm}, we present the \texttt{FGPR} algorithm. We study the theoretical properties of \texttt{FGPR} in Sec. \ref{sec:theory}. In Sec. \ref{sec:exp_sim}-\ref{sec:exp_robot}, we present several empirical results over a range of simulated datasets and real-world datasets. A detailed literature review can be found in Sec. \ref{sec:related}. We conclude our paper in Sec. \ref{sec:con} with a brief discussion. Codes are available on this GitHub link  \github{https://github.com/UMDataScienceLab/Federated_Gaussian_Process}.

\section{The \texttt{FGPR} Algorithm} 
\label{sec:algorithm}

In this section, we describe the problem setting in Sec. \ref{subsec:set} and  introduce \texttt{FGPR} - a federated learning scheme for $\mathcal{GP}$s in Sec. \ref{subsec:train}. We then provide insights on the advantages of \texttt{FGPR} in Sec. \ref{subsec:why}. Specifically, we will show that \texttt{FGPR} is capable of automatically personalizing the global model to each local device. This property allows \texttt{FGPR} to excel in many real-world applications, such as multi-fidelity modeling, where heterogeneity exists.

\subsection{Background}
\label{subsec:set}

We consider the Gaussian process regression. We first briefly review the centralized $\mathcal{GP}$ model. Suppose the training dataset is given as  $D=\left\{\bm{X}, \bm{y} \right\}$, where $\bm{y}=\left[y_{1}, ..., y_{N}\right]^\intercal$, $\bm{X}=\left[x_{1}^\intercal, ..., x_{N}^\intercal\right]$ and $N$ denotes the number of observations. In this paper, we use $|D|=N$ to denote the cardinality of set $D$. Here, $x\in\mathbb{R}^d$ is a $d$-dimensional input and $y\in\mathbb{R}$ is the output. We decompose the output as $y(x)=f(x)+\epsilon(x)$, where
\begin{align*}
    f\sim\mathcal{GP}(0,\mathcal{K}(\cdot,\cdot;\bm{\theta}_{\mathcal{K}})) , \quad \epsilon\overset{\text{i.i.d.}}{\sim} \mathcal{N}(0, \sigma^2),
\end{align*}
and $\mathcal{K}(\cdot,\cdot;\bm{\theta}_{\mathcal{K}})$ is the prior covariance function parameterized by kernel parameters $\bm{\theta}_{\mathcal{K}}$. Given a new observation $x^*$, the goal of $\mathcal{GP}$ regression is to predict $f(x^*)$. By definition, any finite collection of observations from a $\mathcal{GP}$ follows a multivariate normal distribution. Therefore, the joint distribution of $\bm{y}$ and $f(x^*)$ is given as
\begin{align*}
    \begin{bmatrix}\bm{y}\\f(x^*)\end{bmatrix}\sim\mathcal{N}\left(\bm{0},\begin{bmatrix}\bm{K}(\bm{X},\bm{X})+\sigma^2\bm{I}\ & \bm{K}(\bm{X},x^*) \\\bm{K}(x^*,\bm{X})\ & \bm{K}(x^*,x^*)\end{bmatrix}\right)
\end{align*}
where $\bm{K}(\cdot,\cdot):\mathbb{R}^d\times\mathbb{R}^d\to\mathbb{R}$ is a covariance matrix whose entries are determined by the covariance function $\mathcal{K}(\cdot,\cdot;\bm{\theta}_{\mathcal{K}})$. Therefore, the conditional distribution (also known as the posterior predictive distribution) of $f(x^*)$ is given as $\mathcal{N}\left(\mu_{pred}(x^*),\sigma^2_{pred}(x^*)\right)$, where
\begin{equation}
\begin{split}
\label{eq:pred}
    &\mu_{pred}(x^*)=\bm{K}(x^*, \bm{X})\left(\bm{K}(\bm{X}, \bm{X})+\sigma^2\bm{I}\right)^{-1}\bm{y},\\
    &\sigma^2_{pred}(x^*)\\
    &=\bm{K}(x^*, x^*)-\bm{K}(x^*, \bm{X})\left(\bm{K}(\bm{X},\bm{X})+\sigma^2\bm{I}\right)^{-1}\bm{K}(\bm{X},x^*).
\end{split}
\end{equation}
Here, $\mu_{pred}(x^*)$ is a point estimate of $f(x^*)$ and $\sigma^2_{pred}(x^*)$ quantifies the variance. It can be seen that the covariance matrix $\bm{K}(\cdot,\cdot)$ and noise parameter $\sigma^2$ are the most important components in the $\mathcal{GP}$ prediction. The former one is determined by the covariance function $\mathcal{K}(\cdot,\cdot;\bm{\theta}_{\mathcal{K}})$, which is typically assumed to belong to a kernel family parameterized by $\bm{\theta}_{\mathcal{K}}$. In this paper, we denote by $\btheta\coloneqq(\bm{\theta}_{\mathcal{K}},\sigma^2)$ the $\mathcal{GP}$ model parameters. Therefore, predicting an accurate output $f(x^*)$ critically depends on finding a good estimate of $\btheta$. To estimate $\btheta$, the most popular approach is to minimize the log marginal likelihood in the form of
\begin{align}
\label{Eq:likelihood}
    &-\log p(\bm{y}|\bm{X};\btheta)=-\log\int p(\bm{y}|\bm{X},f;\btheta)p(f|\bm{X};\btheta)df\nonumber\\
    &=\frac{1}{2}\left[\bm{y}^\intercal\bm{K}^{-1}(\bm{X},\bm{X})\bm{y} + \log\left|\bm{K}(\bm{X},\bm{X})\right|+N\log(2\pi)\right]
\end{align}
There are numerous optimizers that are readily available to minimize $-\log p(\bm{y}|\bm{X};\btheta)$. In this paper, we resort to stochastic optimization methods such as SGD or Adam \citep{kingma2014adam}. % to understand the underlying theoretical properties of \texttt{FGPR}. 

\begin{remark}
Needless to say, a current critical challenge in FL, is that edge devices have limited compute power. SGD offers an excellent scalability solution to the computational complexity of $\mathcal{GP}s$ which has been a long-standing bottleneck since $\mathcal{GP}s$  require inverting a covariance matrix $\bm{K}(\cdot,\cdot)$ at each iteration of an optimization procedure (see Eq. (\ref{Eq:likelihood})). This operation, in general, incurs a $\mathcal{O}(N^3)$ time complexity. In SGD, only a mini-batch with a size of $M\ll N$ is taken at each iteration; hence allowing $\mathcal{GP}$s to scale to big data regimes. Besides that, and as will become clear shortly, our approach only requires edge devices to do few steps on SGD on their local data. Another notable advantage of SGD is that it offers good generalization power \citep{keskar2016large}. In deep learning, it is well-known that SGD can drive solutions to a flat minimizer that generalizes well \citep{wu2018sgd}. Although this statement is still an open problem in $\mathcal{GP}$, \citet{chen2020stochastic} empirically validate that the solution obtained by SGD generalizes better than other deterministic optimizers.
\end{remark}

In a centralized regime, and given a mini-batch of size $M$, the stochastic gradient (SG) of  the log-likelihood in Eq. \eqref{Eq:likelihood} is
\begin{align*}
    &g(\bm{\theta};\xi) \\   &=\frac{1}{2}\bigg[-\bm{y}_{\xi}^\intercal\bm{K}^{-1}(\bm{X}_{\xi},\bm{X}_{\xi})\frac{\partial \bm{K}(\bm{X}_{\xi},\bm{X}_{\xi})}{\partial\theta}\bm{K}^{-1}(\bm{X}_{\xi},\bm{X}_{\xi})\bm{y}_{\xi} +\text{Tr}\left(\bm{K}^{-1}(\bm{X}_{\xi},\bm{X}_{\xi})\frac{\partial \bm{K}(\bm{X}_{\xi},\bm{X}_{\xi})}{\partial\theta}\right)\bigg] \\
    &= \frac{\Tr\bigg[\bm{K}^{-1}(\bm{X}_{\xi},\bm{X}_{\xi})\left(\bm{I}-\bm{y}_{\xi}\bm{y}_{\xi}^T\bm{K}^{-1}(\bm{X}_{\xi},\bm{X}_{\xi})\right)\frac{\partial \bm{K}(\bm{X}_{\xi},\bm{X}_{\xi})}{\partial\theta} \bigg]}{2}
\end{align*}
where $\xi$ is the set of indices corresponding to a subset of training data with mini-batch size $M$ and $\bm{X}_{\xi}, \bm{y}_{\xi}$ is the respective subset of inputs and outputs indexed by $\xi$. At each iteration $t$, a subset of training data is taken to update model parameters  as
\begin{align*}
    \bm{\theta}^{(t+1)} \leftarrow \bm{\theta}^{(t)} - \eta^{(t)}g(\bm{\theta}^{(t)};\xi^{(t)}),
\end{align*}
where $\eta^{(t)}$ is the learning rate at iteration $t$. This step is repeated several times till some exit condition is met. 

\begin{remark}
Although SGD was a key propeller for deep learning, it has only recently been shown to be applicable to $\mathcal{GP}$ inference \citep{chen2020stochastic}. The key difficulty is that unlike ERM, the log-likelihood cannot be written as a sum over individual data-points due to the existence of correlation. In turn, this renders the stochastic gradient of a mini-batch ($g(\bm{\theta};\xi)$) a biased estimator of the full gradient when taking expectation with respect to the random sampling.
\end{remark}

In the next section, we extend this $\mathcal{GP}$ inference framework to a federated setting.

\subsection{The \texttt{FGPR} Framework}
\label{subsec:train}

Suppose there exists $K\geq 2$ local devices. In this paper, we will use (edge) devices and clients interchangeably. For client $k \in [K]$, the local dataset is given as $D_k=\left\{\bm{X}_k, \bm{y}_k \right\}$ with cardinality $N_k$. We let $N=\sum_{k=1}^{K}N_k$. Denote by $L_k(\btheta;D_k)\coloneqq-\log p(\bm{y}_k|\bm{X}_k;\btheta)$ the loss function for device $k$ and $g_k(\bm{\theta};\xi_k)$ the SG of this loss function with respect to a mini-batch of size $M$ indexed by $\xi$.

\textbf{In FL, our goal is to collaboratively learn a global parameter} $\btheta$ \textbf{that minimizes the global loss function in the form of}
\begin{align}
    L(\btheta)\coloneqq\sum_{k=1}^Kp_k L_k(\btheta;D_k)
\end{align}
where $p_k=\frac{N_k}{\sum_{k=1}^KN_k}$ is the weight parameter for device $k$ such that $\sum_{k=1}^Kp_k=1$.  To fulfill this goal, during each communication period, each local device $k$ runs $E$ steps of SGD and updates model parameters as 
\begin{align*}
    \bm{\theta}^{(t+1)}_k \leftarrow \bm{\theta}^{(t)}_k - \eta^{(t)}g_k(\bm{\theta}^{(t)}_k;\xi^{(t)}_k).
\end{align*}
At the end of each communication round, the central server aggregates model parameters as
\begin{align*}
    \bm{\bar{\theta}} = \sum_{k=1}^Kp_k\bm{\theta}_k.
\end{align*}
The aggregated parameter $\bm{\bar{\theta}}$ is then distributed back to local devices. This cycle is repeated several times till convergence. In this training framework, all devices participate during each communication round. We define this framework as synchronous updating. In reality, however, some local devices are frequently offline or reluctant/slow to respond due to various unexpected reasons. To resolve this issue, we develop an asynchronous updating scheme. Specifically, at the beginning of each communication round ($c$), we select $K_{\text{c}}\in[1,K)$ clients by sampling probability $p_k$ and denote by ${\cal S}_c$ the indices of these clients. During the communication round, the central server aggregates model parameter as
\begin{align*}
    \bm{\bar{\theta}}=\frac{1}{K_{c}}\sum_{k\in\mathcal{S}_c}\bm{\theta}_k.
\end{align*}

The detailed procedure is given in Algorithm \ref{algo:global}.

\begin{remark}
The aggregation strategy used in Algorithm \ref{algo:global} is known as \texttt{FedAvg} \citep{mcmahan2017communication}. Despite being the first proposed aggregation scheme for FL, \texttt{FedAvg} has stood the test in the past couple of years as one of most robust and competitive approaches for model aggregation. That being said, it is also possible to extend our algorithm to different strategies such as different sampling or weighting schemes. 
\end{remark}

% We consider scaling factors $s_1(M)=\tau\log M$ and $s_2(M)=M$, where $\tau$ is a constant scalar. The scaling factors simply ensure components in the gradient have the same scale. 

% \begin{align}
% \label{eq:GP}
%     -\frac{1}{N_k}\log p(\bm{y}_k|\bm{X}_k)=\frac{1}{2N_k}\left[\bm{y}_k^T\bm{K}_{N_k}^{-1}(\btheta_k)\bm{y}_k + \log|\bm{K}_{N_k}(\btheta_k)|+N_k\log(2\pi)\right],
% \end{align}
% where $\btheta_k=(\theta_{1k},\theta_{2k})$ is the model parameter, $\bm{K}_{N_k}(\btheta_k)=\theta_{1k}\bm{K}_{f_k,N_k}+\theta_{2k}\bm{I}_{N_k}$ is the covariance matrix of size $N_k\times N_k$, $K_{f_k,N_k}$ is the kernel matrix, $\bm{I}_{N_k}$ is the identity matrix, and
% $\theta_{1k} (\theta_{2k})$ is the variance (noise) parameter for device $k$. For each $k$, we calculate the stochastic gradient as follow:
% \begin{align*}
%     \big[g_k(\bm{\theta}_k;\xi_k)\big]_l = \frac{1}{2s_l(M_k)}\Tr\bigg[\bm{K}_{\xi_k}^{-1}(\btheta_k)(\bm{I}_m-\bm{y}_{\xi_k}\bm{y}_{\xi_k}^T\bm{K}_{\xi_k}^{-1}(\btheta_k))\frac{\partial \bm{K}_{\xi_k}(\btheta_k)}{\partial\theta_l} \bigg], \ l = 1,2
% \end{align*}
%We consider scaling factors $s_1(M)=\tau\log M$ and $s_2(M)=M$, where $\tau$ is a constant scalar. The scaling factors simply ensure components in the gradient have the same scale. 

\begin{algorithm}[!htbp]
	\SetAlgoLined
	\KwData{number of sampled devices $K_{\text{sample}}$, number of communication rounds $R$, initial model parameter $\bm{\theta}$}
	\For{$c=0:(R-1)$}{
	    Select $K_{\text{sample}}$ clients by sampling probability $p_k$ and denote by ${\cal S}_c$ the indices of these clients\;
	    Server broadcasts $\bm{\theta}$\;
	    \For{$k\in\mathcal{S}_c$}{
	        $\bm{\theta}_k^{(0)}=\bm{\theta}$\;
    		Update model parameter (e.g., using Algorithm \ref{algo:local})\;
		}
		Aggregation $\bm{\bar{\theta}}_c=\frac{1}{K_{\text{sample}}}\sum_{k\in\mathcal{S}_c}\bm{\theta}^{(E)}_k$, Set $\bm{\theta}=\bm{\bar{\theta}}_c$\;
	}
	Return $\bm{\bar{\theta}}_R$.
	\caption{The \texttt{FGPR} algorithm}
	\label{algo:global}
\end{algorithm}

    \vspace{-0.7cm}
\begin{algorithm}[!htbp]

	\SetAlgoLined
	\KwData{index of device $k$, number of local updates $E$, SGD learning rate schedule $\{\eta^{(t)}\}_{t=1}^E$, initial model parameter $\btheta_k^{(0)}$}
% 	\KwResult{model parameter $\bm{\bar{\theta}}_C$.}
	\For{$t=0:(E-1)$}{
	    Randomly sample a subset of data from $D_k$ and denote it as $\xi_k^{(t)}$\;
    	$\bm{\theta}^{(t+1)}_k=\bm{\theta}^{(t)}_k-\eta^{(t)}g_k(\bm{\theta}^{(t)}_k;\xi_k^{(t)})$ \;
	}
	Return $\btheta_k^{(E)}$\;
	\caption{Local update using SGD}
	\label{algo:local}
\end{algorithm}

\subsection{Why a Single Global $\mathcal{GP}$ Model Works?}
\label{subsec:why}
In this paper, we will demonstrate the viability of \texttt{FGPR} in cases where data across devices are both homogeneous or heterogeneous. In heterogeneous  settings, it is often the case that personalized FL approaches are developed where clients eventually retain their own models while borrowing strength from one another. Popular personalization methods usually fine-tune the global model based on local data while encouraging local weights to stay in a small region in the parameter space of the global model \citep{li2021ditto}. This allows a balance between client's shared knowledge and unique characteristic. This literature however is mainly focused on deep learning. 

One natural question is: why does a single global model learned from Algorithm \ref{algo:global} work in \texttt{FGPR}. Here it is critical to note that, unlike deep learning, estimating $\bm{\theta}$ in a $\mathcal{GP}$ is equivalent to learning a prior through which predictions are obtained by conditioning on the observed data and the learned prior. 

More specifically, in the $\mathcal{GP}$, we impose a prior on $f_k$ such that $f_k\sim\mathcal{GP}(0,\mathcal{K}(\cdot,\cdot;\bm{\theta}_{\mathcal{K}}))$. The covariance function is parameterized by $\bm{\theta}_{\mathcal{K}}$. Therefore, learning a global model by \texttt{FGPR} can be viewed as learning a common model prior over $f_k, \forall k$. On the other hand, the posterior predictive distribution at a testing point $x^*$ is given as
\begin{align*}
    &p(f_k^*|\bm{X}_k, \bm{y}_k, x^*)=\int p(f_k^*|x^*,f_k) p(f_k|\bm{X}_k, \bm{y}_k) df_k\\
    &=\int  p(f_k^*|x^*,f_k)\frac{p(\bm{y}_k|\bm{X}_k,f_k)\overbrace{p(f_k)}^{\text{prior}}}{p(\bm{y}_k|\bm{X}_k)}df_k\\ &=\mathcal{N}(\mu_{k,pred}(x^*),\sigma^2_{k,pred}(x^*)),
\end{align*}
where the predictive mean $\mu_{k,pred}(x^*)$ and the predictive variance $\sigma^2_{k,pred}(x^*)$ are defined in Eq. \eqref{eq:pred}.
% \begin{align}
% \label{eq:pred}
%     \mu_{k,pred}(x^*)&=\bm{K}(x^*, \bm{X}_k)(\bm{K}(\bm{X}_k, \bm{X}_k)+\sigma^2\bm{I})^{-1}\bm{y}_k,\\
%     \sigma^2_{k,pred}(x^*)&=\bm{K}(x^*, x^*)-\bm{K}(x^*, \bm{X}_k)(\bm{K}(\bm{X}_k,\bm{X}_k)+\sigma_2\bm{I})^{-1}\bm{K}(\bm{X}_k,x^*).
% \end{align}
From this posterior predictive equation, one can see that the predicted trajectory (and variance) of $\mathcal{GP}$ in device $k$ is affected by both prior distribution and training data $(\bm{X}_k,\bm{y}_k)$ explicitly. For a specific device, the local data themselves embody the personalization role. Therefore, \texttt{FGPR} can automatically tailor a shared global model to a personalized model for each local device. This idea is similar to meta-learning, where one tries to learn a global model that can quickly adapt to a new task.

% On the other hand, regression models such as a polynomial regression or a neural network typically make parametric assumptions on the functional form. For instance, a simple linear regression assumes $y_{k}(x)=\beta x+\epsilon_{k}(x)$. Thus, the predictive equation will depend on the solely depends on weight coefficient $\beta$. Therefore, one single global parameters in those models deteriorate when the data distributions are heterogeneous. 

\begin{figure}[!htbp]
    \centering
    \centerline{\includegraphics[width=0.8\columnwidth]{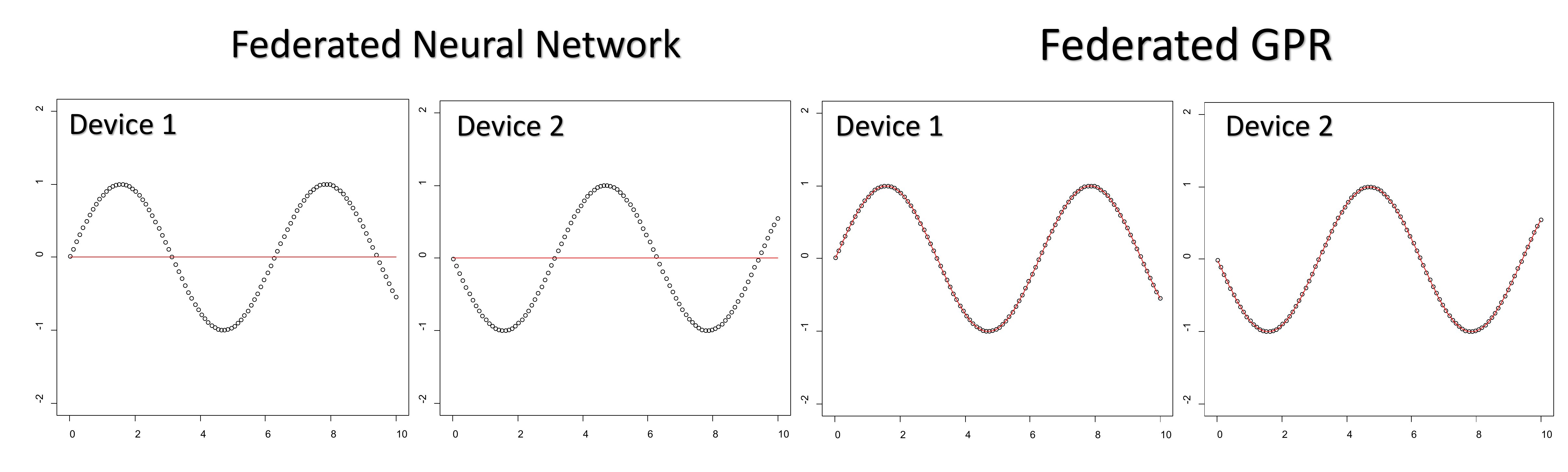}}
    \caption{A simple example that is used to demonstrate the automatic personalization feature of \texttt{FGPR}. In the plot, the black dots are original data and the red lines are fitted curves.}
    \label{fig:global_local}
\end{figure}

To see this, we create a simple and stylized numerical example. Suppose there are two local devices. Device 1 has data that follows $y=\sin(x)$ while device 2 has data that follows $y=-\sin(x)$. Each device has $100$ training points uniformly spread in $[0,10]$. We use \texttt{FedAvg} to train a 2-layer neural network. Unfortunately, a single global model of a neural network simply returns a line, as shown in Figure \ref{fig:global_local}. Mathematically, this example solves
\begin{align*}
    \min_{\btheta} \bigg(\norm{f_{\btheta}-\sin(x)}_2^2 + \norm{f_{\btheta}+\sin(x)}_2^2\bigg),
\end{align*}
where $f_{\btheta}$ is a global neural network parametrized by $\btheta$ and $\norm{\cdot}_2^2$ is a functional on $[0,10]$ defined as $\norm{f}_2^2=\int_{0}^{10}f(x)^2dx.$ By taking derivative of the global function and set it to zero, we obtain
\begin{align*}
    &\frac{\partial}{\partial\btheta}\bigg(\norm{f_{\btheta}-\sin(x)}_2^2 + \norm{f_{\btheta}+\sin(x)}_2^2\bigg)\\
    &=\int_{0}^{10}2\bigg(f_{\btheta}-\sin(x)\bigg)\frac{df_{\btheta}}{d\btheta}+2\bigg(f_{\btheta}+\sin(x)\bigg)\frac{df_{\btheta}}{d\btheta} dx = 0 \\
    &\Rightarrow  \int_{0}^{10} 4f_{\btheta}\frac{df_{\btheta}}{d\btheta} dx = 0 \\
    &\Rightarrow f_{\btheta} = 0.
\end{align*}
This implies that the global model did not learn anything from both datasets and simply returns a constant estimation. To remedy this issue, one needs to additionally implement an additional personalization step that fine-tunes the global model from local data.  This typically introduces extra computational costs or even (hyper-)parameters. On the other hand, a single $\mathcal{GP}$ model learned from \texttt{FGPR} can provide good interpolations for both devices. This demonstrates the advantage of the automatic personalization property of \texttt{FGPR}.

\begin{remark}
Despite \texttt{FGPR} being a global modeling approach, in our empirical section we will compare with personalized FL using NNs when the data distributions are heterogeneous. 
\end{remark}

\section{Theoretical Results}
\label{sec:theory}

Proving convergence of \texttt{FGPR} introduces new challenges due to correlation and the decentralized nature of model estimation. Recall that in ERM (e.g., in deep learning), the loss function for device $k$ can be written as $L_k(\btheta)=\frac{1}{N_k}\sum_{i=1}^{N_k} L_{k,i}(\btheta;\{x_{k,i}^\intercal,y_{k,i}\})$, where $L_{k,i}(\btheta;\{x_{k,i}^\intercal,y_{k,i}\})$ is the loss function evaluated at $\btheta$ using data point $\{x_{k,i}^\intercal,y_{k,i}\}$. In $\mathcal{GP}$s, the loss function cannot be expressed as a summation form since all data points are correlated. This correlation renders the stochastic gradient  a biased estimator of the full gradient. To the best of our knowledge, only a recent work from \citep{chen2020stochastic} has shown theoretical convergence results of centralized $\mathcal{GP}$ in a correlated setting. Adding to that, \texttt{FGPR} aggregates parameters that are estimated on only a partial dataset. 

% Despite this, proving convergence is a decentralized FL setting where data is partitioned across many edge devices parameters and parameters are estimated via  aggregation across local estimates is yet to be investigated and poses many additional challenges. 

In this section, we take a step forward in understanding the theoretical properties of $\mathcal{GP}$ estimated in a federated fashion. Specifically, we provide several probabilistic convergence results of \texttt{FGPR} under both homogeneous and heterogeneous clients and under both full and partial device participation settings.

To proceed, we define $\btheta_{\mathcal{K}}=(\theta_1,\ell)$ such that $\btheta=(\theta_1,\theta_2,\ell)$. Here, $\theta_1$ is the signal variance parameter, $\theta_2=\sigma$ is the noise parameter and $\ell$ is the length parameter. Denote by $\btheta^*\coloneqq(\theta_1^*,\theta_2^*,\ell^*)$ the true data-generating parameter that minimizes the population risk. In other words, $\btheta^*=\argmin_{\btheta}\mathbb{E}\left[\sum_{k=1}^Kp_kL_k(\btheta;D_k)\right]$. We impose a structure on the covariance function such that $\mathcal{K}(\cdot,\cdot;\bm{\theta}_{\mathcal{K}})= \theta^2_1 \mathrm{k}_f(\cdot,\cdot)$ where $\mathrm{k}_f(\cdot,\cdot)$ is a known kernel function. This form of covariance function is ubiquitous and widely adopted. For instance, the Mat\'ern covariance is in the form of
\begin{align*}
    &C_v(x_1,x_2)=\theta^2_1\frac{2^{1-v}}{\Gamma(v)}\left(\sqrt{2v}\frac{\norm{x_1-x_2}}{\ell}\right)^vK_v\left(\sqrt{2v}\frac{\norm{x_1-x_2}}{\ell}\right)  + \theta_2^2\mathbb{I}_{x_1=x_2}
\end{align*}
where $v$ is a positive scalar and $K_v$ is the modified Bessel function of the second kind and $\mathbb{I}$ is an indicator function. In this example, $\mathrm{k}_f(x_1,x_2)=\frac{2^{1-v}}{\Gamma(v)}\left(\sqrt{2v}\frac{\norm{x_1-x_2}}{\ell}\right)^vK_v\left(\sqrt{2v}\frac{\norm{x_1-x_2}}{\ell}\right)$. Another example is the RBF covariance:
\begin{align*}
    C_{RBF}(x_1,x_2)=\theta_1^2\exp\left(\frac{\norm{x_1-x_2}^2}{2\ell^2}\right) + \theta_2^2\mathbb{I}_{x_1=x_2}.
\end{align*}
There are also many other examples such as the Ornstein–Uhlenbeck covariance and the periodic covariance \citep{williams2006gaussian}.

\begin{remark}
A more general setting is to consider the compound covariance function that is in the form of $\mathcal{K}(\cdot,\cdot;\bm{\theta}_{\mathcal{K}})=\sum_{i=1}^{A}\theta_{1i}^2\mathrm{k}_{f_i}(\cdot,\cdot)$. For simplicity, in the theoretical analysis, we assume $A=1$. However, our proof techniques can be easily extended to the scenario where $A>1$.
\end{remark}

In the theoretical analysis, we will show the explicit convergence bounds on $\theta_1$ and $\theta_2$. The convergence behavior of the length parameter $\ell$ is still an open problem \citep{chen2020stochastic}. The key reason is that one needs to apply the eigendecomposition technique to the kernel function and carefully analyze the lower and upper bounds of eigen-functions. The length parameter $\ell$, however, lies in the denominator of a kernel function. In this case, it is extremely challenging to write the kernel function in a form of eigenvalues and bound them. To the best our knowledge, the work that studies convergence results of $\ell$ is still vacant even in centralized regimes.

\subsection{Assumptions}
To derive our convergence results, we make the following assumptions.
\begin{assumption}
\label{assumption:parameter}
The parameter space $\bm{\Theta}$ is a compact and convex subset of $\mathbb{R}^2$. Moreover, $(\theta_1^*,\theta_2^*)^\intercal\in \bm{\Theta}^{\mathrm{o}}$ and $\sup_{(\theta_1,\theta_2)^\intercal\in\bm{\Theta}}\norm{(\theta_1,\theta_2)^\intercal-(\theta_1^*,\theta_2^*)^\intercal}>0$, where $\bm{\Theta}^{\mathrm{o}}$ is the interior of set $\bm{\Theta}$.
% and $(\theta_1^*,\theta_2^*)^\intercal$ are the optimal parameters of $(\theta_1,\theta_2)^\intercal$ that minimize the population risk. In other words, $(\theta_1^*,\theta_2^*)^\intercal$ are the true data generating parameters such that $(\theta_1^*,\theta_2^*)^\intercal =  \argmin_{\theta_1,\theta_2}\mathbb{E}\left[\sum_{k=1}^Kp_k L_k(\btheta;D_k) \right]$.
\end{assumption}
This assumption  indicates that all parameter iterates are bounded and the global minimizer $(\theta_1^*,\theta_2^*)^\intercal$ exists. Without loss of generality, assume the lower (or upper) bound of the parameter space on each dimension is $\theta_{min}$ (or $\theta_{max}$).

\begin{assumption}
\label{assumption:norm}
The norm of the stochastic gradient is bounded. Specifically,
\begin{align*}
    0\leq \norm{g_k(\cdot;\xi_k^{(t)})}\leq G, \text{ for all } k\in[K], t\in[T].
\end{align*}
\end{assumption}

Here $T$ is defined as total number of iteration indices on each device. Mathematically, $T=R(E-1)$ and $[T]=\{0,\ldots,T\}$.

\begin{remark}
It is very common to assume the local loss functions are $L$-smooth, (strongly-)convex or the variance of the stochastic gradient is bounded. Here we do not make those assumptions.
\end{remark}

In the $\mathcal{GP}$ setting, the explicit convergence bound depends on the rate of decay of eigenvalues from a specific type of kernel function. In this paper, we study two types of kernel functions: (1) kernel functions with exponential eigen-decay rates; and (2) kernel functions with polynomial eigen-decay rates. Those translate to the following assumptions.

\begin{assumption*}%\label{a2a}
\label{assumption:smooth}
For each $k\in[K]$, the eigenvalues of kernel function $\mathrm{k}_f$ with respect to probability measure $\mu$ are $\{C_ke^{-b_kj}\}_{j=1}^\infty$, where $b_k>0$ and $C_k<\infty$. Without loss of generality, assume $C_k\leq 1$.
\end{assumption*}

\begin{assumption*}
\label{assumption:non-smooth}
For each $k\in[K]$, the eigenvalues of kernel function $\mathrm{k}_f$ with respect to probability measure $\mu$ are $\{C_kj^{-2b_k}\}_{j=1}^\infty$, where $b_k>\frac{\sqrt{21}+3}{4}$ and $C_k<\infty$. Without loss of generality, assume $C_k\leq 1$.
\end{assumption*}

\begin{remark}
Assumption \ref{assumption:smooth} is satisfied by smooth kernels such as RBF kernels and Assumption \ref{assumption:non-smooth} is satisfied by the non-smooth kernels such as Mat\'ern kernels.
\end{remark}

\subsection{Homogeneous Setting}
\label{subsec:3-1}

We first assume that data across all devices are generated from the same underlying process or distribution (i.e., homogeneous data). Mathematically, it indicates \citep{li2019convergence}
\begin{align*}
    \lim_{N_1,\ldots,N_k\to\infty}\left|\sum_{k=1}^Kp_k L_k(\btheta^*;D_k)-\sum_{k=1}^Kp_k L_k(\btheta_k^*;D_k)\right|=0.
\end{align*}
% \lim_{N_k\to\infty, \forall k\in[K]}
We briefly parse this expression. Since the data distribution across all devices are homogeneous, we know, for each $k$, $\btheta_k^*=\btheta^*$ as $N_k\to\infty$. Therefore, $\sum_{k=1}^Kp_k L_k(\btheta^*;D_k)=\sum_{k=1}^Kp_k L_k(\btheta_k^*;D_k)$. In Sec. \ref{subsec:hetero}, we will consider the heterogeneous data settings which is often more realistic in real-world applications.

To derive the convergence result, we scale $[g_k(\bm{\theta};\xi_k)]_1$ by a constant factor $s_1(M_k)=\tau\log M_k$ and $[g_k(\bm{\theta};\xi_k)]_2$ by $s_2(M_k)=M_k$, where $[g_k(\bm{\theta};\xi_k)]_i$ is the $i$-th component in the stochastic gradient. Those scaling factors are introduced to ensure $[g_k(\bm{\theta};\xi_k)]_1$ and $[g_k(\bm{\theta};\xi_k)]_2$ have the same scale in the theoretical analysis. To see this, we have
\begin{align*}
    &[g_k(\bm{\theta};\xi_k)]_1 \\
    &= \frac{\Tr\bigg[\bm{K}^{-1}(\bm{X}_{\xi},\bm{X}_{\xi})\left(\bm{I}-\bm{y}_{\xi}\bm{y}_{\xi}^T\bm{K}^{-1}(\bm{X}_{\xi},\bm{X}_{\xi})\right)\frac{\partial \bm{K}(\bm{X}_{\xi},\bm{X}_{\xi})}{\partial\theta_1} \bigg] }{2}\\
    &\asymp \log M_k.
\end{align*}
Therefore, scaling $[g_k(\bm{\theta};\xi_k)]_1$ by $\mathcal{O}(\log M_k)$ ensures that it has the same scale as $\frac{[g_k(\bm{\theta};\xi_k)]_2}{M_k}$.

\begin{remark}
The aforementioned scaling factors are only needed for convergence results. In practice, we observe that $\tau$ has minimal influence on the model performance. 
\end{remark}

Our first Theorem shows that \texttt{FGPR} using RBF kernels converges if all devices participated in the training. 

% We provide an explicit convergence bound on $|\bar{\theta}_1^{(T)}-\theta_1^*|^2+|\bar{\theta}_2^{(T)}-\theta_2^*|^2$ at iteration $T$, where $\theta_1^*,\theta_2^*$ are the optimal model parameters such that
% \begin{align*}
%     (\theta_1^*,\theta_2^*) = \argmin_{\theta_1,\theta_2}\sum_{k=1}^Kp_k L_k(\btheta;D_k).
% \end{align*}

\begin{theorem}
\label{theorem:parameter_full_exp}
(RBF kernels, synchronous update) Suppose Assumptions \ref{assumption:parameter}-\ref{assumption:smooth} hold. At each communication round, assume $|\mathcal{S}_c|=K$. If $\eta^{(t)}=\mathcal{O}(\frac{1}{t})$ (i.e., a decay learning rate scheduler), then for some constants $\beta_1, C_{\btheta}, c_{\btheta}>0, \epsilon_k\in(0,\frac{1}{2})$, at iteration $T$, with probability at least $\min_{k}\left(1-C_{\btheta}T\exp\left\{-c_{\btheta}\left(\log M_k\right)^{2\epsilon_k}\right\}\right)$,
\begin{align*}
    &\left|\bar{\theta}_1^{(T)}-\theta_1^*\right|^2+\left|\bar{\theta}_2^{(T)}-\theta_2^*\right|^2\\
    &\qquad \leq\frac{2\beta_1^2\left(8(E-1)^2+2\right)G^2}{T+1}+\mathcal{O}\left(\max_k\frac{\log M_k}{M_k} +\sum_{k=1}^Kp_k(\log M_k)^{\epsilon_k-\frac{1}{2}}\right),
\end{align*}
and with probability at least $$\min_k\left(1-C_{\btheta}\left(\log\left( M_k^{\epsilon_k-\frac{1}{2}}\right)\right)^4T\exp\left\{-c_{\theta}M_k^{2\epsilon_k}\right\}\right),$$
\begin{align*}
    &\norm{\bar{\theta}_2^{(T)}-\theta_2^*}_2^2\leq\frac{2\beta_1^2\left(8(E-1)^2+2\right)G^2}{T+1}+\mathcal{O}\left(\max_k\frac{\log M_k}{M_k}  + \sum_{k=1}^Kp_kM_k^{\epsilon_k-\frac{1}{2}}\right).
\end{align*}
Here, constants $\beta_1, C_{\btheta}, c_{\btheta}$ only depend on $\theta_{min},\theta_{max}$ and $\{b_k\}_{k=1}^K$.
\end{theorem}

\begin{remark}
Recall that $T$ is the number of iterations. Theorem \ref{theorem:parameter_full_exp} implies that, with a high probability, the parameter iterate converges to the global optimal parameter at a rate of $\mathcal{O}(\frac{1}{T})$. This is credited to the unique structure of the $\mathcal{GP}$ loss function, which we refer to as relaxed convexity (See Lemma 4 and Lemma 5 in the Appendix). However, the $\mathcal{GP}$ loss function is \textbf{non-convex} \citep{williams2006gaussian}.
\end{remark}

\begin{remark}
In the upper bound, there is a term $\frac{2\beta_1^2\left(8(E-1)^2+2\right)G^2}{T+1}\sim\frac{(E-1)^2}{T+1}$, where $E$ is the number of local SGD steps. To ensure this term is decreasing with respect to $T$, one need to ensure $E$ does not exceed $\Omega(\sqrt{T})$. Otherwise, the \texttt{FGPR} will not converge. For instance, if $E=T$, then the \texttt{FGPR} is equivalent to the one-shot communication approach \citep{zhang2013communication}. If $E>T$, it means there is no communication among local devices.
\end{remark}

\begin{remark}
In addition to the $\mathcal{O}(\frac{1}{T})$ term, there is also a statistical error term $\mathcal{O}(\max_k\frac{\log M_k}{M_k} + \sum_{k=1}^Kp_kM_k^{\epsilon_k-\frac{1}{2}})$ that appeared in the upper bound. Theoretically, it indicates that a large batch size is capable of reducing error in parameter estimation. %However, from the computational perspective, a large batch size also incurs heavy computation burden due to the matrix inversion operation.
\end{remark}

\begin{remark}
From Theorem \ref{theorem:parameter_full_exp}, it can be seen that $\norm{\bar{\theta}_2^{(T)}-\theta_2^*}_2^2$ has smaller error term $\mathcal{O}\left(\sum_{k=1}^Kp_kM_k^{\epsilon_k-\frac{1}{2}}\right)$ than $\mathcal{O}\left(\sum_{k=1}^Kp_k(\log M_k)^{\epsilon_k-\frac{1}{2}}\right)$. This implies that the noise parameter $\theta_2$ is easier to estimate than $\theta_1$. This is intuitively understandable since $\theta_1$ has more complicated eigenvalue structure $(\bm{K}_{f,N})$ than $\theta_2$ (which is simply the identity matrix).
\end{remark}

Next, we study the convergence behavior under the asynchronous update (i.e., partial device participation) framework. In this scenario, only a portion of devices are actively sending their model parameters to the central server at each communication round.

\begin{theorem}
\label{theorem:parameter_partial_exp}
(RBF kernels, asynchronous update) Suppose Assumption \ref{assumption:parameter}-\ref{assumption:smooth} hold. At each communication round, assume $|\mathcal{S}_c|=K_{sample}<K$ number of devices are sampled according to the sampling probability $p_k$. If $\eta^{(t)}=\mathcal{O}(\frac{1}{t})$, then for some constants $C_{\btheta}, c_{\btheta}>0, \epsilon_k\in(0,\frac{1}{2})$, at iteration $T$, with probability at least $\min_{k}\left(1-C_{\btheta}T\exp\left(-c_{\btheta}\left(\log M_k\right)^{2\epsilon_k}\right)\right)$,
\begin{align*}
    &\mathbb{E}_{\mathcal{S}_c}\left\{\left|\bar{\theta}_1^{(T)}-\theta_1^*\right|^2+\left|\bar{\theta}_2^{(T)}-\theta_2^*\right|^2\right\}\\
    &\qquad \leq\frac{2\beta_1^2\left(\frac{1}{|\mathcal{S}_c|}4E^2+8(E-1)^2+2\right)G^2}{T+1}+\mathcal{O}\left(\max_k\frac{\log M_k}{M_k} +\sum_{k=1}^Kp_k(\log M_k)^{\epsilon_k-\frac{1}{2}}\right),
\end{align*}
and with probability at least $$\min_k\left(1-C_{\btheta}\left(\log\left( M_k^{\epsilon_k-\frac{1}{2}}\right)\right)^4T\exp\left\{-c_{\btheta}M_k^{2\epsilon_k}\right\}\right),$$
\begin{align*}
    &\mathbb{E}_{\mathcal{S}_c}\left\{\norm{\bar{\theta}_2^{(T)}-\theta_2^*}_2^2\right\}\\
    &\leq\frac{2\beta_1^2\left(\frac{1}{|\mathcal{S}_c|}4E^2+8(E-1)^2+2\right)G^2}{T+1}+\mathcal{O}\left(\max_k\frac{\log M_k}{M_k} + \sum_{k=1}^Kp_kM_k^{\epsilon_k-\frac{1}{2}}\right).
\end{align*}
Here $\mathbb{E}_{\mathcal{S}_c}(x)\coloneqq\sum_{k=1}^Kp_kx_k$.
\end{theorem}

\begin{remark}
Under the asynchronous update setting, a similar convergence guarantee holds. The only difference is that the number of active devices $|\mathcal{S}_c|$ plays a role in the upper bound. Numerically, the ratio $\frac{E^2}{|\mathcal{S}_c|}$ enlarges the upper bound and impedes the convergence rate. As $|\mathcal{S}_c|$ grows (i.e., more devices participate in the training), the ratio $\frac{E^2}{|\mathcal{S}_c|}$ decreases. 
\end{remark}

% +N_k^{\frac{(2+\alpha_k)(4b_k+3)}{4b_k(2b_k-1)}-1}

Our next theorem provides explicit convergences rate for \texttt{FGPR} with Mat\'ern kernels, under both  a synchronous and asynchronous update scheme.

\begin{theorem}
\label{theorem:parameter_matern}
(Mat\'ern kernels) Suppose Assumptions \ref{assumption:parameter}-\ref{assumption:norm} and \ref{assumption:non-smooth} hold, 
% \begin{enumerate}

    (1) At each communication round, assume $|\mathcal{S}_c|=K$. If $\eta^{(t)}=\mathcal{O}(\frac{1}{t})$, then for some constants $C_{\btheta},c_{\btheta}>0$, $\beta_1>0,b_k>\frac{(\sqrt{21}+3)}{4}$ and $0<\alpha_k<\frac{1}{2}$, with probability at least $\min_k\left(1-C_{\btheta}T(\log( M_k^{\epsilon_k-\frac{1}{2}}))^4\exp\{-c_{\btheta}M_k^{2\epsilon_k}\}\right)$, 
    \begin{align*}
    &\norm{\bar{\theta}_2^{(T)}-\theta_2^*}_2^2\leq\frac{2\beta_1^2\left(8(E-1)^2+2\right)G^2}{T+1} \\
    &\qquad + \mathcal{O}\left(\max_kM_k^{-\frac{8b_k^2-12b_k-6-3\alpha_k-4\alpha_kb_k}{8b_k^2-4b_k}}\right)+ \mathcal{O}\left(\sum_{k=1}^Kp_kM_k^{\epsilon_k-\frac{1}{2}}\right).
    \end{align*}
    Additionally, 
    \begin{align*}
    &\norm{\nabla L(\bbtheta^{(T)})}^2_2\leq \frac{2\beta_1^2\left(8(E-1)^2+2\right)G^2}{4\theta_{min}^4(T+1)}+ \mathcal{O}\left(\max_k\left\{M_k^{\frac{(2+\alpha_k)(4b_k+3)}{4b_k(2b_k-1)}-1}+\sum_{k=1}^Kp_kM_k^{\epsilon_k-\frac{1}{2}}\right\}\right).
    \end{align*}
    
    (2) At each communication round, assume $|\mathcal{S}_c|=K_{sample}$, number of devices are sampled according to the sampling probability $p_k$. If $\eta^{(t)}=\mathcal{O}(\frac{1}{t})$, then for some constants $C_{\btheta},c_{\btheta}>0$, $\beta_1>0,b_k>\frac{(\sqrt{21}+3)}{4}$ and $0<\alpha_k<\frac{1}{2}$, with probability at least $\min_k\left(1-C_{\btheta}T\left(\log\left( M_k^{\epsilon_k-\frac{1}{2}}\right)\right)^4\exp\{-c_{\btheta}M_k^{2\epsilon_k}\}\right)$, 
    \begin{align*}
    &\mathbb{E}_{\mathcal{S}_c}\left\{\norm{\bar{\theta}_2^{(T)}-\theta_2^*}_2^2\right\}\\
    &\leq\frac{2\beta_1^2\left(\frac{4E^2}{|\mathcal{S}_c|}+8(E-1)^2+2\right)G^2}{T+1} + \mathcal{O}\left(\max_kM_k^{-\frac{8b_k^2-12b_k-6-3\alpha_k-4\alpha_kb_k}{8b_k^2-4b_k}}\right)  + \mathcal{O}\left(\sum_{k=1}^Kp_kM_k^{\epsilon_k-\frac{1}{2}}\right).
    \end{align*}
      Additionally, 
    \begin{align*}
    &\mathbb{E}_{\mathcal{S}_c}\left\{\norm{\nabla L(\bbtheta^{(T)})}^2_2\right\}\\
    &\leq \frac{2\beta_1^2\left(\frac{4E^2}{|\mathcal{S}_c|}+8(E-1)^2+2\right)G^2}{4\theta_{min}^4(T+1)}+ \mathcal{O}\left(\max_k\left\{M_k^{\frac{(2+\alpha_k)(4b_k+3)}{4b_k(2b_k-1)}-1}+\sum_{k=1}^Kp_kM_k^{\epsilon_k-\frac{1}{2}}\right\}\right).
    \end{align*}
% \end{enumerate}
\end{theorem}

\begin{remark}
It can be seen that the \texttt{FGPR} using Mat\'ern kernel has larger statistical error than the one using RBF kernel. In the RBF kernel, the statistical error is partially affected by $\mathcal{O}\left(\max_k\frac{log M_k}{M_k}\right)$ (Theorems \ref{theorem:parameter_full_exp},\ref{theorem:parameter_partial_exp}) while this term becomes $\mathcal{O}\left(\max_kM_k^{-\frac{8b_k^2-12b_k-6-3\alpha_k-4\alpha_kb_k}{8b_k^2-4b_k}}\right)$ in the Mat\'ern kernel. The latter one is larger since $b_k>\frac{(\sqrt{21}+3)}{4}$ and $\alpha_k\in(0,0.5)$. This difference arises from the fact that the Mat\'ern kernel has slower eigenvalue decay rate (determined by $b_k$) than the RBF kernel (i.e., polynomial vs. exponential). This slow decay rate leads to slower convergence and larger statistical error. When $b_k$ becomes larger, the decay rate becomes faster and the influence of $\mathcal{O}\left(\max_kM_k^{-\frac{8b_k^2-12b_k-6-3\alpha_k-4\alpha_kb_k}{8b_k^2-4b_k}}\right)$ gets smaller. In this case, the statistical error is dominated by $\mathcal{O}\left(\sum_{k=1}^Kp_kM_k^{\epsilon_k-\frac{1}{2}}\right)$, which is same as the one in the RBF kernel.
\end{remark}

\begin{remark}
In addition to the convergence bound on parameter iterates, we also provide an upper bound on the full gradient $\norm{\nabla L(\bbtheta^{(T)})}^2_2$. This bound scales the same as the bound for $\norm{\bar{\theta}_2^{(T+1)}-\theta_2^*}_2^2$. 
\end{remark}

\begin{remark}
For Mat\'ern kernel, there is no explicit convergence guarantee for parameter $\bar{\theta}_1$. The reason is that it is very hard to derive the lower and upper bounds for the SG for Mat\'ern kernel. However, Theorem \ref{theorem:parameter_matern} shows that both $\bar{\theta}_2$ and the full gradient converge at rates of $\mathcal{O}(\frac{1}{T})$ subject to statistical errors.
\end{remark}

%~~~~~~~~~~~~~~~~~~~~~~~~~~~~

\subsection{Heterogeneous Setting}
\label{subsec:hetero}

In this section, we consider the scenario where data from all devices are generated from several different processes or distributions. Equivalently, this indicates
\begin{align*}
    \mathbb{P}\left(
    \left|\sum_{k=1}^Kp_k L_k(\btheta^*;D_k)-\sum_{k=1}^Kp_k L_k(\btheta_k^*;D_k)\right|=0\right)=0.
\end{align*}
Since the data are heterogeneous, the weighted average of $L_k(\btheta_k^*;D_k)$ can be very different from $L(\btheta^*)$. 

% Here, we add a probability statement since $\sum_{k=1}^Kp_k L_k(\btheta^*;D_k)=\sum_{k=1}^Kp_k L_k(\btheta_k^*;D_k)$ can happen numerically. 

% In this case, it is difficult to provide a bound on $\norm{\bbtheta^{(T)}-\btheta^*}_2$. However, we can prove that our proposed algorithm converges to a stationary point at a rate of $\mathcal{O}(\frac{1}{T})$.

% +\frac{\log N_k}{N_k}

\begin{theorem}
\label{theorem:exp_hetero}
(RBF kernels) Suppose Assumption \ref{assumption:parameter}-\ref{assumption:smooth} hold. At each communication round, assume $|\mathcal{S}_c|=K$. If $\eta^{(t)}=\mathcal{O}(\frac{1}{t})$, then for some constants $C_{\btheta}, c_{\btheta}>0, \epsilon_k\in(0,\frac{1}{2})$, at iteration $T$, with probability at least $\min_k\{1-C_{\btheta}(\log( M_k^{\epsilon_k-\frac{1}{2}}))^4T\exp\{-c_{\btheta}M_k^{2\epsilon_k}\}\}$,
\begin{align*}
    &\norm{\nabla L(\bbtheta^{(T)})}^2_2\leq \frac{2\beta_1^2\left(8(E-1)^2+2\right)G^2}{4\theta_{min}^4(T+1)} +\mathcal{O}\left(\max_k\frac{\log M_k}{M_k}+\sum_{k=1}^Kp_kM_k^{\epsilon_k-\frac{1}{2}}\right).
\end{align*}
On the other hand, at each communication round, assume $|\mathcal{S}_c|=K_{sample}$ number of devices are sampled according to the sampling probability $p_k$. Then we have
\begin{align*}
    &\mathbb{E}_{\mathcal{S}_c}\left\{\norm{\nabla L(\bbtheta^{(T)})}^2_2\right\}\leq \frac{2\beta_1^2\left(\frac{1}{|\mathcal{S}_c|}4E^2+8(E-1)^2+2\right)G^2}{4\theta_{min}^4(T+1)} +\mathcal{O}\left(\max_k\frac{\log M_k}{M_k}+\sum_{k=1}^Kp_kM_k^{\epsilon_k-\frac{1}{2}}\right).
\end{align*}
\end{theorem}

\begin{remark}
In the heterogeneous setting, we show that the \texttt{FGPR} algorithm will converge to a critical point of $L(\cdot)$ at a rate of $\mathcal{O}(\frac{1}{T})$ subject to a statistical error. The upper bound of $\norm{\nabla L(\bbtheta^{(T)})}^2_2$ has the same form as the one for $\norm{\bar{\theta}_2-\theta^*_2}_2^2$.
\end{remark}

For the Mat\'ern Kernel, we have the same upper bounds on $\norm{\nabla L(\bbtheta^{(T)})}^2_2$ and $\mathbb{E}_{\mathcal{S}_c}\left\{\norm{\nabla L(\bbtheta^{(T)})}^2_2\right\}$ as those in the Theorem \ref{theorem:parameter_matern}. This implies that the heterogeneous data distribution has little to no influence on the convergence behavior. The reason is that the heterogeneous definition is based on true parameters $\btheta^*$ and $\{\btheta^*_k\}_{k=1}^K$. The main Theorem, however, states that our algorithm will converge to a stationary point. In the non-convex scenario, the stationary point might be different from the true parameter.

% \begin{theorem}
% \label{theorem:mat_hetero}
% (Mat\'ern kernels) Suppose Assumption \ref{assumption:parameter}-\ref{assumption:norm} and \ref{assumption:non-smooth} hold. At each communication round, assume $|\mathcal{S}_c|=K$. If $\eta^{(t)}=\mathcal{O}(\frac{1}{t})$, then for some constants $C_{\btheta},c_{\btheta}>0$, $\beta_1>0,b_k>(\sqrt{21}+3)/4$ and $0<\alpha_k<1/2$, with probability at least $\min\{1-C_{\btheta}(T+1)(\log( M_k^{\epsilon_k-\frac{1}{2}}))^4\exp\{-c_{\btheta}M_k^{2\epsilon_k}\}\}$, 
% \begin{align*}
%     &\norm{\nabla L(\bbtheta^{(T)})}^2_2\leq \frac{2\beta_1^2\left(8(E-1)^2+2\right)G^2}{4\theta_{min}^4T}+ \mathcal{O}\left(\max_k\left\{M_k^{\frac{(2+\alpha_k)(4b_k+3)}{4b_k(2b_k-1)}-1}+\sum_{k=1}^Kp_kM_k^{\epsilon_k-\frac{1}{2}}+N_k^{\frac{(2+\alpha_k)(4b_k+3)}{4b_k(2b_k-1)}-1}\right\}\right).
% \end{align*}
% On the other hand, at each communication round, assume $|\mathcal{S}_c|=\floor{\kappa K}, \kappa\in(0,1)$, number of devices are sampled according to the sampling probability $p_k$. Then we have
% \begin{align*}
%     &\mathbb{E}_{\mathcal{S}_c}\left\{\norm{\nabla L(\bbtheta^{(T)})}^2_2\right\}\\
%     &\leq \frac{2\beta_1^2\left(\frac{4E^2}{|\mathcal{S}_c|}+8(E-1)^2+2\right)G^2}{4\theta_{min}^4T}+ \mathcal{O}\left(\max_k\left\{M_k^{\frac{(2+\alpha_k)(4b_k+3)}{4b_k(2b_k-1)}-1}+\sum_{k=1}^Kp_kM_k^{\epsilon_k-\frac{1}{2}}+N_k^{\frac{(2+\alpha_k)(4b_k+3)}{4b_k(2b_k-1)}-1}\right\}\right).
% \end{align*}
% \end{theorem}

Overall, in this theoretical section, we show that the \texttt{FGPR} is guaranteed to converge under both homogeneous setting and heterogeneous setting, regardless of the synchronous updating or the synchronous updating.

% \label{app:exp}
%     \begin{figure*}[!htbp]
%     \vskip -0.1in
%     \centering
%     \centerline{\includegraphics[width=\columnwidth]{figs/fig1.pdf}}
%     \caption{1}
%     \label{fig:femnist}
%     \vskip -0.1in
% \end{figure*}

% \label{app:exp}
%     \begin{figure*}[!htbp]
%     \vskip -0.1in
%     \centering
%     \centerline{\includegraphics[width=\columnwidth]{figs/fig2.pdf}}
%     \caption{1}
%     \label{2}
%     \vskip -0.1in
% \end{figure*}

% \label{app:exp}
%     \begin{figure*}[!htbp]
%     \vskip -0.1in
%     \centering
%     \centerline{\includegraphics[width=\columnwidth]{figs/fig3.pdf}}
%     \caption{1}
%     \label{3}
%     \vskip -0.1in
% \end{figure*}

% \label{app:exp}
%     \begin{figure*}[!htbp]
%     \vskip -0.1in
%     \centering
%     \centerline{\includegraphics[width=\columnwidth]{figs/fig4.pdf}}
%     \caption{1}
%     \label{4}
%     \vskip -0.1in
% \end{figure*}

% $\\
%     &=\frac{1}{2N}\Tr\left[(\theta^{(t)}_1\bm{K}_{f,N}+\theta^{(t)}_2\bm{I}_N)^{-1}(\bm{I}_N-(\theta^*_1\bm{K}_{f,N}+\theta^*_2\bm{I}_N)(\theta^{(t)}_1\bm{K}_{f,N}+\theta^{(t)}_2\bm{I}_N)^{-1})\frac{\partial (\theta^{(t)}_1\bm{K}_{f,N}+\theta^{(t)}_2\bm{I}_N)}{\partial\theta_i^{(t)}} \right]\\
%     &=\frac{1}{2N}(\theta_1^{(t)}-\theta_1^*)\sum_{j=1}^N\frac{\lambda_{1j}\lambda_{ij}}{\left(\theta_1^{(t)}\lambda_{1j}+\theta_2^{(t)}\lambda_{2j}\right)^2} + \frac{1}{2N}(\theta_2^{(t)}-\theta_2^*)\sum_{j=1}^N\frac{\lambda_{2j}\lambda_{ij}}{\left(\theta_1^{(t)}\lambda_{1j}+\theta_2^{(t)}\lambda_{2j}\right)^2}$

\section{Proof of Concept}
\label{sec:exp_sim}

We start by validating the theoretical results obtained in Sec. \ref{subsec:3-1}. We also provide sample experiments that shed light on key properties of \texttt{FGPR}.

\begin{figure*}[!htbp]
    \vskip -0.1in
    \centering
    \centerline{\includegraphics[width=0.9\columnwidth]{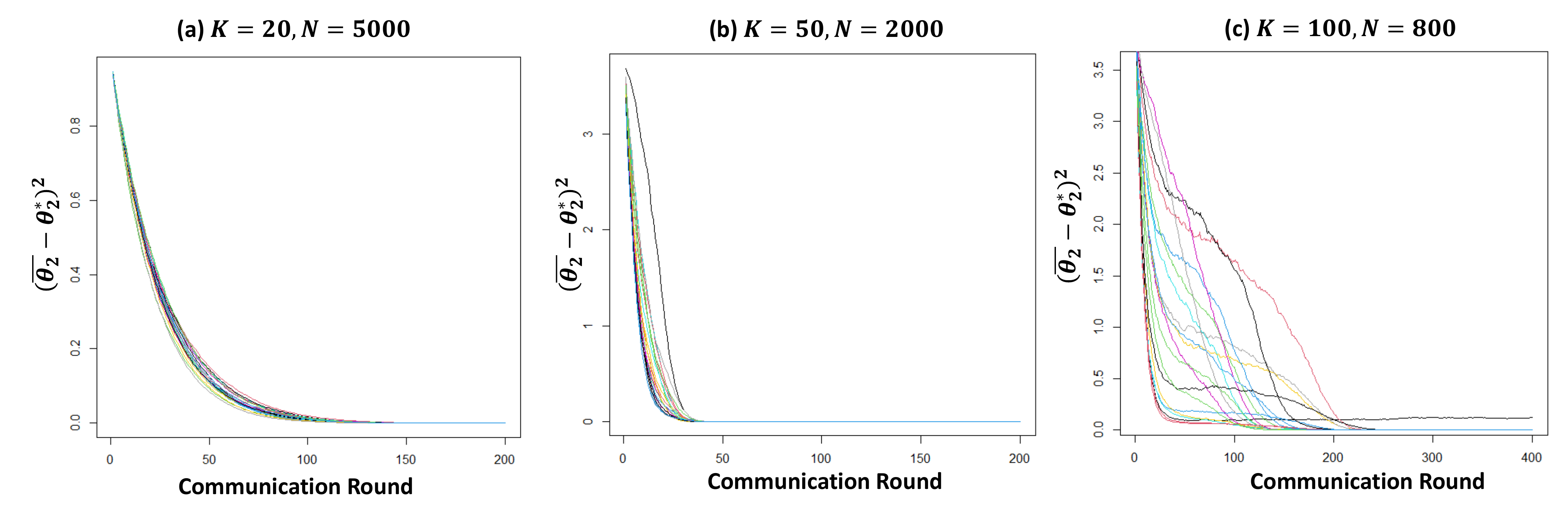}}
    \caption{(Mat\'ern$-3/2$ kernel) Evolution of $\norm{\bar{\theta}_2-\theta^*_2}_2^2$ over training epochs. The input dimension is 1. In the plot, each color represents an independent run.}
    \label{fig:d1}
    \vskip -0.2in
\end{figure*}

\begin{figure*}[!htbp]
    \vskip -0.1in
    \centering
    \centerline{\includegraphics[width=0.9\columnwidth]{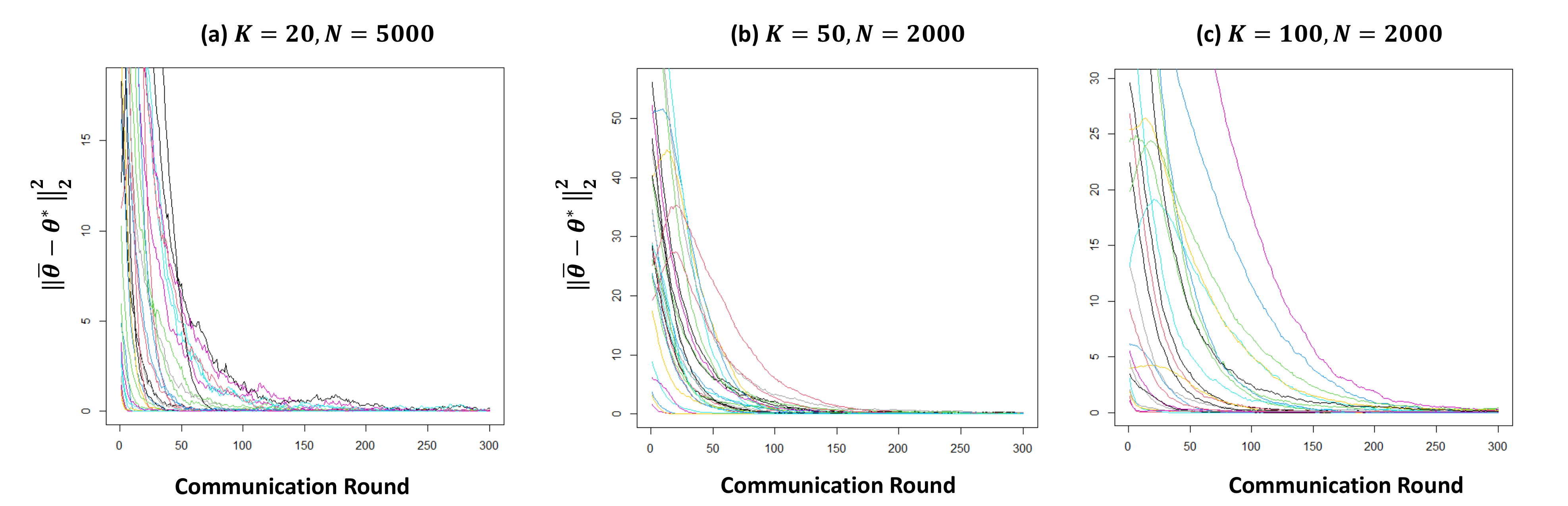}}
    \caption{(RBF kernel) Evolution of $\norm{\bar{\theta}-\theta^*}_2^2$ over training epochs. In the plot, each color represents an independent run. The input dimension $d$ is different for each run and $d\in\{1,\ldots,10\}$.}
    \label{fig:d2}
    \vskip -0.2in
\end{figure*}

\paragraph{Example 1: Homogeneous Setting with Balanced Data} We generate data from a $\mathcal{GP}$ with zero-mean and both a RBF and Mat\'ern$-3/2$ kernel. 
We consider $\theta_1\in[0.1,10]$, $\theta_2\in[0.01, 1]$ and a length parameter $\bm{\ell}\in[0.01,1]^d$. The input space is a $d$-dimensional unit cube $[0,1]^d$ in $\mathbb{R}^d$ with $d\in\{1,\ldots,10\}$ and the dimension of the output is one. We conduct 20 independent experiments. In each experiment, we first randomly sample $\theta_1,\theta_2,\bm{\ell}$ and $d$ to generate data samples from the $\mathcal{GP}$. In each scenario, we set $N_k=\frac{N}{K}$. This setting is homogeneous since, in each independent experiment, data among each device are generated from the same underlying stochastic process. We consider three scenarios: (1) $K=20, N=5000$, (2) $K=50,N=2000$, (3) $K=100,N=800$. Results are reported in Figure \ref{fig:d1} (Mat\'ern Kernel) and \ref{fig:d2} (RBF kernel). It can be seen that the convergence rate follows the $\mathcal{O}(\frac{1}{T})$ patterns. In Figure \ref{fig:d1}(c), we observe some fluctuations due to the small sample size. In this setting, since $N=800$, each device only has $M_k=8$ data points on average. Recall that in the theoretical analysis, we have shown that the convergence rate is also affected by $\mathcal{O}(\sum_{k=1}^Kp_kM_k^{\epsilon_k-0.5})$. 

\newpage

\begin{figure*}[!htbp]
    \centering
    \centerline{\includegraphics[width=0.5\columnwidth]{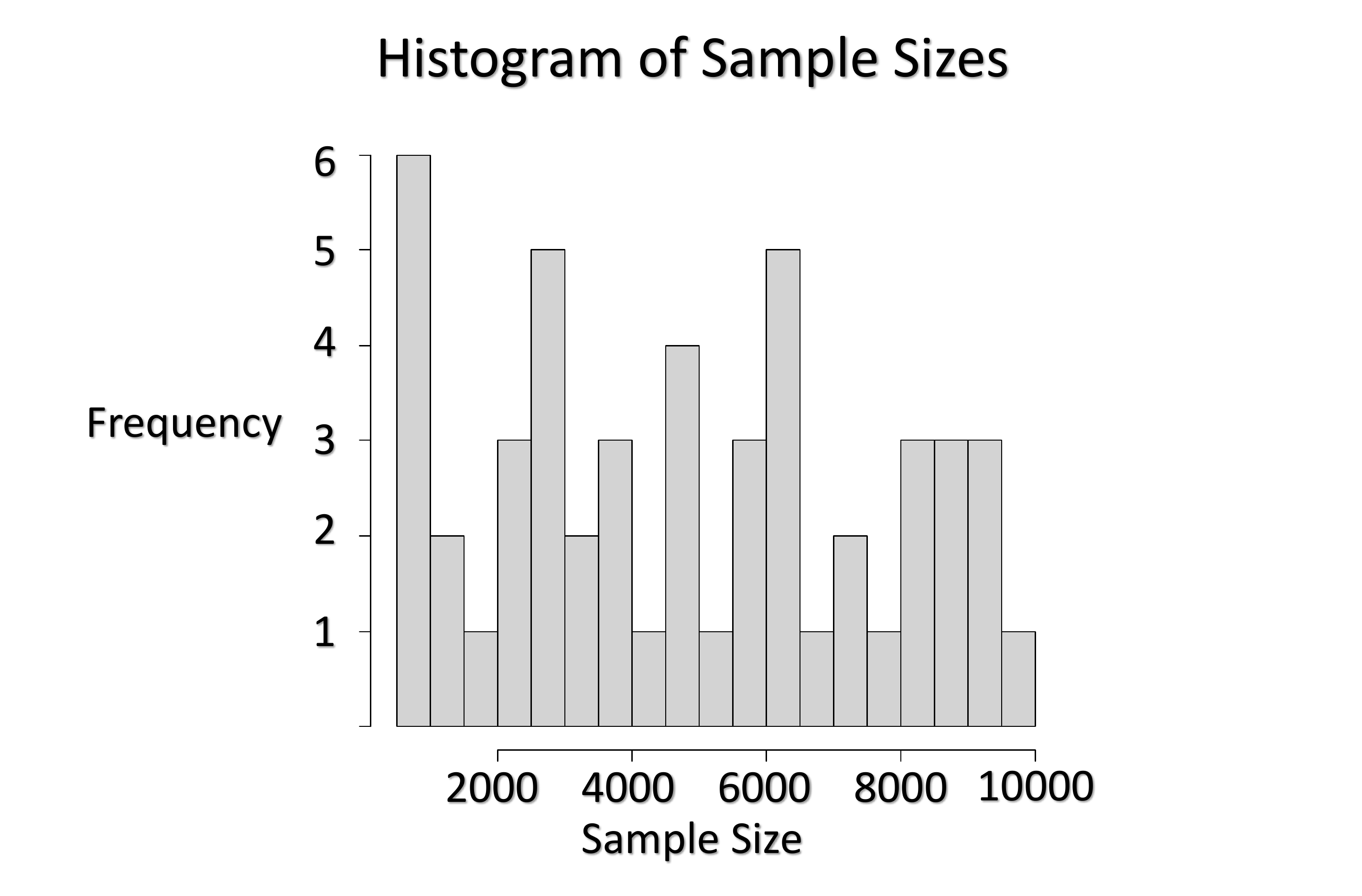}}
    \caption{Histogram of Sample Sizes.}
    \label{fig:hist}
    \vskip -0.1in
\end{figure*}

\begin{figure*}[!htbp]
    \vskip -0.2in
    \centering
    \centerline{\includegraphics[width=\columnwidth]{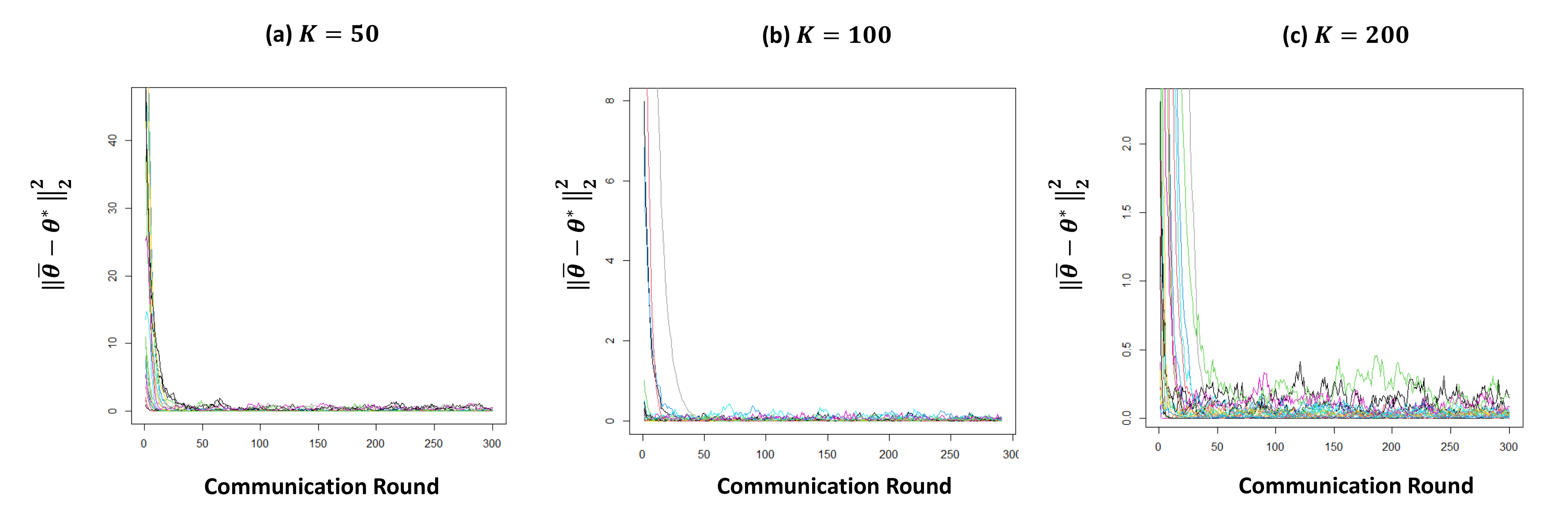}}
    \caption{(RBF kernel) Evolution of $\norm{\bar{\theta}-\theta^*}_2^2$ over training epochs using imbalanced data. In the plot, each color represents an independent run. The input dimension $d$ is different for each run and $d\in\{1,\ldots,10\}$.}
    \label{fig:d3}
    \vskip -0.1in
\end{figure*}

\textbf{Example 2: Homogeneous Setting with Unbalanced Data.} We use the same data-generating strategy as Example 1, but the sample sizes are unbalanced. Specifically, the number of data points in each device ranges from 10 to 10,000. The histogram of data distribution from one experiment is given in Figure \ref{fig:hist}. The convergence curves are plotted in Figure \ref{fig:d3}. Again, the convergence rate agrees with our theoretical finding. This simple example reveals a critical property of \texttt{FGPR}: \texttt{FGPR} can help devices with few observations recover true parameters (subject to statistical errors) or reduce prediction errors. We will further demonstrate this advantage in the heterogeneous setting in Sec. \ref{sec:exp_fidelity}.

% \begin{wrapfigure}{r}{0.45\textwidth}
%   \begin{center}
%     \includegraphics[width=0.5\textwidth]{figs/hist.pdf}
%   \end{center}
%   \caption{Histogram of Sample Sizes}
%   \label{fig:hist}
% \end{wrapfigure}

\textbf{Example 3: The Ability to Recover Accurate Prediction for Badly Initialized $\mathcal{GP}$.} When training an FL algorithm, it is not uncommon to initialize model parameter $\btheta$ near a bad stationary point. Here we provide one toy example. We simulate data from $y=sin(x)+\epsilon(x)$, where  $\epsilon\sim\mathcal{N}(0,0.2)$ and create two clients. Each client has 100 data points that are uniformly sampled from $[0,1]$. We artificially find a bad initial parameter $\btheta$ such that the fitted curve is just a flat line. This can be achieved by finding a $\btheta$ whose noise parameter $\theta_2$ is large. In this case, $\btheta=(1, 10, 1)$ and the fitted model interprets all data as noise and simply returns a flat line. 

\begin{figure*}[!htbp]
    \centering
    \centerline{\includegraphics[width=0.5\columnwidth]{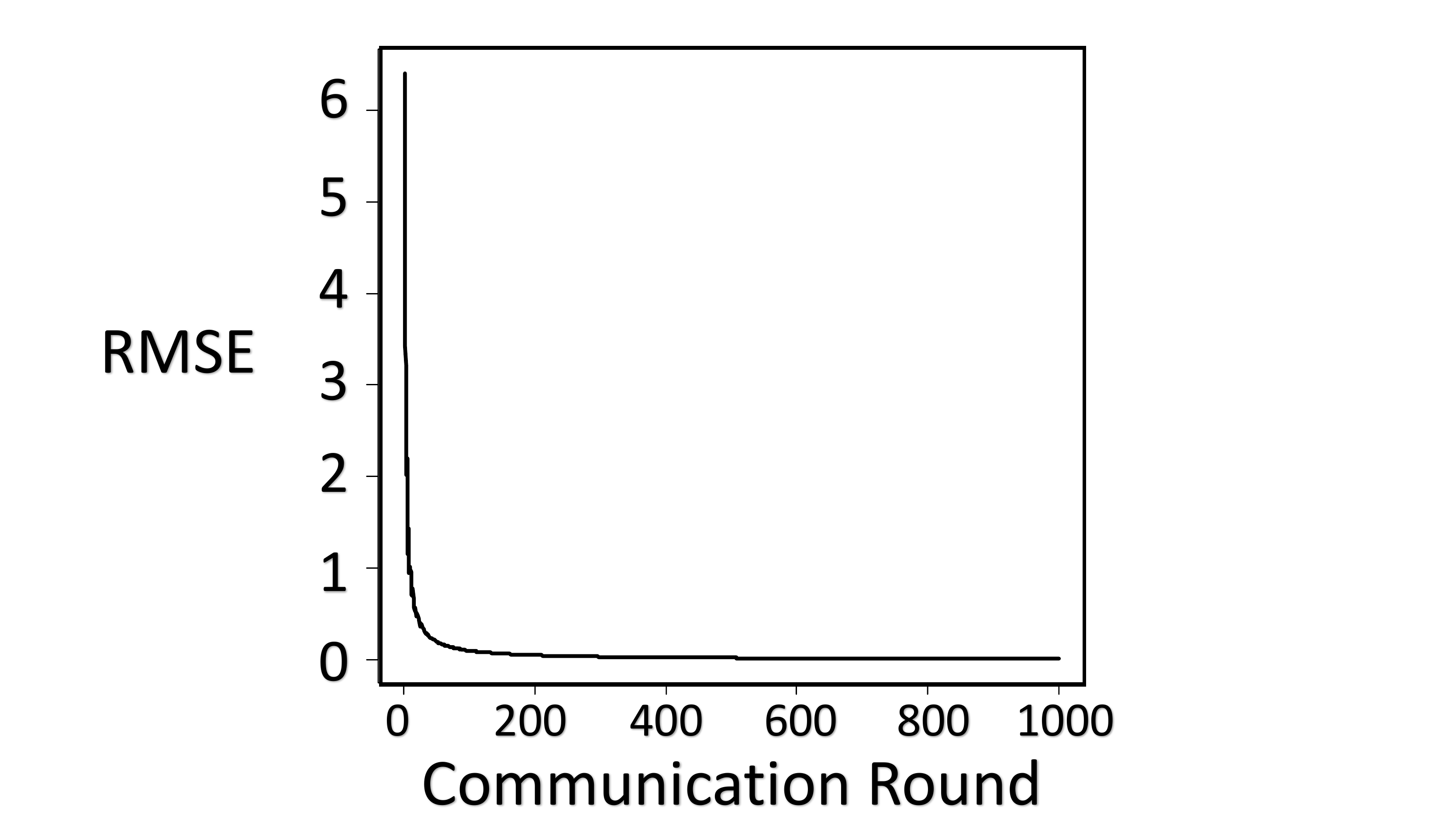}}
    \caption{Evolution of RMSE in Example 3.}
    \label{fig:d4}
\end{figure*}

We find that \texttt{FGPR} is robust to parameter initialization. We plot the evolution of averaged RMSE versus training epoch in Figure \ref{fig:d4}. It can be seen that, even when the parameter is poorly initialized, \texttt{FGPR} can still correct the wrong initialization after several communication rounds. This credits to the stochasticity in the SGD method. It is known that, in the ERM framework, SGD method can escape a bad stationary solution and converge to a solution that generalizes better \citep{wu2018sgd}.

% \begin{figure*}[!htbp]
%     \vskip -0.1in
%     \centering
%     \centerline{\includegraphics[width=\columnwidth]{figs/bad_gp.pdf}}
%     \caption{Results from Example 3. Each red line is the fitted $\mathcal{GP}$ curve and the dots represent the original training data.}
%     \label{fig:d4}
%     \vskip -0.1in
% \end{figure*}
% % \subsection{Multi-fidelity }

\section{Application I: Multi-fidelity Modeling}
\label{sec:exp_fidelity}

For many computer experiments, high-fidelity numerical simulations of complex physical processes typically require a significant amount of time and budget. This limits the number of data points researchers can collect and affects the modeling accuracy due to insufficient data. A major work trend has been proposed to augment the expensive data source with cheaper surrogates to overcome this hindrance. Multi-fidelity models are designed to fuse scant but accurate observations (i.e., high-fidelity, HF) with cheap and biased approximations (i.e., low-fidelity, LF) to improve the HF model performance.

Denote by $f_h$ a high-fidelity function and $f_l$ a low-fidelity function. Multi-fidelity approaches \citep{bailly2019multifidelity, cutajar2019deep, brevault2020overview} aim to use $f_l$ to better predict $f_h$. During the past decades, many multi-fidelity models have been proposed to fulfill this goal. We refer to \citep{fernandez2016review} and  \citep{peherstorfer2018survey} for detailed literature reviews. Among all of the methods, $\mathcal{GP}$ based approaches have caught most attention in this area due to their ability to incorporate prior beliefs, interpolate complex functional patterns and quantify uncertainties \citep{fernandez2016review}. The last ability is critical to fuse observations across different fidelities effectively. 

Within many applications, two specific models have been shown to be very competitive \citep{brevault2020overview}; the auto-regressive (\texttt{AR}) and the Deep $\mathcal{GP}$ (\texttt{Deep}) approaches. Both approaches model $f_h$ as shown below 
\begin{align*}
    f_h(x) = \rho(f_l(x), x) + \Delta(x),
\end{align*}
where $\rho(\cdot,\cdot)$ is a space-dependent non-linear transformation and $\Delta(x)$ is a bias term modeled through a $\mathcal{GP}$. 

More specifically, the \texttt{AR} model \citep{kennedy2000predicting} sets the transformation as a linear mapping such that $\rho(f_l(x), x)=\rho_c f_l(x)$, where $\rho_c$ is a constant. It then imposes a $\mathcal{GP}$ prior on $f_l$ and accordingly obtains its posterior $f^*_l$. As a result, one can derive the closed-form posterior distribution $p(f_h|f^*_l,x,y)$ and obtain the posterior predictive equation of the high-fidelity model. On the other hand, the \texttt{Deep} model \citep{cutajar2019deep} treats $\rho(f_l(x), x)$ as a deep Gaussian process to uncover highly complex relationships among $f_l$ and $f_h$ . \texttt{Deep} is one of the state-of-the-art multi-fidelity models. For more details, please refer to \citep{brevault2020overview}.

%The \texttt{NAR} method assigns a $\mathcal{GP}$ prior to $\rho(\cdot,\cdot)$ such that $ \rho(f_l(x), x)$ is a composition of two $\mathcal{GP}$s. The \texttt{NAR} then uses sampling techniques to approximate the posterior distribution of $f_h$ \citep{perdikaris2017nonlinear}. 

% During the past decades,  $\mathcal{GP}$-based multi-fidelity modeling techniques such as auto-regressive (\texttt{AR}) model and the deep Gaussian process model (\texttt{Deep}) \citep{brevault2020overview} have seen immense successes in this area. 

Nowadays, as data privacy gains increased importance, having access to data across multiple fidelities is often impractical as multiple clients can own data. This imposes a key challenge in multi-fidelity modeling approaches as effective inference on expensive high-fidelity models often necessitates the need to borrow strength from other information sources. Fortunately, in such a case, \texttt{FGPR} is a potential candidate that learns a $\mathcal{GP}$ prior without sharing data.

% Consider this example, in a robot control policy design experiment, a national laboratory often collects real-time data from a field test that costs a significant amount of budget. On the other hand, small research institutes usually try to replicate the system with a smaller budget by creating lower-fidelity models and, as such, borrowing strength across different fidelities is critical. 

In this section, we test the viability of \texttt{FGPR} in multi-fidelity modeling. In the context of \texttt{FGPR}, each client has one level of data fidelity and our goal is to improve HF model predictions. During the training process, clients only share their model parameters with a central orchestrator. We benchmark \texttt{FGPR} with several state-of-the-art multi-fidelity models. Interestingly, our results (Table \ref{table:RMSE}) show that \texttt{FGPR} not only preserves privacy but also can provide superior performance than centralized multi-fidelity approaches. 

% This demonstrates that, not only does \texttt{FGPR} preserve privacy but also can improve model accuracy across various state-of-the-art. 

% We train \texttt{FGPR} using learning rate $0.01$ with batch size $16$. The communication step happens every 2 local epochs. The number of communication round is 100. 

Below we detail the benchmark models: (1) \texttt{Separate} which simply fits a single $\mathcal{GP}$ to the HF dataset without any communication. This means the HF dataset does not use any information from the LF dataset; (2) the \texttt{AR} method \citep{kennedy2000predicting}. \texttt{AR} is the most classical and widely-used multi-fidelity modeling approach \citep{laurenceau2008building, fernandez2016review, bailly2019multifidelity}; (3) the \texttt{Deep} model \citep{damianou2013deep} highlighted above. All output values are standardized into mean 0 and variance 1.

We start with two simple illustrative examples from \citep{cutajar2019deep} and then benchmark all models on five well-known models in the multi-fidelity literature.

\textbf{Example 1: Linear Example} - We first present a simple one dimensional linear example where $x\in[0,1]$. The low and high fidelity models are given by \citep{cutajar2019deep}
\begin{align*}
    y_l(x) &= \frac{1}{2}y_h(x) + 10(x-\frac{1}{2}) + 5,\\
    y_h(x) &= (6x-2)^2\sin(12x-4),
\end{align*}
where $y_l(\cdot)$ is the output from the LF model and $y_h(\cdot)$ is the output from the HF model. We simulate 100 data points from the LF model and 20 data points from the HF model. The number of testing data points is 1,000. 

% The root mean square error (RMSE) from LF model is $6.66\times 10^{-5}$ and the RMSE from HF model is $7.67\times 10^{-3}$. Next, we fit local $\mathcal{GP}$ on each device without communication (i.e., use local models). The RMSE are $7.30\times10^{-4}$ and $7.93\times10^{-3}$, respectively. 

\textbf{Example 2: Nonlinear Example} - The one dimensional non-linear example ($x\in[0,2]$) is given as \citep{cutajar2019deep}
\begin{align*}
    y_l(x) &= \cos(15x),\\
    y_h(x) &= x\exp^{y_l(2x-0.2)}-1.
\end{align*}
We use the same data-generating strategy in Example 1.

\begin{figure*}[!htbp]
    \vskip -0.1in
    \centering
    \centerline{\includegraphics[width=\columnwidth]{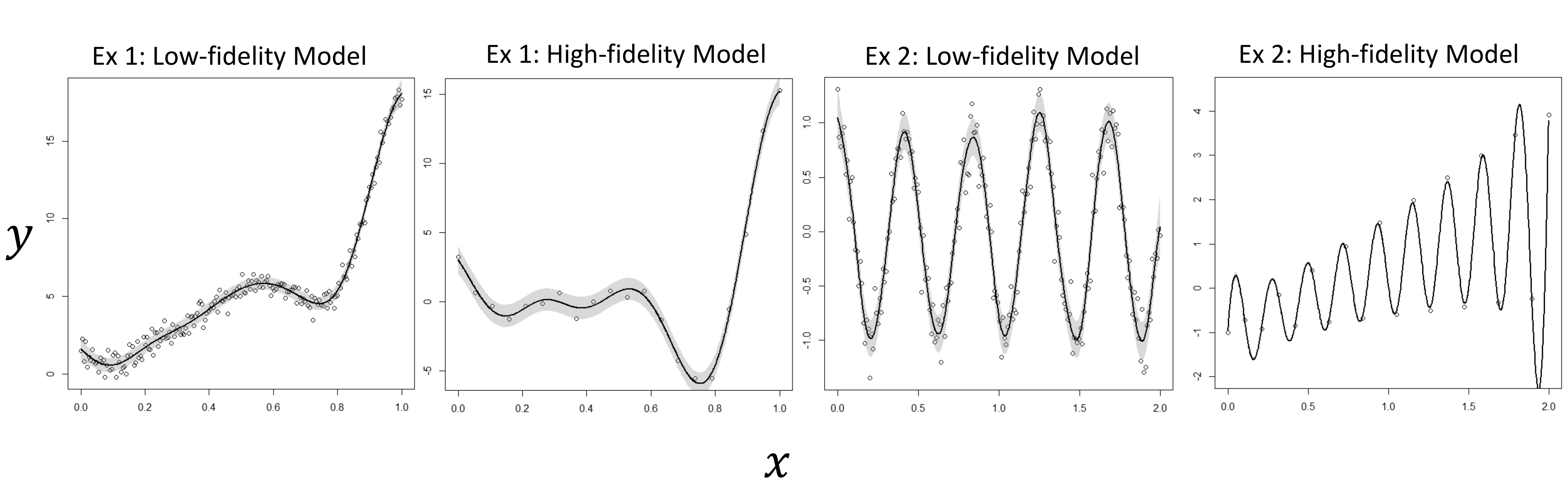}}
    \caption{Results of Example 1 and 2. The black lines are predicted mean and grey areas are the 95\% confidence intervals.}
    \label{fig:MF1}
    \vskip -0.1in
\end{figure*}

\begin{figure*}[!htbp]
    \vskip -0.1in
    \centering
    \centerline{\includegraphics[width=0.6\columnwidth]{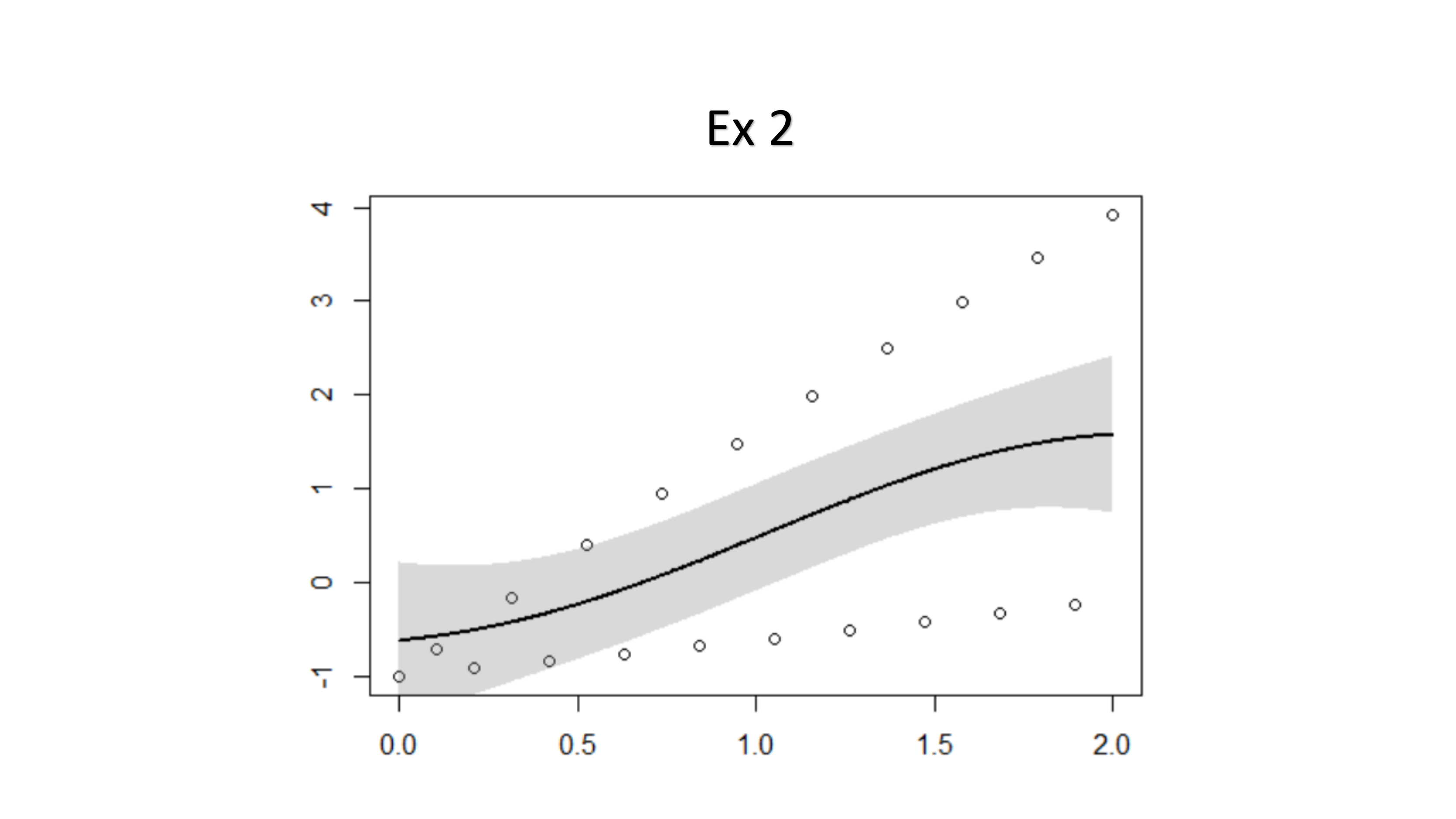}}
    \caption{Results of Example 2.}
    \label{fig:MF2}
    \vskip -0.1in
\end{figure*}

The results from both examples are plotted in Figure \ref{fig:MF1}. The results provide a simple proof-of-concept that the learned  \texttt{FGPR} is able to accurately predict the HF model despite sparse observations. Additionally, \texttt{FGPR} can also adequately capture uncertainties (grey areas in Figure \ref{fig:MF1}) in predictions. The results also confirm our insights on automatic personalization in Sec. \ref{subsec:why} whereby a single global model was able to adequately fit both HF and LF datasets. Here we conduct one additional comparison study on Example 2. We train a $\mathcal{GP}$ model solely using high-fidelity dataset. The fitted curve is plotted in Figure \ref{fig:MF2}. It can be seen that, without borrowing any information from the LF dataset, the fitted curve is a polynomial line that with large noise. This example further demonstrates the advantage of \texttt{FGPR}: the shared global model parameter encodes key information (e.g., trend, pattern) from the low-fidelity dataset such that the high-fidelity dataset can exploit this information to fit a more accurate surrogate model.

% \begin{figure*}[!htbp]
%     \vskip -0.1in
%     \centering
%     \centerline{\includegraphics[width=0.7\columnwidth]{figs/MF_example_2.pdf}}
%     \caption{Example 2.}
%     \label{fig:MF2}
%     \vskip -0.1in
% \end{figure*}

% \begin{tabular}{@{}c@{}}RMSE-LF \\ RMSE-HF \end{tabular}

Next, we consider a range of benchmark problems that are widely used in the multi-fidelity literature \citep{cutajar2019deep, brevault2020overview}. We defer the full specifications of those problems to the Appendix. For each experiment, we generate 1,000 testing points uniformly on the input domain.

\begin{itemize}
    \item \textbf{CURRIN:} CURRIN \citep{currin1991bayesian, xiong2013sequential} is a two-dimensional problem that is widely used for multi-fidelity computer simulation models.
    \item \textbf{PARK:} The PARK function \citep{cox2001statistical, xiong2013sequential} lies in a four-dimensional space ($\bm{x}\in(0,1]^4$). This function is used as an illustrative example for parameter calibration and design of experiments.
    \item \textbf{BRANIN:} BRANIN is widely used as a test function for metamodeling in computer experiments. In this example, there are three fidelity levels \citep{perdikaris2017nonlinear, cutajar2019deep}. 
    \item \textbf{Hartmann-3D:} Similar to BRANIN, this is a 3-level multi-fidelity dataset where the input space is $[0,1]^3$.
    \item \textbf{Borehole Model:} The Borehole model is an 8-dimensional physical model that simulates water flow through a borehole \citep{moon2012two, gramacy2012gaussian, xiong2013sequential}. 
\end{itemize}

Each experiment is repeated 30 times and we report RMSEs of model performance on the true HF model, along with the standard deviations in Table \ref{table:RMSE}. 

\begin{table*}[!htbp]
\centering
\scriptsize
\caption{RMSEs and standard deviations compared to the true HF model. Each experiment is repeated 30 times. The sample size is in a format of HF/MF/LF, where MF represents a medium-fidelity model.}
\begin{tabular}{cccccc}
\hline
RMSE-HF & Sample Size & \textbf{\texttt{FGPR}} & \texttt{Separate} & \texttt{AR} & \texttt{Deep} \\ \hline
CURRIN & $40/0/200$ & $\bm{0.148\pm 0.056}$ & $0.301\pm 0.080$ & $0.295\pm0.052$ & $0.252\pm0.064$ \\ \hline
PARK & $50/0/300$ & $\bm{0.012\pm 0.002}$ & $0.052\pm0.006$ & $0.035\pm0.001$  & $0.013\pm0.001$   \\ \hline
BRANIN & $20/40/200$ & $0.260\pm0.065$ & $0.374\pm0.089$ & $0.335\pm0.070$ & $\bm{0.213\pm0.085}$ \\ \hline
Hartmann-3D & $50/100/200$ & $ \bm{0.365\pm0.074}$ & $0.456\pm0.087$ & $0.412\pm0.067$ & $0.383\pm0.092$ \\ \hline
Borehole & $50/0/200$ & $\bm{0.604\pm0.006}$ & $0.633\pm0.006$ & $0.615\pm0.005$ & $0.622\pm0.007$ \\ \hline
\end{tabular}
\label{table:RMSE}
\end{table*}

First, it can be seen in Table \ref{table:RMSE}, \texttt{FGPR} consistently yields smaller RMSE than \texttt{Separate}. This confirms that \texttt{FGPR} is able to borrow strength across multi-fidelity datasets. More importantly, we find that \texttt{FGPR} can even achieve superior performance compared to the \texttt{AR} and \texttt{Deep} benchmarks. This implies that one can avoid centralized approaches without compromising accuracy. In summary, the results show that \texttt{FGPR} can serve as a compelling candidate for privacy-preserving multi-fidelity modeling in the modern era of statistics and machine learning.

Below, we also detail an interesting technical observation.

\begin{remark}
In our settings, the weight coefficient $p_k$ for the HF is low compared to LF as HF clients have less data. For instance, in the CURRIN example, the HF coefficient is $p_1=\frac{40}{240}=0.17$. Therefore, the global parameter is averaged with higher weights for the LF model. Yet, the model excels in predicting the HF model. This again goes back to the fact that, unlike deep learning based FL approaches, \texttt{FGPR} is learning a joint prior on the functional space. The scarce HF data alone cannot learn a strong prior, yet, with the help of the LF data, such prior can be learned effectively. That being said, it may be interesting to investigate the adaptive assignment of $p_k$, yet this requires additional theoretical analysis.  
%Since we are focusing on the HF dataset, one might wonder why we assign smaller weight to it. The reason is that device with HF dataset does not have sufficient amounts of data to fit an accurate surrogate. Therefore, it needs to borrow more information from LF dataset (i.e., higher weight coefficient for the LF device). Based on all of our empirical studies, this assignment consistently provides inferior performance than \texttt{FGPR}. It is possibly true that we can consider some middle values between small $p_1$ and $1$ in the future work. However, it needs additional theoretical analysis. Despite this, the current approach to device $p_i$ has already provided satisfactory empirical performance. 
\end{remark}

\begin{figure*}[!htbp]
    \vskip -0.1in
    \centering
    \centerline{\includegraphics[width=0.5\columnwidth]{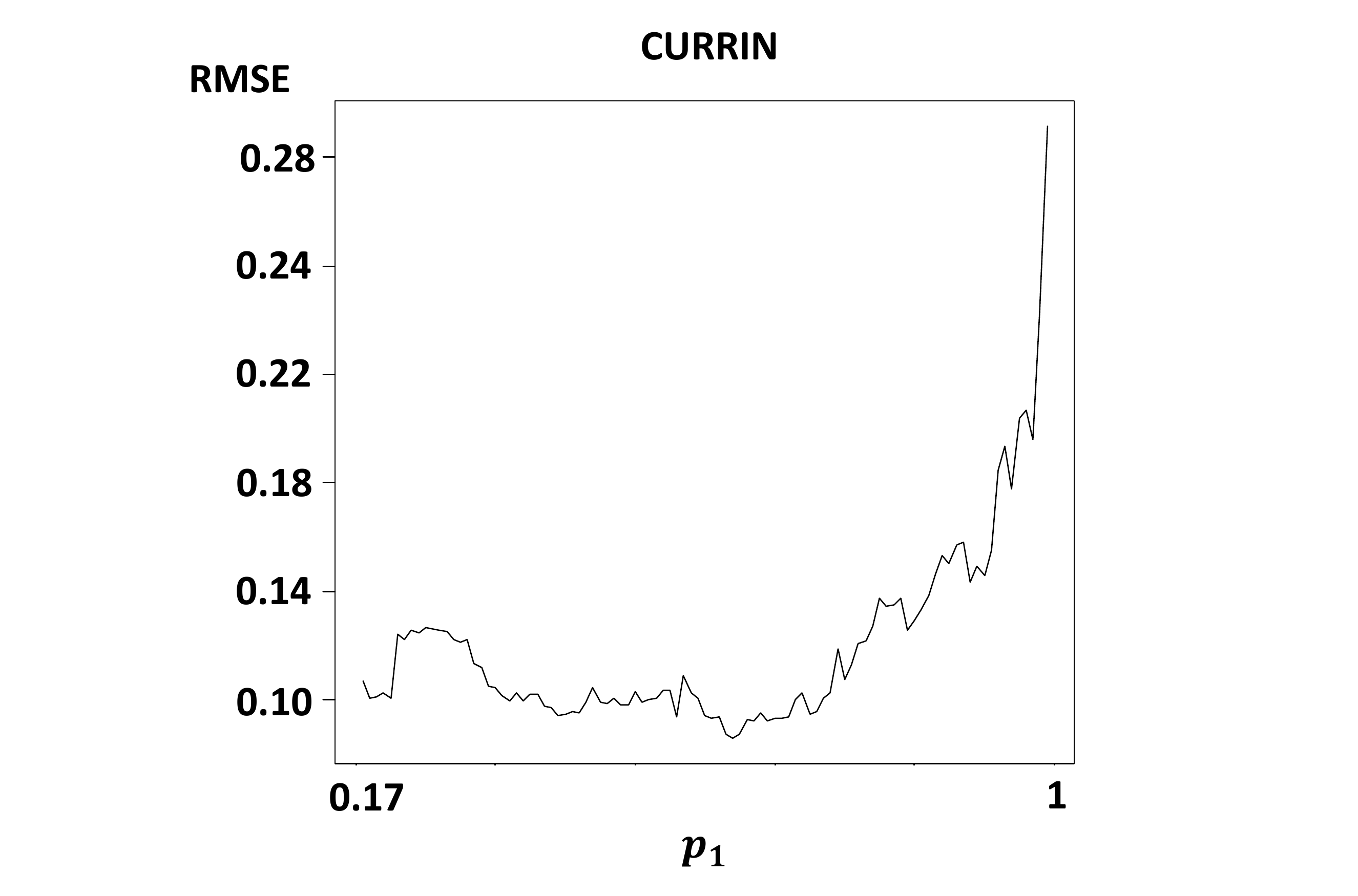}}
    \caption{Ablation Study (CURRIN)}
    \label{fig:MF-ablation}
    \vskip -0.1in
\end{figure*}

On par with Remark 17, we conduct an ablation study on $p_i$ using the CURRIN function. Specifically, we use the same sample size (i.e., $N_1=40$, $N_2=200$) but we gradually increase $p_1$ from $0.17$ to 1 and decrease $p_2$ from $0.83$ to $0$. We plot the RMSE versus $p_1$ in Figure \ref{fig:MF-ablation}. It can be seen that the RMSE remains consistent when we moderately increase $p_1$. However, once $p_1$ passes a threshold, the RMSE increases sharply. Again this is because the increased weight to HF can be misleading due to the scarcity of HF data.

\section{Application II: NASA Aircraft Gas Turbine Engines}
\label{sec:exp_nasa}

In our second case study, we consider degradation signals generated from aircraft gas turbine engines using the NASA Commercial Modular Aero-Propulsion System Simulation (C-MAPSS) tools (\hyperlink{https:
//ti.arc.nasa.gov/tech/dash/groups/pcoe/}{NASA dataset Link}). The dataset consists of 100 engines and contains time-series degradation signals collected from multiple sensors installed on the engines. The goal of the experiment is to predict the degradation signals for test engines in a federated paradigm. To do so, we assume that each client/device is a single engine and all engines are aiming to collaboratively learn a predictive degradation model. 
% We randomly sample 60 engines as training units and 40 engines as testing units. 

% At each communication round, we set $|\mathcal{S}_c|=10$.

We briefly describe our training procedures. We randomly divide the 100 engines into 60 training engines and 40 testing engines. For each testing unit $k$, we randomly split the data on each device into a 50\% training dataset $D_{k,\text{train}}\coloneqq(\bm{X}_{k,\text{train}},\bm{y}_{k,\text{train}})$ and a 50\% testing dataset $D_{k,\text{test}}\coloneqq(\bm{X}_k^*,\bm{y}_k^*)$, where $\bm{y}^*_k=\left[y^*_{k,1}, ...y^*_{k,|D_{k,\text{test}}|}\right]^\intercal$, $\bm{X}_k^*=\left[{x^*_{k,1}}^\intercal, ...,{x^*_{k,|D_{k,\text{test}}|}}^\intercal\right]$. Recall that in Sec. \ref{sec:algorithm}, we define $|D_{k,\text{test}}|$ as the number of data points in the set $D_{k,\text{test}}$. We first train \texttt{FGPR} using the 60 training units and obtain a final aggregated global model parameter $\btheta$. The testing unit $k$ then directly uses this global parameter  $\btheta$ and $D_{k,\text{train}}$ to predict outputs $[f({x^*_{k,1}}^\intercal),\cdots,f({x^*_{k,|D_{k,\text{test}}|}}^\intercal)]$ at testing locations $\bm{X}_k^*$ without any additional training. 

We benchmark \texttt{FGPR} with the following models. 
\begin{enumerate}
    \item \texttt{Polynomial}: All signal trajectories exhibit polynomial patterns and therefore a polynomial regression is often employed to analyze this dataset \citep{liu2013data, yan2016multiple,song2018statistical}. More specifically, we train a polynomial regression using \texttt{FedAvg}. During the training process, each device updates the  coefficients of a polynomial regression in the form of  $y_k(x)=\sum_{i=0}^p\beta_{ik}x^i+\epsilon_k(x)$, where $\{\beta_{ik}\}_{i=0}^p$ are model parameters. This update is done by running gradient descent to minimize the local sum squared error. The central server aggregates the parameters using \texttt{FedAvg} and broadcasts the aggregated parameter to all devices in the following communication round. Here, we conduct experiments with different $p\in\{1,\ldots,20\}$ and select the best $p$ with the smallest averaged testing RMSE (will be defined shortly). Our empirical study find that $p=10$ provides the best performance. 
    \item \texttt{Neural}: we train a $q$-layer neural network using \texttt{FedAvg} \citep{mcmahan2017communication}. Similar to \texttt{Polynomial}, we test the performance of the neural network with different $q\in\{1,\ldots,20\}$. The best value is $2$.
    % \item \texttt{Separate} $\mathcal{GP}$. A separate $\mathcal{GP}$ is trained on the individual local data of the 40 testing devices
\end{enumerate}

The performance of each model is measured by the averaged RMSE across all 40 testing devices defined as follows:
\begin{align*}
    \text{RMSE}=\frac{1}{40}\sum_{k=1}^{40}\sqrt{\frac{\sum_{i=1}^{|D_{k,\text{test}}|}(f({x^*_{k,i}}^\intercal)-y_{k,i})^2}{|D_{k,\text{test}}|}}.
\end{align*}

The averaged RMSE and the standard deviation of RMSE across all testing devices are reported in Table \ref{table:RMSE_NASA}. Each experiment is repeated 30 times. The outputs on each device are scaled to be a mean 0 and variance 1 sequence.

\begin{table}[!htbp]
\centering
\begin{tabular}{cccc}
\hline
\begin{tabular}{@{}c@{}}Averaged RMSE $\times 10$ \\ std of RMSE $\times 10$ \end{tabular}& \texttt{FGPR}  & \texttt{Polynomial} & \texttt{Neural} \\ \hline
% Sensor 2, Scenario (I) & \begin{tabular}{@{}c@{}}$\bm{xx}$\\$\bm{xxx}$\end{tabular}   & &  &\\ \hline
Sensor 2 & \begin{tabular}{@{}c@{}}$\bm{5.45\ (0.01)}$\\ $\bm{0.87\ (0.02)}$ \end{tabular} & \begin{tabular}{@{}c@{}}$6.79\ (0.01)$\\ $0.98\ (0.02)$ \end{tabular} & \begin{tabular}{@{}c@{}}$6.47\ (0.05)$\\ $1.02\ (0.01)$ \end{tabular} \\ \hline
% Sensor 7, Scenario (I) &     & &  &\\ \hline
Sensor 7 & \begin{tabular}{@{}c@{}}$\bm{5.76\ (0.03)}$\\ $\bm{0.76\ (0.01)}$ \end{tabular} & \begin{tabular}{@{}c@{}}$6.55\ (0.02)$\\ $0.89\ (0.04)$ \end{tabular} & \begin{tabular}{@{}c@{}}$6.71\ (0.02)$\\ $0.85\ (0.03)$ \end{tabular}\\ \hline
\end{tabular}
\caption{Averaged RMSE (line 1 in each cell) and standard deviation (std) of RMSE (line 2 in each cell) across all testing devices for the NASA data. Each experiment is repeated 30 times. }
% \textbf{The $\texttt{Separate}$ is trained on the 40 testing devices.}
\label{table:RMSE_NASA}
\end{table}

% From Table \ref{table:RMSE_NASA}, we can obtain some important insights. First, it can be seen that \texttt{FGPR} outperform other benchmark models. Notably, it also provides superior performance to the \texttt{Separate} approach in Scenario (I). The key reason is that the same type of sensors across all engines have very similar patterns (i.e., homogeneous patterns). Therefore, the federated method explores richer information across different devices than the separate modeling approach. Second, 

From Table \ref{table:RMSE_NASA}, we can obtain some important insights. First, \texttt{FGPR} consistently yields lower averaged RMSE than other benchmark models. This illustrates the good transferability of \texttt{FGPR}. More concretely, a shared global model can provide accurate surrogates even on untrained devices. This feature is in fact very helpful in transfer learning or online learning. For instance, the shared global model can be used as an initial parameter for fine-tuning on the streaming data. Second, \texttt{FGPR} also provides smaller standard deviation of RMSE across all devices. This credits to the automatic personalization feature encoded in $\mathcal{GP}$. In the next section, we will compare our model with a state-of-the-art personalized FL framework to further demonstrate the advantage of \texttt{FGPR}.

% Here, we also observe that the performances of \texttt{Polynomial} and \texttt{Neural} are unsatisfactory. This is because those benchmark models are lack of personalization features. 

% Next, we let all 100 engines participate in the training. In each device, the data are randomly split into a 60\% training set and a 40\% testing set. At each communication round, we select $10\%$ of units to be activated (i.e., $|\mathcal{S}_c|=10$). Results are presented in Table ???

\section{Application III: Robotics}
\label{sec:exp_robot}

We now test the performance of \texttt{FGPR} on a robotic dataset. 

To enable accurate movement of a robot, one needs to control the joint torques \citep{nguyen2008computed}. Joint torques can be computed by many existing inverse dynamics models. However, in real-world applications, the underlying physical process is highly complex and often hard to derive using first principles. Data-driven models were proposed as an appealing alternative to handle complex functional patterns and, more importantly, quantify uncertainties \citep{nguyen2011model}. The goal of this section to test  \texttt{FGPR} as a data-driven approach to  accurately compute joint torques at different joints positions, velocities and accelerations.

% The goal is to accurately compute joint torques at different joints positions, velocities and accelerations. This allows the design of more precise and compliant controls for robots. For instance, in a factory automation system, there is a planning server that assigns one or several tasks to robotic workers and orchestrates their movement. This server typically assigns a trajectory that takes robot from one place to another destination, by specifying the desired locations, velocities and accelerations. To ensure the movement is accurate, one need to control the joint torques \citep{nguyen2008computed}. Those joint torques can be computed by many existing inverse dynamics models. However, in the real-world application, the real physical process is typically more involved and renders those model inaccurate: the movement of robot might also affected by uncontrollable environmental factors or so. Therefore, regression model become an appealing alternative since it can handle complex functional patterns and, most importantly, quantity uncertainties \citep{nguyen2011model}. 

To this end, we test \texttt{FGPR} using a Mat\'ern-$3/2$ kernel on learning an inverse dynamics problem for a seven degrees-of-freedom SARCOS anthropomorphic robot arm \citep{williams2006gaussian, bui2018partitioned}. This task contains $d=21$ dimensional input and $7$ dimensional output with 44,484 points for training and 4,449 points for testing. Since \texttt{FGPR} is a single-output FL framework, we only use one output each time (See Table \ref{table:RMSE_robot}). Our goal is to accurately predict the forces used at different joints given the joints' input information. We randomly partition the data into 25 devices. Overall, each device has around 1850 training points and 180 testing points each. We found that the data pattern in each device is high dimensional, highly non-linear and heterogeneous. Therefore, the neural network trained from a simple \texttt{FedAvg} failed. To resolve this issue, we train the \texttt{Neural} using a state-of-the-art personalized FL framework \texttt{Ditto} \citep{li2021ditto}. In \texttt{Ditto}, each local device solve two optimization problems. The first is the same as \texttt{FedAvg} and to find $\bm{\theta}$, while the second derives personalized parameters $\bm{v}_k$ for each client $k$ by solving
\begin{align*}
    \min_{\bm{v}_k}h_k(\bm{v}_k;\btheta)\coloneqq F_k(\bm{v}_k)+\frac{\lambda}{2}\norm{\bm{v}_k-\btheta}_2^2
\end{align*}
where $F_k$ is the local loss function, $\lambda$ is a regularization parameter and $\btheta$ is the shared global parameter. The idea behind \texttt{Ditto} is clear: in addition to updating a shared global parameter $\btheta$, each device also maintains its own personalized solution $\bm{v}_k$. Yet, the regularization term ensures that this $\bm{v}_k$ should be close to $\btheta$ such that one can retain useful information learned from a global model. 

In Table \ref{table:RMSE_robot}, we present results from Output 1, 3, 5 and 7. 
Under the heterogeneous setting, \texttt{FGPR} alone can still provide lower averaged RMSE than the personalized \texttt{Neural} benchmark model. This credits to (1) the flexible prior regularization in the $\mathcal{GP}$ regression that can avoid potential model over-fitting; and most importantly, (2) the intrinsic personalization capability of \texttt{FGPR}.

% \begin{table}[!htbp]
% \centering
% \begin{tabular}{cccc}
% \hline
% \begin{tabular}{@{}c@{}}Averaged RMSE $\times 10$ \\ std of RMSE $\times 10$\end{tabular}& \texttt{FGPR} & \texttt{Separate} & \texttt{Neural}\\ \hline
% Output 1 & \begin{tabular}{@{}c@{}}$\bm{2.75\ (0.00)}$ \\ $\bm{1.84\ (0.01)}$ \end{tabular} &    \begin{tabular}{@{}c@{}}$3.22\ (0.02)$ \\ $2.44\ (0.00)$ \end{tabular} & \begin{tabular}{@{}c@{}}$3.01\ (0.01)$ \\ $1.70\ (0.00)$ \end{tabular} \\ \hline
% Output 3 & \begin{tabular}{@{}c@{}}$\bm{2.42\ (0.03)}$ \\ $\bm{1.57\ (0.01)}$ \end{tabular} & \begin{tabular}{@{}c@{}}$2.93\ (0.01)$ \\ $1.75\ (0.03)$ \end{tabular} & \begin{tabular}{@{}c@{}}$3.05\ (0.06)$ \\ $2.11\ (0.02)$ \end{tabular} \\ \hline
% Output 5 & \begin{tabular}{@{}c@{}}$2.20\ (0.05)$ \\ $1.29\ (0.02)$ \end{tabular} & \begin{tabular}{@{}c@{}}$\bm{2.11\ (0.03)}$ \\ $\bm{1.33\ (0.01)}$ \end{tabular} & \begin{tabular}{@{}c@{}}$2.89\ (0.09)$ \\ $1.37\ (0.02)$ \end{tabular}\\ \hline
% Output 7 & \begin{tabular}{@{}c@{}}$\bm{2.38\ (0.01)}$ \\ $\bm{1.44\ (0.02)}$ \end{tabular} & \begin{tabular}{@{}c@{}}$2.56\ (0.03)$ \\ $1.39\ (0.01)$ \end{tabular} & \begin{tabular}{@{}c@{}}$2.90\ (0.02)$ \\ $1.50\ (0.01)$ \end{tabular}\\ \hline
% \end{tabular}
% \caption{Averaged RMSE (line 1 in each cell) and standard deviation (std) of RMSE (line 2 in each cell) across all testing devices for the robotics data. Each experiment is repeated 30 times. We provide the standard deviation of each performance measure in the bracket.}
% \label{table:RMSE_robot}
% \end{table}

\begin{table*}[!htbp]
\centering
\scriptsize
\caption{Averaged RMSE (line 1 in each cell) and standard deviation (std) of RMSE (line 2 in each cell) across all testing devices for the robotics data. Each experiment is repeated 30 times. We provide the standard deviation of each performance measure in the bracket.}
\begin{tabular}{cccccc}
\hline
\begin{tabular}{@{}c@{}}Averaged RMSE $\times 10$ \\ std of RMSE $\times 10$\end{tabular}& Output 1 &  Output 3 & Output 5 & Output 7\\ \hline
\texttt{FGPR} & \begin{tabular}{@{}c@{}}$\bm{2.75\ (0.00)}$ \\ $\bm{1.84\ (0.01)}$ \end{tabular} & \begin{tabular}{@{}c@{}}$\bm{2.42\ (0.03)}$ \\ $\bm{1.57\ (0.01)}$ \end{tabular} &  \begin{tabular}{@{}c@{}}$\bm{2.20\ (0.05)}$ \\ $\bm{1.29\ (0.02)}$ \end{tabular} & \begin{tabular}{@{}c@{}}$\bm{2.38\ (0.01)}$ \\ $\bm{1.44\ (0.02)}$ \end{tabular} \\ \hline
\texttt{Neural} & \begin{tabular}{@{}c@{}}$3.01\ (0.01)$ \\ $1.70\ (0.00)$ \end{tabular}  & \begin{tabular}{@{}c@{}}$3.05\ (0.06)$ \\ $2.11\ (0.02)$ \end{tabular} & \begin{tabular}{@{}c@{}}$2.89\ (0.09)$ \\ $1.37\ (0.02)$ \end{tabular} & \begin{tabular}{@{}c@{}}$2.90\ (0.02)$ \\ $1.50\ (0.01)$ \end{tabular}\\ \hline
\end{tabular}
\label{table:RMSE_robot}
\end{table*} 

% Under the heterogeneous setting, \texttt{FGPR} alone can still provide lower averaged RMSE than the personalized \texttt{Neural} benchmark model. This credits to (1) the flexible prior regularization in the $\mathcal{GP}$ regression that can avoid potential model over-fitting; and most importantly, (2) the native personalized predictive equation lies in the $\mathcal{GP}$ regression model. Here we note that compared to the \texttt{Separate} approach, the \texttt{FGPR} is sometimes marginal or even inferior. This is intuitively understandable as the sample size in each local device is large (around $1,850$) such that a local  $\mathcal{GP}$ regression model is capable of providing an accurate surrogate without communication.

\section{Related Work}
\label{sec:related}

Most of the existing FL literature has focused on developing deep learning algorithms and their applications in image classification and natural language processing. Please refer to \citep{kontar2021internet} for an in-depth review of FL literature. Here we briefly review some related papers that tackle data heterogeneity. One popular trend \citep{li2018federated, zhang2020fedpd, pathak2020fedsplit} uses regularization techniques to allay heterogeneity. For instance, \texttt{FedProx} \citep{li2018federated} adds a quadratic regularizer to the client objective to limit the impact of heterogeneity by penalizing local updates that move far from the global model. Alternatively, personalized models were proposed. Such models usually follow an alternating train-then-personalize approach where a global model is learned, and the personalized model is regularized to stay within its vicinity \citep{kirkpatrick2017overcoming,dinh2020personalized, li2021ditto}. Other approaches \citep{arivazhagan2019federated, liang2020think} use different layers of a network to represent global and personalized solutions. More recently, researchers have tried to remove the dependence on a global model for personalization by following a multi-task learning philosophy \citep{smith2017federated}. Yet, such models can only handle simple linear formulations.  

To date, little to no literature exists beyond deep learning. Perhaps, the closest field where alternative regression approaches were investigated is distributed learning for distributed systems. Distributed approaches for MCMC, PCA, logistic and quantile regression have been proposed \citep{zhang2013communication, wang2013parallelizing, lee2017communication,lin2017distributed, chen2019quantile, chen2021distributed, chen2021first, fan2021communication}. However, distributed learning and FL have several fundamental differences. In distributed systems, clients are compute nodes within a centralized regime. Hence clients can shuffle, randomize and randomly partition the data. As such distributed learning approaches, most assume data are homogeneous \citep{jordan2018communication}. Some few exceptions yet exists \citep{duan2020heterogeneity}. In contrast, in FL, data partitions are fixed and only reside at the edge. Besides, distributed systems are connected by large bandwidth infrastructure, allowing models to aggregate after each optimization iterate. On the other hand, devices in FL usually have limited bandwidth and unreliable internet connection. Those difficulties cause infrequent communication and partial device participation \citep{bonawitz2019towards, li2020federated}.

\section{Conclusion}
\label{sec:con}

In this paper, we extend the standard $\mathcal{GP}$ regression model to a federated setting, \texttt{FGPR}. We use both theory and a wide range of experiments to justify the viability of our proposed framework. We highlight the unique capability of \texttt{FGPR} to provide automatic personalization and strong transferability on untrained devices. 

\texttt{FGPR} may find value in meta-learning as it provides an inherent Bayesian perspective on this topic. Other interesting research directions include extending the current framework to a multi-output $\mathcal{GP}$ model. The challenge lies in capturing the correlation across output in federated paradigm.  Another possible direction arises from the theoretical perspective of $\texttt{FGPR}$. In this work, we only provide theoretical guarantees on noise/variance parameters and the gradient norm. Studying the convergence behavior of length parameters is another crucial but challenging future research direction.

\newpage

\section{Appendix}

\section{Multi-fidelity Modeling}

\textbf{Example 3: CURRIN} The CURRIN \citep{currin1991bayesian, xiong2013sequential} is a two-dimensional problem that is widely used for multi-fidelity computer simulation models. Given the input domain $\bm{x}\in[0,1]^2$, the high-fidelity model is 
\begin{align*}
    y_h(\bm{x})=\left[1-\exp\left(-\frac{1}{2x_2}\right)\right]\frac{2300x_1^3+1900x_1^2+2092x_1+60}{100x_1^3+500x_1^2+4x_1+20}
\end{align*}
whereas the low-fidelity model is given by
\begin{align*}
    y_l(\bm{x})&=\frac{1}{4}[y_h(x_1+0.05, x_2+0.05)+y_h(x_1+0.05,\max(0,x_2-0.05))]\\
    &\qquad +\frac{1}{4}[y_h(x_1-0.05, x_2+0.05)+y_h(x_1-0.05,\max(0,x_2-0.05))].
\end{align*}
We collect 40 data points from the HF model and 200 data points from the LF model. The number of testing data points is 1,000. 

\textbf{Example 4: PARK} The PARK function \citep{cox2001statistical, xiong2013sequential} is a four-dimensional problem ($\bm{x}\in(0,1]^4$) where the high-fidelity model is given as
\begin{align*}
    y_h(\bm{x})=\frac{x_1}{2}\left[\sqrt{1+(x_2+x_3^2)\frac{x_4}{x_1^2}}-1\right] + (x_1+3x_4)\exp[1+\sin(x_3)],
\end{align*}
while the low-fidelity model is
\begin{align*}
    y_l(\bm{x})=\left[1+\frac{\sin(x_1)}{10}\right]y_h(\bm{x}) - 2x_1 + x_2^2 + x_3^2 + 0.5.
\end{align*}

\textbf{Example 5: BRANIN} In this example, there are three fidelity levels \citep{perdikaris2017nonlinear, cutajar2019deep}:
\begin{align*}
    y_h&=\left(\frac{-1.275x_1^2}{\pi^2}+\frac{5x_1}{\pi}+x_2-6\right)^2 + \left(10-\frac{5}{4\pi}\right)\cos(x_1)+10,\\
    y_m&=10\sqrt{y_h(\bm{x}-2)} + 2(x_1-0.5) -3(3x_2-1)-1,\\
    y_l&=y_m(1.2(\bm{x}+2))-3x_2+1,\\
    x&\in[-5,10]\times[0,15]
\end{align*}
where $y_m(\cdot)$ represents the output from the medium-fidelity (MF) model. 

\textbf{Example 6: Hartmann-3D} Similar to Example 5, this is a 3-level multi-fidelity dataset where the input space is $[0,1]^3$. The evaluation of observations with fidelity $t$ is defined as \citep{cutajar2019deep}
\begin{align*}
    y_t(\bm{x})=\sum_{i=1}^4\alpha_i\exp\left(-\sum_{j=1}^3A_{ij}(x_j-P_{ij})^2\right)
\end{align*}
where 
\begin{align*}
    A&=\begin{bmatrix}3&10&30\\0.1&10&35\\3&10&30\\0.1&10&35\end{bmatrix}, P = \begin{bmatrix}0.3689&0.1170&0.2673\\0.4699 &0.4387 &0.7470\\0.1091 &0.8732& 0.5547\\0.0381& 0.5743 &0.8828\end{bmatrix},\\
    \bm{\alpha}&=(1.0, 1.2, 3.0, 3.2)^\intercal,\bm{\alpha}_t=\bm{\alpha}+(3-t)\bm{\delta},\bm{\delta}=(0.01, -0.01, -0.1, 0.1)^\intercal.
\end{align*}

\textbf{Example 7: Borehole Model} The Borehole model is an 8-dimensional physical model that simulates water flow through a borehole \citep{moon2012two, gramacy2012gaussian, xiong2013sequential}. The high-fidelity model is given as
\begin{align*}
    y_h(\bm{x}) = \frac{2\pi x_3(x_4-x_6)}{\ln(x_2/x_1)[1+2x_7x_3/(\ln(x_2/x_1)x_1^2x_8)+x_3/x_5]}
\end{align*}
where $x_1\in[0.05, 0.15], x_2\in[100, 50000], x_3\in[63070, 115600], x_4\in[990, 1110], x_5\in[63.1, 115], x_6\in[700, 820], x_7\in[1120, 1680], x_8\in[9855, 12045]$. The low-fidelity model is
\begin{align*}
    y_l(\bm{x}) = \frac{5\pi x_3(x_4-x_6)}{\ln(x_2/x_1)[1.5+2x_7x_3/(\ln(x_2/x_1)x_1^2x_8)+x_3/x_5]}.
\end{align*}

\section{Important Lemmas}
\label{app:lemma}

In this section, we present some key lemmas used in our theoretical analysis. We defer the proofs of those Lemmas into Section \ref{app:lemma_proof}.

\begin{lemma}
\label{lemma:1}
(Theorem 4 in \citet{braun2006accurate}) Let $\Ker$ be a Mercer kernel on a probability space $\mathcal{X}$ with probability measure $\mu$, satisfying $\Ker(x,x)\leq 1$ for all $x\in\mathcal{X}$, with eigenvalues $\{\lambda_i^*\}_{i=1}^\infty$. Let $\bm{K}_{f,N}$ be the empirical kernel matrix evaluated on data $\bm{X}$ $i.i.d.$ sampled from $\mu$, then with probability at least $1-\delta$, the eigenvalues of $\lambda_j(\bm{K}_{f,N})$ satisfies the following bound for $1\leq j\leq N$ and $1\leq r\leq N$:
\begin{align*}
    \left|\frac{\lambda_j(\bm{K}_{f,N})}{N}-\lambda_j^* \right|\leq\lambda_j^*C(r,N)+H(r,N),
\end{align*}
where 
\begin{align*}
    C(r,N)&<r\sqrt{\frac{2}{N\lambda_r^*}\log\frac{2r(r+1)}{\delta}} + \frac{4r}{3N\lambda_r^*}\log\frac{2r(r+1)}{\delta},\\
    H(r,N)&<\lambda_r^*+\sum_{i=r+1}^\infty\lambda_i^*+\sqrt{\frac{2\sum_{i=r+1}^\infty\lambda_i^*}{N}\log\frac{2}{\delta}} + \frac{2}{3N}\log\frac{2}{\delta}.
\end{align*}
Alternatively, $C(r,N)$ and $H(r,N)$ can also be bounded as follows:
\begin{align*}
    C(r,N)&<r\sqrt{\frac{r(r+1)}{N\delta \lambda_r^*}},\\
    H(R,N)&<\lambda_r^*+\sum_{i=r+1}^\infty\lambda_i^*+\sqrt{\frac{2\sum_{i=r+1}^\infty\lambda_i^*}{N\delta}}.
\end{align*}
\end{lemma}

This Lemma is proved in \citet{braun2006accurate}.

\begin{lemma}
\label{lemma:2}
\citep{chen2020stochastic} Under Assumptions \ref{assumption:parameter}-\ref{assumption:smooth}, in device $k$, for any $0<\epsilon_k,\alpha_k<1$, $C_{1k}(\alpha_k,b_k)>0$ and $N_k>C_{2k}(\epsilon_k,b_k)$, then with probability at least $1-\frac{2}{N_k^{\alpha_k}}$, we have
\begin{align*}
    \frac{\epsilon_k\log N_k}{8b_k\theta_{max}^2}\leq\sum_{j=1}^{N_k}\frac{\lambda_{1j}^2}{\left(\theta_{1k}^{(t)}\lambda_{1j}+\theta_{2k}^{(t)}\lambda_{2j}\right)^2}&\leq \frac{4+2\alpha_k}{b_k\theta_{min}^2}\log N_k\\
    \frac{N_k-C_{1k}(\alpha_k,b_k)\log N_k}{4\theta_{max}^2}\leq\sum_{j=1}^{N_k}\frac{\lambda_{2j}^2}{\left(\theta_{1k}^{(t)}\lambda_{1j}+\theta_{2k}^{(t)}\lambda_{2j}\right)^2}&\leq \frac{N_k}{\theta_{min}^2}\\
    \sum_{j=1}^{N_k}\frac{\lambda_{1j}\lambda_{2j}}{\left(\theta_{1k}^{(t)}\lambda_{1j}+\theta_{2k}^{(t)}\lambda_{2j}\right)^2}&\leq \frac{5+2\alpha_k}{7b_k\theta_{min}^2}\log N_k.
\end{align*}
\end{lemma}
This Lemma is proved in \citet{chen2020stochastic}. Here note that we omit the subscript $k$ in the eigenvalues $\lambda$ for simplicity. The full notation should be, for example, $\lambda_{1jk}^2$ for device $k$.

\begin{lemma}
\label{lemma:2-2}
Under Assumption \ref{assumption:parameter}-\ref{assumption:norm} and \ref{assumption:non-smooth}, for any $0<\alpha_k<\frac{8b_k^2-12b_k-6}{4b_k+3}$, with probability at least $1-\frac{1}{N_k^{1+\alpha_k}}$, the following inequalities hold:
\begin{align*}
    \sum_{j=1}^{N_k}\frac{\lambda_{1j}^2}{\left(\theta_{1k}^{(t)}\lambda_{1j}+\theta_{2k}^{(t)}\lambda_{2j}\right)^2}&\leq N_k^{\frac{(2+\alpha_k)(4b_k+3)}{4b_k(2b_k-1)}}\left(\frac{1}{\theta_{min}^2}+\frac{C^2_{mat,k}(4b_k+3)}{\theta_{min}^2(8b_k^2-8b_k-3)}\right),\\
    \sum_{j=1}^{N_k}\frac{\lambda_{1j}\lambda_{2j}}{\left(\theta_{1k}^{(t)}\lambda_{1j}+\theta_{2k}^{(t)}\lambda_{2j}\right)^2}&\leq N_k^{\frac{(2+\alpha_k)(4b_k+3)}{4b_k(2b_k-1)}}\left(\frac{1}{\theta_{min}^2}+\frac{C_{mat,k}(4b_k+3)}{\theta_{min}^2(4b_k^2-6b_k-3)}\right),\\
      \frac{N_k-C_{mat,k}N_k^{\frac{(2+\alpha_k)(4b_k+3)}{4b_k(2b_k-1)}}}{4\theta_{max}}\leq\sum_{j=1}^{N_k}\frac{\lambda_{2j}^2}{\left(\theta_{1k}^{(t)}\lambda_{1j}+\theta_{2k}^{(t)}\lambda_{2j}\right)^2}&\leq\frac{N_k}{\theta_{min}^2}.
\end{align*}
\end{lemma}
Lemma \ref{lemma:2-2} provides several bounds to constrain the eigenvalues of Mat\'ern kernel.

\begin{lemma}
\label{lemma:3} 
Under Assumption \ref{assumption:parameter}-\ref{assumption:smooth}, with probability at least $1-2TM^{-c}$, the following inequality holds for any $k\in[K]$ and $0\leq t < T$:
\begin{align*}
    \langle \btheta_k^{(t)}-\btheta^*, g_k^*(\btheta_k^{(t)}) \rangle \geq \frac{\gamma_k}{2}\norm{\btheta_k^{(t)}-\btheta_k^*}_2^2 - C_{3k}(\alpha_k,b_k)\frac{\log M_k}{M_k},
\end{align*}
where $\gamma_k=\min\left\{\frac{1}{32\tau b_k\theta_{max}^2},\frac{1}{4\theta_{max}^2}-\frac{8\theta_{max}^2}{\tau b_k\theta_{min}^4}\right\}$  and $C_{3k}(\alpha_k,b_k)=\frac{1}{64b_k}+\frac{C_{1k}(\alpha_k,b_k)}{8}- \frac{4\theta_{max}^2}{b\theta_{min}^2}$.
\end{lemma}

\begin{lemma}
\label{lemma:3-2} 
Under Assumption \ref{assumption:parameter}-\ref{assumption:norm} and \ref{assumption:non-smooth}, with probability at least $1-\frac{1}{M_k^{1+\alpha_k}}$, the following inequality holds:
\begin{align*}
    \left[g_k^*(\btheta_k^{(t)})\right]_2(\theta_{2k}^{(t)}-\theta_{2k}^*)\geq\frac{\gamma_k}{2}(\theta_{2k}^{(t)}-\theta_{2k}^*)^2-(\theta_{max}-\theta_{min})^2M_k^{\frac{(2+\alpha_k)(4b_k+3)}{4b_k(2b_k-1)}-1}\left(\frac{1}{2\theta_{min}^2}+\frac{C_{mat,k}(4b_k+3)}{2\theta_{min}^2(4b_k^2-6b_k-3)}\right),
\end{align*}
where $\gamma_k\coloneqq\frac{1}{2M_k}\frac{M_k-C_{mat,k}M_k^{\frac{(2+\alpha_k)(4b_k+3)}{4b_k(2b_k-1)}}}{4\theta_{max}}$. 
\end{lemma}

\begin{lemma}
\label{lemma:4}
\citep{chen2020stochastic}
Under Assumption \ref{assumption:parameter}-\ref{assumption:norm}, for any $\phi>0$, we have
\begin{align*}
    P\left(\sup_{\bm{\theta}}\frac{N}{s_i(N)}\left|[\nabla L(\btheta)]_i-[\nabla L^*(\btheta)]_i\right|>C_{\btheta}\phi\right)\leq \delta(\phi), i = 1,2.
\end{align*}
Furthermore, if assumption \ref{assumption:smooth} holds and $s_i(N)=\tau\log N$, then for $N>C_{\bm{\theta}},c_{\bm{\theta}}>0$, we have
\begin{align*}
    \delta(\phi)\leq\frac{C_{\btheta}}{N^{c_{\btheta}}}+C_{\btheta}(\log\phi)^4\exp\{-c_{\btheta}\log N \min\{\phi^2,\phi\}\}.
\end{align*}
If assumption \ref{assumption:smooth} or \ref{assumption:non-smooth} holds and $s_i(N)=N$, then
\begin{align*}
    \delta(\phi)\leq C_{\btheta}(\log\phi)^4\exp\{-c_{\btheta} N \min\{\phi^2,\phi\}\}.
\end{align*}
\end{lemma}

This Lemma is provided in \citet{chen2020stochastic}.

\section{Proof of Theorems}
\label{app:theorem_proof}

\subsection{Detailed Notations}
% In this proof, we borrow some key notations from a conventional theoretical FL paper written by \cite{li2019convergence}. 

Let $\btheta_k^{(t)}$ be the model parameter maintained in the $k^{th}$ device at the $t^{th}$ step. Let $\mathcal{I}_E=\{cE \ | \ c=1,2,\ldots,C\}$ be the set of global aggregation steps. If $t+1\in\mathcal{I}_E$, then the central server collects model parameters from active devices and aggregates all of those model parameters. Motivated by \cite{li2019convergence}, we introduce an intermediate parameter $\bm{v}_k^{(t+1)}\coloneqq \bm{\theta}^{(t)}_k - \eta^{(t)}g_k(\bm{\theta}^{(t)}_k;\xi^{(t)}_k)$. It can be seen that $\btheta^{(t+1)}_k=\bm{v}_k^{(t+1)}$ if $t+1\notin\mathcal{I}_E$ and  $\btheta^{(t+1)}_k=\sum_{k=1}^Kp_k\bm{v}_k^{(t+1)}$ otherwise. Let $\bm{\bar{v}}^{(t)}=\sum_{k=1}^Kp_k\bm{v}_k^{(t)}$ and $\bbtheta^{(t)}=\sum_{k=1}^Kp_k\btheta_k^{(t)}$. The central server can only obtain $\bbtheta^{(t)}$ when $t+1\in\mathcal{I}_E$. The term $\bm{\bar{v}}^{(t)}$ is introduced for the purpose of proof and is inaccessible in practice. We further define $g^{(t)}=\sum_{k=1}^Kp_kg_k(\btheta_k^{(t)};\xi_k^{(t)})$. 
% and $\bar g^{(t)}=\mathbb{E}g^{(t)}$.

%~~~~~~~~~~~~~~~~~~~~~~~~~~~~~~~~~~~~~~~~~~~%
%~~~~~~~~~~~~~~~~~~~~~~~~~~~~~~~~~~~~~~~~~~~%
%~~~~~~~~~~~~~~~~~~~~~~~~~~~~~~~~~~~~~~~~~~~%
%~~~~~~~~~~~~~~~~~~~~~~~~~~~~~~~~~~~~~~~~~~~%
%~~~~~~~~~~~~~~~~~~~~~~~~~~~~~~~~~~~~~~~~~~~%
%~~~~~~~~~~~~~~~~~~~~~~~~~~~~~~~~~~~~~~~~~~~%
%~~~~~~~~~~~~~~~~~~~~~~~~~~~~~~~~~~~~~~~~~~~%
%~~~~~~~~~~~~~~~~~~~~~~~~~~~~~~~~~~~~~~~~~~~%

\subsection{Proof of Theorem \ref{theorem:parameter_full_exp}}

Under the scenario of full device participation, we have $\bbtheta^{(t+1)}=\bar{\bm{v}}^{(t+1)}$ for all $t$. By definition of $\bar{\bm{v}}^{(t)}$, we have
\begin{align*}
    \norm{\bar{\bm{v}}^{(t+1)}-\btheta^*}_2^2&=\norm{\bbtheta^{(t)}-\eta^{(t)}g^{(t)}-\btheta^*}_2^2\\
    &=\underbrace{\norm{\bbtheta^{(t)}-\btheta^*}_2^2}_{\text{A}} \underbrace{- 2\eta^{(t)}\langle\bbtheta^{(t)}-\btheta^*,g^{(t)}\rangle}_{\text{B}} +\eta^{(t)2}\underbrace{\norm{g^{(t)}}_2^2}_{\text{C}}.
\end{align*}

We can write B as
\begin{align*}
    \text{B}&=-2\eta^{(t)}\langle\bbtheta^{(t)}-\btheta^*,g^{(t)}\rangle=-2\eta^{(t)}\langle\bbtheta^{(t)}-\btheta^*,\sum_{k=1}^Kp_kg_k(\btheta_k^{(t)};\xi_k^{(t)})\rangle\\
    &=-2\eta^{(t)}\sum_{k=1}^Kp_k\langle\bbtheta^{(t)}-\btheta^*,g_k(\btheta_k^{(t)};\xi_k^{(t)})\rangle\\
    &=-2\eta^{(t)}\sum_{k=1}^Kp_k\langle\bbtheta^{(t)}-\btheta^{(t)}_k,g_k(\btheta_k^{(t)};\xi_k^{(t)})\rangle-2\eta^{(t)}\sum_{k=1}^Kp_k\langle\btheta^{(t)}_k-\btheta^*,g_k(\btheta_k^{(t)};\xi_k^{(t)})\rangle.
\end{align*}

By Cauchy-Schwarz inequality and inequality of arithmetic and geometric means, we can simplify the first term in B as
\begin{align*}
    -2\langle\bbtheta^{(t)}-\btheta^{(t)}_k,g_k(\btheta_k^{(t)};\xi_k^{(t)})\rangle&\leq2\frac{\sqrt{\eta^{(t)}}}{\sqrt{\eta^{(t)}}}\norm{\bbtheta^{(t)}-\btheta^{(t)}_k}\norm{g_k(\btheta_k^{(t)};\xi_k^{(t)})}\\
    &\leq 2\frac{\frac{1}{\eta^{(t)}}\norm{\bbtheta^{(t)}-\btheta^{(t)}_k}^2+\eta^{(t)}\norm{g_k(\btheta_k^{(t)};\xi_k^{(t)})}^2}{2}\\
    &\leq\left(\frac{1}{\eta^{(t)}}\norm{\bbtheta^{(t)}-\btheta^{(t)}_k}_2^2 + \eta^{(t)}\norm{g_k(\btheta_k^{(t)};\xi_k^{(t)})}_2^2\right).
\end{align*}
By Lemma \ref{lemma:3}, we can simplify the second term in B as
\begin{align*}
    &-2\eta^{(t)}\langle\btheta^{(t)}_k-\btheta^*,g_k(\btheta_k^{(t)};\xi_k^{(t)})\rangle=-2\eta^{(t)}\langle\btheta^{(t)}_k-\btheta^*,g_k(\btheta_k^{(t)};\xi_k^{(t)})+g_k^*(\btheta_k^{(t)})-g_k^*(\btheta_k^{(t)})\rangle\\
    &\leq-2\eta^{(t)}\frac{\gamma_k}{2}\norm{\btheta^{(t)}_k-\btheta^*}_2^2+2\eta^{(t)}C_{3k}(\alpha_k,b_k)\frac{\log M_k}{M_k}-2\eta^{(t)}\langle\btheta^{(t)}_k-\btheta^*,g_k(\btheta_k^{(t)};\xi_k^{(t)})-g_k^*(\btheta_k^{(t)})\rangle.
\end{align*}

By Assumption \ref{assumption:norm},
\begin{align*}
    \text{C}=\norm{g^{(t)}}_2^2=\norm{\sum_{k=1}^Kp_kg_k(\btheta_k^{(t)};\xi_k^{(t)})}^2\leq \left(\sum_{k=1}^K\norm{p_kg_k(\btheta_k^{(t)};\xi_k^{(t)})}\right)^2\leq \left(\sum_{k=1}^Kp_kG\right)^2=G^2.
\end{align*}
Combining A, B and C together, we obtain
\begin{align*}
    &\norm{\bar{\bm{v}}^{(t+1)}-\btheta^*}_2^2\\
    &\leq\norm{\bbtheta^{(t)}-\btheta^*}_2^2 + \eta^{(t)}\sum_{k=1}^Kp_k\left(\frac{1}{\eta^{(t)}}\norm{\bbtheta^{(t)}-\btheta^{(t)}_k}_2^2 + \eta^{(t)}G^2\right) \\
    &\qquad -2\eta^{(t)}\sum_{k=1}^Kp_k\frac{\gamma_k}{2}\norm{\btheta^{(t)}_k-\btheta^*}_2^2+2\eta^{(t)}\sum_{k=1}^Kp_kC_{3k}(\alpha_k,b_k)\frac{\log M_k}{M_k} + \eta^{(t)2}G^2\\
    &\qquad -2\eta^{(t)}\sum_{k=1}^Kp_k\langle\btheta^{(t)}_k-\btheta^*,g_k(\btheta_k^{(t)};\xi_k^{(t)})-g_k^*(\btheta_k^{(t)})\rangle\\
    &=\norm{\bbtheta^{(t)}-\btheta^*}_2^2 + \underbrace{\sum_{k=1}^Kp_k\norm{\bbtheta^{(t)}-\btheta^{(t)}_k}_2^2}_{\text{D}} + \eta^{(t)2}G^2 \\
    &\qquad -2\eta^{(t)}\underbrace{\sum_{k=1}^Kp_k\frac{\gamma_k}{2}\norm{\btheta^{(t)}_k-\btheta^*}_2^2}_{\text{E}}+2\eta^{(t)}\sum_{k=1}^Kp_kC_{3k}(\alpha_k,b_k)\frac{\log M_k}{M_k} + \eta^{(t)2}G^2\\
    &\qquad -2\eta^{(t)}\sum_{k=1}^Kp_k\langle\btheta^{(t)}_k-\btheta^*,g_k(\btheta_k^{(t)};\xi_k^{(t)})-g_k^*(\btheta_k^{(t)})\rangle.
\end{align*}
Since the aggregation step happens each $E$ steps, for any $t\geq 0$, there exists a $t_0\leq t$ such that $t-t_0\leq E-1$ and $\btheta_k^{(t_0)}=\bbtheta^{(t_0)}$ for all $k\in[K]$. Since $\eta^{(t)}$ is non-increasing, for all $t-t_0\leq E-1$, we can simplify D as 
\begin{align*}
    \text{D}&=\sum_{k=1}^Kp_k\norm{\bbtheta^{(t)}-\btheta^{(t)}_k}_2^2=\sum_{k=1}^Kp_k\norm{(\btheta^{(t)}_k-\bbtheta^{(t_0)})-(\bbtheta^{(t)}-\bbtheta^{(t_0)})}_2^2\\
    &\leq\sum_{k=1}^Kp_k\norm{\btheta^{(t)}_k-\bbtheta^{(t_0)}}_2^2+\underbrace{\sum_{k=1}^Kp_k\norm{\bbtheta^{(t)}-\bbtheta^{(t_0)}}_2^2}_{\sum p_k=1}\\
    &=\sum_{k=1}^Kp_k\norm{\sum_{t=t_0}^{t-1}\eta^{(t)}g_k(\btheta_k^{(t)};\xi_k^{(t)})}_2^2+\norm{\sum_{k=1}^Kp_k\sum_{t=t_0}^{t-1}\eta^{(t)}g_k(\btheta_k^{(t)};\xi_k^{(t)})}^2_2\\
    &\leq \sum_{k=1}^Kp_k(t-t_0)\sum_{t=t_0}^{t-1}\eta^{(t)2}\norm{g(\btheta_k^{(t)};\xi_k^{(t)})}_2^2 +  \sum_{k=1}^Kp_k\norm{\sum_{t=t_0}^{t-1}\eta^{(t)}g_k(\btheta_k^{(t)};\xi_k^{(t)})}^2_2\\
    &\leq  2\sum_{k=1}^Kp_k(E-1)\sum_{t=t_0}^{t-1}\eta^{(t)2}\norm{g(\btheta_k^{(t)};\xi_k^{(t)})}_2^2\leq 2\sum_{k=1}^Kp_k(E-1)\sum_{t=t_0}^{t-1}\eta^{(t_0)2}G^2 = 2\sum_{k=1}^Kp_k(E-1)^2\eta^{(t_0)2}G^2.
\end{align*}
Without loss of generality, assume $\eta^{(t_0)}\leq2\eta^{(t)}$ since the learning rate is decreasing. Therefore, $\text{D}\leq 8(E-1)^2\eta^{(t)2}G^2$. To simplify E, we have
\begin{align*}
    \text{E}=\sum_{k=1}^Kp_k\frac{\gamma_k}{2}\norm{\btheta^{(t)}_k-\btheta^*}_2^2&\geq \min_k\gamma_k\frac{1}{2}\sum_{k=1}^Kp_k\norm{\btheta^{(t)}_k-\btheta^*}_2^2\\
    &\geq \min_k\gamma_k\frac{1}{2}\norm{\sum_{k=1}^Kp_k(\btheta^{(t)}_k-\btheta^*)}_2^2= \min_k\gamma_k\frac{1}{2}\norm{\bbtheta^{(t)}-\btheta^*}_2^2,
\end{align*}
using Jensen's inequality.

Therefore, we obtain
\begin{align*}
    &\norm{\bar{\bm{v}}^{(t+1)}-\btheta^*}_2^2\\
    &\leq\norm{\bbtheta^{(t)}-\btheta^*}_2^2 + 8(E-1)^2\eta^{(t)2}G^2+\eta^{(t)2}G^2\\
    &\qquad -2\eta^{(t)}\min_k\gamma_k\frac{1}{2}\norm{\bbtheta^{(t)}-\btheta^*}_2^2+2\eta^{(t)}\sum_{k=1}^Kp_kC_{3k}(\alpha_k,b_k)\frac{\log M_k}{M_k} + \eta^{(t)2}G^2\\
    &\qquad -2\eta^{(t)}\sum_{k=1}^Kp_k\langle\btheta^{(t)}_k-\btheta^*,g_k(\btheta_k^{(t)};\xi_k^{(t)})-g_k^*(\btheta_k^{(t)})\rangle\\
    &=\left(1-2\eta^{(t)}\min_k\gamma_k\frac{1}{2}\right)\norm{\bbtheta^{(t)}-\btheta^*}_2^2+(8(E-1)^2\eta^{(t)2}+2\eta^{(t)2})G^2+2\eta^{(t)}\sum_{k=1}^Kp_kC_{3k}(\alpha_k,b_k)\frac{\log M_k}{M_k}\\
    &\qquad -2\eta^{(t)}\sum_{k=1}^Kp_k\langle\btheta^{(t)}_k-\btheta^*,g_k(\btheta_k^{(t)};\xi_k^{(t)})-g_k^*(\btheta_k^{(t)})\rangle\\
    &\leq\left(1-2\eta^{(t)}\min_k\gamma_k\frac{1}{2}\right)\norm{\bbtheta^{(t)}-\btheta^*}_2^2+(8(E-1)^2\eta^{(t)2}+2\eta^{(t)2})G^2+2\eta^{(t)}\max_kC_{3k}(\alpha_k,b_k)\frac{\log M_k}{M_k}\\
    &\qquad -2\eta^{(t)}\sum_{k=1}^Kp_k\langle\btheta^{(t)}_k-\btheta^*,g_k(\btheta_k^{(t)};\xi_k^{(t)})-g_k^*(\btheta_k^{(t)})\rangle\\
    &=\left(1-2\eta^{(t)}\min_k\gamma_k\frac{1}{2}\right)\norm{\bbtheta^{(t)}-\btheta^*}_2^2+\left(8(E-1)^2+2\right)\eta^{(t)2}G^2+2\eta^{(t)}\bigg(\max_kC_{3k}(\alpha_k,b_k)\frac{\log M_k}{M_k}\\
    &\qquad -\sum_{k=1}^Kp_k\langle\btheta^{(t)}_k-\btheta^*,g_k(\btheta_k^{(t)};\xi_k^{(t)})-g_k^*(\btheta_k^{(t)})\rangle\bigg).
\end{align*}
Since $\frac{3}{2\min_k\gamma_k}\leq\beta_1\leq\frac{2}{\min_k\gamma_k}$ and $\eta^{(t)}=\frac{\beta_1}{t}$ for all $t\geq 1$. Here we set $\eta^{(0)}=\beta_1$. We will show
\begin{align*}
    &\norm{\bbtheta^{(t)}-\btheta^*}_2^2\leq\frac{2\beta_1^2\left(8(E-1)^2+2\right)G^2}{t} \\
    &\qquad + \sum_{u=0}^{t-1}\left(2\eta^{(u+1)}\prod_{v=u+2}^t(1-\eta^{(v)}\min_k\gamma_k)\right)\bigg(\max_kC_{3k}(\alpha_k,b_k)\frac{\log M_k}{M_k} -\sum_{k=1}^Kp_k\langle\btheta^{(u)}_k-\btheta^*,g_k(\btheta_k^{(u)};\xi_k^{(u)})-g_k^*(\btheta_k^{(u)})\rangle\bigg)
\end{align*}
by induction. When $t=1$, we have
\begin{align*}
    &\norm{\bbtheta^{(1)}-\btheta^*}_2^2\leq\left(8(E-1)^2+2\right)\beta_1^2G^2+2\beta_1\bigg(\max_kC_{3k}(\alpha_k,b_k)\frac{\log M_k}{M_k}\\
    &\qquad -\sum_{k=1}^Kp_k\langle\btheta^{(0)}_k-\btheta^*,g_k(\btheta_k^{(0)};\xi_k^{(0)})-g_k^*(\btheta_k^{(0)})\rangle\bigg)
\end{align*}
since $\left(1-2\eta^{(0)}\min_k\gamma_k\frac{1}{2}\right)<0$. Assume the inequality holds for $t=l\geq 1$, then we have
\begin{align*}
    &\norm{\bbtheta^{(l+1)}-\btheta^*}_2^2\\
    &\leq\left(1-2\eta^{(l)}\min_k\gamma_k\frac{1}{2}\right)\bigg\{\frac{2\beta_1^2\left(8(E-1)^2+2\right)G^2}{l} \\
    &\qquad + \sum_{u=0}^{l-1}2\eta^{(u+1)}\prod_{v=u+2}^{l}(1-\eta^{(v)}\min_k\gamma_k)\bigg(\max_kC_{3k}(\alpha_k,b_k)\frac{\log M_k}{M_k} -\sum_{k=1}^Kp_k\langle\btheta^{(u)}_k-\btheta^*,g_k(\btheta_k^{(u)};\xi_k^{(u)})-g_k^*(\btheta_k^{(u)})\rangle\bigg)\bigg\}\\
    &\qquad +\left(8(E-1)^2+2\right)\eta^{(l)2}G^2+2\eta^{(l)}\bigg(\max_kC_{3k}(\alpha_k,b_k)\frac{\log M_k}{M_k}-\sum_{k=1}^Kp_k\langle\btheta^{(l)}_k-\btheta^*,g_k(\btheta_k^{(l)};\xi_k^{(l)})-g_k^*(\btheta_k^{(l)})\rangle\bigg)\\
    &\leq\frac{2\beta_1^2\left(8(E-1)^2+2\right)G^2}{l+1}\\
    &\qquad +  \sum_{u=0}^{l}2\eta^{(u+1)}\prod_{v=u+2}^{l}(1-\eta^{(v)}\min_k\gamma_k)\bigg(\max_kC_{3k}(\alpha_k,b_k)\frac{\log M_k}{M_k} -\sum_{k=1}^Kp_k\langle\btheta^{(u)}_k-\btheta^*,g_k(\btheta_k^{(u)};\xi_k^{(u)})-g_k^*(\btheta_k^{(u)})\rangle\bigg).
\end{align*}
To derive above inequality, we can first show that
\begin{align*}
    \left(1-2\eta^{(l)}\min_k\gamma_k\frac{1}{2}\right)\frac{2\beta_1^2\left(8(E-1)^2+2\right)G^2}{l}  + \left(8(E-1)^2+2\right)\eta^{(l)2}G^2\leq \frac{2\beta_1^2\left(8(E-1)^2+2\right)G^2}{l+1}
\end{align*}
as long as $\beta_1\geq\frac{3l+1}{2l+2}\frac{1}{\min_k\gamma_k}$. This is true since the right-hand side is always less or equal to $\frac{3}{2\min_k\gamma_k}$. The remaining part in the above inequality is apparent since $1-2\eta^{(l)}\min_k\gamma_k\frac{1}{2}\leq 1$. Thus, the proof of the induction step is complete. Using this fact, it can be shown that
\begin{align*}
    &\norm{\bbtheta^{(t+1)}-\btheta^*}_2^2\\
    &\leq\frac{2\beta_1^2\left(8(E-1)^2+2\right)G^2}{t+1}\\
    &\qquad +  \sum_{u=0}^{t}2\eta^{(u+1)}\prod_{v=u+2}^{t}(1-\eta^{(v)}\min_k\gamma_k)\bigg(\max_kC_{3k}(\alpha_k,b_k)\frac{\log M_k}{M_k} -\sum_{k=1}^Kp_k\langle\btheta^{(u)}_k-\btheta^*,g_k(\btheta_k^{(u)};\xi_k^{(u)})-g_k^*(\btheta_k^{(u)})\rangle\bigg)\\
    &\leq\frac{2\beta_1^2\left(8(E-1)^2+2\right)G^2}{t+1}\\
    &\qquad +  \sum_{u=0}^{t}\left(2\eta^{(u+1)}\prod_{v=u+2}^{t}(1-\eta^{(v)}\min_k\gamma_k)\right)\bigg(\max_kC_{3k}(\alpha_k,b_k)\frac{\log M_k}{M_k}\bigg)\\ 
    &\qquad -\sum_{u=0}^{t}\left(2\eta^{(u+1)}\prod_{v=u+2}^{t}(1-\eta^{(v)}\min_k\gamma_k)\right)\bigg(\sum_{k=1}^Kp_k\langle\btheta^{(u)}_k-\btheta^*,g_k(\btheta_k^{(u)};\xi_k^{(u)})-g_k^*(\btheta_k^{(u)})\rangle\bigg)\\
    &\leq\frac{2\beta_1^2\left(8(E-1)^2+2\right)G^2}{t+1}\\
    &\qquad +  \sum_{u=0}^{t}\frac{2\beta_1}{t+1}\bigg(\max_kC_{3k}(\alpha_k,b_k)\frac{\log M_k}{M_k}\bigg)+\sum_{u=0}^{t}\frac{2\beta_1}{t+1}\bigg(\sum_{k=1}^Kp_k\norm{\btheta^{(u)}_k-\btheta^*}_2\norm{g_k(\btheta_k^{(u)};\xi_k^{(u)})-g_k^*(\btheta_k^{(u)})}_2\bigg)\\
    &\leq\frac{2\beta_1^2\left(8(E-1)^2+2\right)G^2}{t+1}\\
    &\qquad +  \bigg(\max_kC_{3k}(\alpha_k,b_k)\frac{\log M_k}{M_k}\bigg)+\sum_{u=0}^{t}\frac{2\beta_1}{t+1}\bigg(\sum_{k=1}^Kp_k\norm{\btheta^{(u)}_k-\btheta^*}_2\norm{g_k(\btheta_k^{(u)};\xi_k^{(u)})-g_k^*(\btheta_k^{(u)})}_2\bigg)\\
    &\leq\frac{2\beta_1^2\left(8(E-1)^2+2\right)G^2}{t+1}\\
    &\qquad +  \max_kC_{3k}(\alpha_k,b_k)\frac{\log M_k}{M_k}+\frac{2\beta_1}{t+1}\sum_{u=0}^{t}\bigg(\sum_{k=1}^K\sqrt{2}p_k(\theta_{max}-\theta_{min})\norm{g_k(\btheta_k^{(u)};\xi_k^{(u)})-g_k^*(\btheta_k^{(u)})}_2\bigg)\\
    &\leq\frac{2\beta_1^2\left(8(E-1)^2+2\right)G^2}{t+1}\\
    &\qquad + 2\beta_1 \max_kC_{3k}(\alpha_k,b_k)\frac{\log M_k}{M_k}+2\beta_1\max_{0\leq u\leq t}\bigg(\sum_{k=1}^K\sqrt{2}p_k(\theta_{max}-\theta_{min})\norm{g_k(\btheta_k^{(u)};\xi_k^{(u)})-g_k^*(\btheta_k^{(u)})}_2\bigg)
\end{align*}
In the third inequality, we use the Cauchy–Schwarz inequality and the fact that $2\eta^{(u+1)}\prod_{v=u+2}^{t}(1-\eta^{(v)}\min_k\gamma_k)\leq2\frac{\beta_1}{u+1}\prod_{v=u+2}^{t}(1-\frac{3}{2v})\leq\frac{2\beta_1}{t+1}$. 

Let $\phi_k=(\log M_k)^{\epsilon_k-\frac{1}{2}}$. By Lemma \ref{lemma:4} and using a union bound over $u$, with probability at least $1-C_{\btheta}(T+1)\exp(-c_{\btheta}(\log M_k)^{2\epsilon_k})$, we have 
\begin{align*}
    \max_{0\leq u\leq t}\norm{g_k(\btheta_k^{(u)};\xi_k^{(u)})-g_k^*(\btheta_k^{(u)})}_2\leq C_{\btheta} (\log M_k)^{\epsilon_k-\frac{1}{2}}.
\end{align*}
Therefore,
\begin{align*}
    &\norm{\bbtheta^{(t+1)}-\btheta^*}_2^2\\
    &\leq\frac{2\beta_1^2\left(8(E-1)^2+2\right)G^2}{t+1}+2\beta_1\max_kC_{3k}(\alpha_k,b_k)\frac{\log M_k}{M_k} + 2\beta_1\max_{0\leq u\leq t}\sum_{k=1}^K\sqrt{2}p_k(\theta_{max}-\theta_{min})  C_{\btheta} (\log M_k)^{\epsilon_k-\frac{1}{2}}\\
    &=\frac{2\beta_1^2\left(8(E-1)^2+2\right)G^2}{t+1}+2\beta_1\max_kC_{3k}(\alpha_k,b_k)\frac{\log M_k}{M_k} +2\sqrt{2}\beta_1(\theta_{max}-\theta_{min}) C_{\btheta} \sum_{k=1}^Kp_k(\log M_k)^{\epsilon_k-\frac{1}{2}}.
\end{align*}

Using the same proof technique, we can also derive a same bound on $\norm{\bar{\theta}_2^{(t+1)}-\theta_2^*}_2^2$. Let $\phi_k=M_k^{\epsilon_k-\frac{1}{2}}$. By Lemma \ref{lemma:4} and using a union bound over $u$, with probability at least $1-C_{\theta}(t+1)(\log( M_k^{\epsilon_k-\frac{1}{2}}))^4\exp\{-c_{\theta}M_k^{2\epsilon_k}\}$,
\begin{align*}
    \max_{0\leq u\leq t}\norm{g_k(\btheta_k^{(u)};\xi_k^{(u)})-g_k^*(\btheta_k^{(u)})}_2\leq C M_k^{\epsilon_k-\frac{1}{2}}.
\end{align*}
Therefore,
\begin{align*}
    &\norm{\bar{\theta}_2^{(t+1)}-\theta_2^*}_2^2\\
    &\leq\frac{2\beta_1^2\left(8(E-1)^2+2\right)G^2}{t+1}+2\beta_1\max_kC_{3k}(\alpha_k,b_k)\frac{\log M_k}{M_k} +2\beta_1\max_{0\leq u\leq t}\sum_{k=1}^K\sqrt{2}p_k(\theta_{max}-\theta_{min}) C M_k^{\epsilon_k-\frac{1}{2}}\\
    &=\frac{2\beta_1^2\left(8(E-1)^2+2\right)G^2}{t+1}+2\beta_1\max_kC_{3k}(\alpha_k,b_k)\frac{\log M_k}{M_k} +2\sqrt{2}\beta_1(\theta_{max}-\theta_{min}) C \sum_{k=1}^Kp_kM_k^{\epsilon_k-\frac{1}{2}}.
    % \\ &\leq \frac{8K^2\left(4(E-1)^2+2\right)G^2}{(\min_k\gamma_k)^2(t+1)}+\left(2\beta_1\max_kC_{3k}(\alpha_k,b_k)\frac{(\log M)^{-\epsilon}}{M} +2C\tau\beta_1\max_{0\leq u\leq t}\sum_{k=1}^Kp_k(\theta_{max}-\theta_{min})\frac{1}{M}\right)(\log M)^{\epsilon+1}.
\end{align*}

%~~~~~~~~~~~~~~~~~~~~~~~~~~~~~~~~~~~~~~~~~~~%
%~~~~~~~~~~~~~~~~~~~~~~~~~~~~~~~~~~~~~~~~~~~%
%~~~~~~~~~~~~~~~~~~~~~~~~~~~~~~~~~~~~~~~~~~~%
%~~~~~~~~~~~~~~~~~~~~~~~~~~~~~~~~~~~~~~~~~~~%
%~~~~~~~~~~~~~~~~~~~~~~~~~~~~~~~~~~~~~~~~~~~%
%~~~~~~~~~~~~~~~~~~~~~~~~~~~~~~~~~~~~~~~~~~~%
%~~~~~~~~~~~~~~~~~~~~~~~~~~~~~~~~~~~~~~~~~~~%
%~~~~~~~~~~~~~~~~~~~~~~~~~~~~~~~~~~~~~~~~~~~%

\subsection{Proof of Theorem \ref{theorem:parameter_partial_exp}}

We slightly modify the definition of $\btheta^{(t+1)}_k$ such that $\btheta^{(t+1)}_k=\frac{1}{|\mathcal{S}_c|}\sum_{k\in\mathcal{S}_c}\bm{v}_k^{(t+1)}$ if $t+1\in\mathcal{I}_E$. Under the scenario of asynchronous update, it can be seen that $\bbtheta^{(t+1)}\neq\bar{\bm{v}}^{(t+1)}$. Therefore, we want to establish a bound on the difference $\norm{\bbtheta^{(t+1)}-\bar{\bm{v}}^{(t+1)}}^2_2$. We have
\begin{align*}
    \norm{\bbtheta^{(t+1)}-\btheta^*}_2^2&=\norm{\bbtheta^{(t+1)}-\bar{\bm{v}}^{(t+1)}+\bar{\bm{v}}^{(t+1)}-\btheta^*}_2^2\\
    &=\underbrace{\norm{\bbtheta^{(t+1)}-\bar{\bm{v}}^{(t+1)}}_2^2}_{\text{A}} + \underbrace{\norm{\bar{\bm{v}}^{(t+1)}-\btheta^*}_2^2}_{\text{B}}+\underbrace{2\langle\bbtheta^{(t+1)}-\bar{\bm{v}}^{(t+1)},\bar{\bm{v}}^{(t+1)}-\btheta^*\rangle}_{\text{C}}.
\end{align*}

We can show
\begin{align*}
    \mathbb{E}_{\mathcal{S}_c}\left\{\bbtheta^{(t+1)}\right\}=\mathbb{E}_{\mathcal{S}_c}\left\{\frac{1}{|\mathcal{S}_c|}\sum_{k\in\mathcal{S}_c}\bm{v}_k^{(t+1)} \right\}=\frac{1}{|\mathcal{S}_c|}\sum_{k\in\mathcal{S}_c}\mathbb{E}_{\mathcal{S}_c}\left\{\bm{v}_k^{(t+1)} \right\}=\mathbb{E}_{\mathcal{S}_c}\left\{\bm{v}_1^{(t+1)} \right\}=\sum_{k=1}^Kp_k\bm{v}_k^{(t+1)}=\bar{\bm{v}}^{(t+1)}
\end{align*}
since the sampling distribution is identical. Therefore, $\mathbb{E}_{\mathcal{S}_c}[\text{C}]=0$.

For part A, we have
\begin{align*}
    &\mathbb{E}_{\mathcal{S}_c}\left\{\norm{\bbtheta^{(t+1)}-\bar{\bm{v}}^{(t+1)}}_2^2\right\}\\
    &=\mathbb{E}_{\mathcal{S}_c}\left\{\frac{1}{|\mathcal{S}_c|^2}\sum_{k\in\mathcal{S}_c}\norm{\bm{v}_k^{(t+1)}-\bar{\bm{v}}^{(t+1)}}_2^2\right\}=\frac{1}{|\mathcal{S}_c|}\sum_{k=1}^Kp_k\norm{\bm{v}_k^{(t+1)}-\bar{\bm{v}}^{(t+1)}}_2^2.
\end{align*}
The first equality uses the fact that $\bm{v}_k^{(t+1)}$ is independent of each other and is an unbiased estimator of $\bar{\bm{v}}^{(t+1)}$. Therefore, we have
\begin{align*}
    \sum_{k=1}^Kp_k\norm{\bm{v}_k^{(t+1)}-\bar{\bm{v}}^{(t+1)}}_2^2&=\sum_{k=1}^Kp_k\norm{\bm{v}_k^{(t+1)}-\bbtheta^{(t_0)}-(\bbtheta^{(t_0)}-\bar{\bm{v}}^{(t+1)})}_2^2\\
    &\leq \sum_{k=1}^Kp_k\norm{\bm{v}_k^{(t+1)}-\bbtheta^{(t_0)}}_2^2\\
    &\leq\sum_{k=1}^Kp_k\norm{\bm{v}_k^{(t+1)}-\btheta_k^{(t_0)}}_2^2\\
    &=\sum_{k=1}^Kp_k\norm{\sum_{i=t_0}^t \eta^{(i)}g_k(\btheta_k^{(i)};\xi_k^{(i)})}_2^2\\
    &\leq\sum_{k=1}^Kp_k\sum_{i=t_0}^tE\norm{ \eta^{(i)}g_k(\btheta_k^{(i)};\xi_k^{(i)})}_2^2\\
    &\leq E^2\eta^{(t_0)2}G^2\leq 4E^2\eta^{(t)2}G^2
\end{align*}
where $t_0=t-E+1$ is the iteration where communication happens. Therefore, 
\begin{align*}
    \mathbb{E}_{\mathcal{S}_c}\left\{\norm{\bbtheta^{(t+1)}-\bar{\bm{v}}^{(t+1)}}_2^2\right\}\leq\frac{4E^2\eta^{(t)2}G^2}{|\mathcal{S}_c|}.
\end{align*}

For part B, we can follow the exact proof in Theorem \ref{theorem:parameter_full_exp} to get an upper bound after taking expectation with respect to $\mathcal{S}_c$. In a nutshell, we can obtain
\begin{align*}
    &\mathbb{E}_{\mathcal{S}_c}\left\{\norm{\bbtheta^{(t+1)}-\btheta^*}_2^2\right\}\\
    &\leq \frac{4E^2\eta^{(t)2}G^2}{|\mathcal{S}_c|} + \left(1-2\eta^{(t)}\min_k\gamma_k\frac{1}{2}\right)\norm{\bbtheta^{(t)}-\btheta^*}_2^2+\left(8(E-1)^2+2\right)\eta^{(t)2}G^2+2\eta^{(t)}\bigg(\max_kC_{3k}(\alpha_k,b_k)\frac{\log M_k}{M_k}\\
    &\qquad -\sum_{k=1}^Kp_k\langle\btheta^{(t)}_k-\btheta^*,g_k(\btheta_k^{(t)};\xi_k^{(t)})-g_k^*(\btheta_k^{(t)})\rangle\bigg).
\end{align*}
Following the induction proof in Theorem \ref{theorem:parameter_full_exp}, we have
\begin{align*}
    &\mathbb{E}_{\mathcal{S}_c}\left\{\norm{\bbtheta^{(t+1)}-\btheta^*}_2^2\right\}\\
    &\leq\frac{2\beta_1^2\left(\frac{1}{|\mathcal{S}_c|}4E^2+8(E-1)^2+2\right)G^2}{t+1}+2\beta_1\max_kC_{3k}(\alpha_k,b_k)\frac{\log M_k}{M_k} +2\sqrt{2}\beta_1(\theta_{max}-\theta_{min}) C \sum_{k=1}^Kp_kM_k^{\epsilon_k-\frac{1}{2}}.
\end{align*}

\subsection{Proof of Theorem \ref{theorem:parameter_matern}}

\textbf{Convergence of Parameter Iterate}

Define $C_{4k}\coloneqq(\theta_{max}-\theta_{min})^2\left(\frac{1}{2\theta_{min}^2}+\frac{C_{mat,k}(4b_k+3)}{2\theta_{min}^2(4b_k^2-6b_k-3)}\right)$. Following the same proof strategy in Theorem \ref{theorem:parameter_full_exp} and using Lemma \ref{lemma:2-2} and \ref{lemma:3-2}, we can show that
\begin{align*}
    &\norm{\bar{\theta}_2^{(t+1)}-\theta_2^*}_2^2\\
    &\leq\frac{2\beta_1^2\left(8(E-1)^2+2\right)G^2}{t+1} +  \sum_{u=0}^{t}2\eta^{(u+1)}\prod_{v=u+2}^{t}(1-\eta^{(v)}\min_k\gamma_k)\bigg(\max_kC_{4k}M_k^{\frac{(2+\alpha_k)(4b_k+3)}{4b_k(2b_k-1)}-1} -\\
    &\qquad \sum_{k=1}^Kp_k\langle\theta^{(u)}_{2k}-\theta_2^*,[g_k(\btheta_k^{(u)};\xi_k^{(u)})]_2-[g_k^*(\btheta_k^{(u)})]_2\rangle\bigg)\\
    &\leq\frac{2\beta_1^2\left(8(E-1)^2+2\right)G^2}{t+1} + 2\beta_1 \max_kC_{4k}M_k^{\frac{(2+\alpha_k)(4b_k+3)}{4b_k(2b_k-1)}-1}\\
    &\qquad +2\beta_1\max_{0\leq u\leq t}\bigg(\sum_{k=1}^K\sqrt{2}p_k(\theta_{max}-\theta_{min})\norm{[g_k(\btheta_k^{(u)};\xi_k^{(u)})]_2-[g_k^*(\btheta_k^{(u)})]_2}_2\bigg).
\end{align*}
Let $\phi_k=M_k^{\epsilon_k-\frac{1}{2}}$. By Lemma \ref{lemma:4}, for any $0<\alpha_k<\frac{8b_k^2-12b_k-6}{4b_k+3}, \epsilon_k<\frac{1}{2}$, with probability at least $1-C_{\btheta}(t+1)(\log( M_k^{\epsilon_k-\frac{1}{2}}))^4\exp\{-c_{\btheta}M_k^{2\epsilon_k}\}$, we have
\begin{align*}
    \max_{0\leq u\leq t}\norm{[g_k(\btheta_k^{(u)};\xi_k^{(u)})]_2-[g_k^*(\btheta_k^{(u)})]_2}_2\leq C_{\btheta} M_k^{\epsilon_k-\frac{1}{2}}.
\end{align*}
Therefore,
\begin{align*}
    &\norm{\bar{\theta}_2^{(t+1)}-\theta_2^*}_2^2\\
    &\leq\frac{2\beta_1^2\left(8(E-1)^2+2\right)G^2}{t+1}  +  2\beta_1 \max_kC_{4k}M_k^{\frac{(2+\alpha_k)(4b_k+3)}{4b_k(2b_k-1)}-1}+2\beta_1\bigg(\sum_{k=1}^K\sqrt{2}p_k(\theta_{max}-\theta_{min})C_{\btheta} M_k^{\epsilon_k-\frac{1}{2}}\bigg)\\
    &=\frac{2\beta_1^2\left(8(E-1)^2+2\right)G^2}{t+1} + \mathcal{O}\left(\max_kM_k^{-\frac{8b_k^2-12b_k-6-3\alpha_k-4\alpha_kb_k}{8b_k^2-4b_k}}\right) + \mathcal{O}\left(\sum_{k=1}^Kp_kM_k^{\epsilon_k-\frac{1}{2}}\right).
\end{align*}

The partial device participation proof is similar to Theorem \ref{theorem:parameter_partial_exp}. Again, using Lemma \ref{lemma:2-2} and \ref{lemma:3-2}, we can show that
\begin{align*}
    &\mathbb{E}_{\mathcal{S}_c}\left\{\norm{\bar{\theta}_2^{(t+1)}-\theta_2^*}_2^2\right\}\\
    &\leq\frac{2\beta_1^2\left(\frac{4E^2}{|\mathcal{S}_c|}+8(E-1)^2+2\right)G^2}{t+1}  +  2\beta_1 \max_kC_{4k}M_k^{\frac{(2+\alpha_k)(4b_k+3)}{4b_k(2b_k-1)}-1}+2\beta_1\bigg(\sum_{k=1}^K\sqrt{2}p_k(\theta_{max}-\theta_{min})C_{\btheta} M_k^{\epsilon_k-\frac{1}{2}}\bigg)\\
    &=\frac{2\beta_1^2\left(\frac{4E^2}{|\mathcal{S}_c|}+8(E-1)^2+2\right)G^2}{t+1} + \mathcal{O}\left(\max_kM_k^{-\frac{8b_k^2-12b_k-6-3\alpha_k-4\alpha_kb_k}{8b_k^2-4b_k}}\right) + \mathcal{O}\left(\sum_{k=1}^Kp_kM_k^{\epsilon_k-\frac{1}{2}}\right).
\end{align*}

\textbf{Convergence of Full Gradient}

We follow the same proof strategy in Theorem \ref{theorem:exp_hetero}. We defer this proof to the subsection after it.

%~~~~~~~~~~~~~~~~~~~~~~~~~~~~~~~~~~~~~~~~~~~%
%~~~~~~~~~~~~~~~~~~~~~~~~~~~~~~~~~~~~~~~~~~~%
%~~~~~~~~~~~~~~~~~~~~~~~~~~~~~~~~~~~~~~~~~~~%
%~~~~~~~~~~~~~~~~~~~~~~~~~~~~~~~~~~~~~~~~~~~%
%~~~~~~~~~~~~~~~~~~~~~~~~~~~~~~~~~~~~~~~~~~~%
%~~~~~~~~~~~~~~~~~~~~~~~~~~~~~~~~~~~~~~~~~~~%
%~~~~~~~~~~~~~~~~~~~~~~~~~~~~~~~~~~~~~~~~~~~%
%~~~~~~~~~~~~~~~~~~~~~~~~~~~~~~~~~~~~~~~~~~~%

\subsection{Proof of Theorem \ref{theorem:exp_hetero}}
\begin{proof}

% To do so, our first goal is to bound the gradient norm by the model parameter iterate (Part I). Kindly note we do not use Lipschitz constant in this proof. Our second goal is therefore to bound the norm of parameter iterate (Part II). Given results from Part I and II, we can achieve our final goal.
Our final goal is to bound the squared norm of full gradient $\norm{\nabla L(\bbtheta)}^2_2=\norm{\sum_{k=1}^Kp_k\nabla L_k(\bbtheta;D_k)}_2^2$. We define a conditional expectation of $\nabla L_k(\bbtheta^{(t)};D_k)$ as $\nabla L_k^*(\bbtheta^{(t)})\coloneqq\mathbb{E}\left(\nabla L_k(\bbtheta^{(t)};D_k)|\bm{X}_k\right)$. By the definition of $\nabla L_k^*(\bbtheta^{(t)})$, for $i\in\{1,2\}$, we have
\begin{align*}
    &\left[\nabla L_k^*(\bbtheta^{(t)})\right]_i \\
    &= \frac{1}{2N_k}\Tr\left[\bm{K}_{N_k}(\bbtheta^{(t)})^{-1}\left(\bm{I}_{N_k}-\bm{K}_{N_k}(\btheta_k^*)\bm{K}_{N_k}(\bbtheta^{(t)})^{-1}\right)\frac{\partial \bm{K}_{N_k}(\bbtheta^{(t)})}{\partial\bar{\theta}_i^{(t)}} \right]\\
    &=\frac{1}{2N_k}\Tr\left[\bm{K}_{N_k}(\bbtheta^{(t)})^{-1}\left(\bm{K}_{N_k}(\bbtheta^{(t)})-\bm{K}_{N_k}(\btheta_k^*)\right)\bm{K}_{N_k}(\bbtheta^{(t)})^{-1}\frac{\partial \bm{K}_{N_k}(\bbtheta^{(t)})}{\partial\bar{\theta}_i^{(t)}} \right].
\end{align*}
where $\btheta_k^*\coloneqq(\theta_{1k}^*,\theta_{2k}^*)$ is the set of optimal model parameters for device $k$. By definition, $\bm{K}_{N_k}(\btheta_k)=\theta_{1k}\bm{K}_{f,N_k}+\theta_{2k}\bm{I}_{N_k}$, where $\theta_{1k},\theta_{2k}$ are device-specific model parameters. Therefore, we obtain
\begin{align*}
    &\left[\nabla L_k^*(\bbtheta^{(t)})\right]_i \\
    &=\frac{1}{2{N_k}}\Tr\left[\bm{K}_{N_k}(\bbtheta^{(t)})^{-1}\left( (\bar{\theta}_1^{(t)}-\theta_{1k}^*)\bm{K}_{f,N_k}+ (\bar{\theta}_2^{(t)}-\theta_{2k}^*)\bm{I}_{N_k} \right)\bm{K}_{N_k}(\bbtheta^{(t)})^{-1}\frac{\partial \bm{K}_{N_k}(\bbtheta^{(t)})}{\partial\bar{\theta}_i^{(t)}} \right],
\end{align*}
where $\bar{\theta}_i^{(t)}=\sum_{k=1}^Kp_k\theta_{ik}^{(t)}, i=1,2$.

By Eigendecomposition, we can write $\bm{K}_{f,N_k}=\bm{Q}_{N_k}\bm{\Lambda}_{N_k}\bm{Q}_{N_k}^{-1}$ where $\bm{Q}_{N_k}$ contains eigenvectors of $\bm{K}_{f,N_k}$, $\bm{\Lambda}_{N_k}\coloneqq\text{diag}(\lambda_{11},\lambda_{12},\ldots,\lambda_{1N_k})$ is a diagonal matrix with eigenvalues of $\bm{K}_{f,N_k}$ and $\lambda_{1j}$ is the $j^{th}$ largest eigenvalue of $\bm{K}_{f,N_k}$. Here note that the values of $\lambda_{\cdot\cdot}$ are different for each device $k$. For simplicity, we drop the notation $k$ in the eigenvalues unless there is an ambiguity. When $i=1$, we can simplify $\left[\nabla L_k^*(\bbtheta^{(t)})\right]_i$ as
\begin{align*}
    &\left[\nabla L_k^*(\bbtheta^{(t)})\right]_1 \\
    &=\frac{1}{2N_k}\Tr\left[\bm{K}_{N_k}(\bbtheta^{(t)})^{-1}\left( (\bar{\theta}_1^{(t)}-\theta_{1k}^*)\bm{K}_{f,N_k}+ (\bar{\theta}_1^{(t)}-\theta_{2k}^*)\bm{I}_{N_k} \right)\bm{K}_{N_k}(\bbtheta^{(t)})^{-1}\bm{K}_{f,N_k} \right]\\
    &=\frac{1}{2N_k}(\bar{\theta}_1^{(t)}-\theta_{1k}^*)\sum_{j=1}^{N_k}\frac{\lambda_{1j}^2}{\left(\bar{\theta}_{1}^{(t)}\lambda_{1j}+\bar{\theta}_{2}^{(t)}\lambda_{2j}\right)^2} + \frac{1}{2N_k}(\bar{\theta}_2^{(t)}-\theta_{2k}^*)\sum_{j=1}^{N_k}\frac{\lambda_{2j}\lambda_{1j}}{\left(\bar{\theta}_{1}^{(t)}\lambda_{1j}+\bar{\theta}_{2}^{(t)}\lambda_{2j}\right)^2}.
\end{align*}
where $\lambda_{2j}=1$ is the $j^{th}$ largest eigenvalue of $\bm{I}_{N_k}$. Similarly, it can be shown that, when $i=2$,
\begin{align*}
    &\left[\nabla L^*(\bbtheta^{(t)})\right]_2 \\
    &=\frac{1}{2N_k}(\bar{\theta}_1^{(t)}-\theta_{1k}^*)\sum_{j=1}^{N_k}\frac{\lambda_{1j}\lambda_{2j}}{\left(\bar{\theta}_{1}^{(t)}\lambda_{1j}+\bar{\theta}_{2}^{(t)}\lambda_{2j}\right)^2} + \frac{1}{2N_k}(\bar{\theta}_2^{(t)}-\theta_{2k}^*)\sum_{j=1}^{N_k}\frac{\lambda_{2j}^2}{\left(\bar{\theta}_{1}^{(t)}\lambda_{1j}+\bar{\theta}_{2}^{(t)}\lambda_{2j}\right)^2}.
\end{align*}

Our first goal is to bound eigenvalues of $\bm{K}_{f,N_k}$ using Lemma \ref{lemma:1} and \ref{lemma:2}. 

% Then we can bound $\sum_{j=1}^{N_k}\frac{\lambda_{1j}^2}{\left(\theta_1^{(t)}\lambda_{1j}+\theta_2^{(t)}\lambda_{2j}\right)^2}$, $\sum_{j=1}^{N_k}\frac{\lambda_{1j}\lambda_{2j}}{\left(\theta_1^{(t)}\lambda_{1j}+\theta_2^{(t)}\lambda_{2j}\right)^2}$ and $\sum_{j=1}^{N_k}\frac{\lambda_{2j}^2}{\left(\theta_1^{(t)}\lambda_{1j}+\theta_2^{(t)}\lambda_{2j}\right)^2}$. 

% By deriving bounds on those three terms, we can then bound $\left[\nabla L^*(\bbtheta^{(t)})\right]_i$ for $i=1,2$.

\paragraph{Part I: Bounding eigenvalues}
By Lemma \ref{lemma:1} and \ref{lemma:2}, for any $0<\epsilon_k,\alpha_k<1$, $C_{1k}(\alpha,b)>0$ and $N_k>C_{2k}(\epsilon_k,b_k)$, with probability at least $1-\frac{3}{N_k^{\alpha_k}}$, 
\begin{align*}
    \frac{\epsilon_k\log N_k}{8b_k\theta_{max}^2}\leq\sum_{j=1}^{N_k}\frac{\lambda_{1j}^2}{\left(\bar{\theta}_{1}^{(t)}\lambda_{1j}+\bar{\theta}_{2}^{(t)}\lambda_{2j}\right)^2}&\leq \frac{4+2\alpha_k}{b_k\theta_{min}^2}\log N_k\\
    \frac{N_k-C_{1k}(\alpha_k,b_k)\log N_k}{4\theta_{max}^2}\leq\sum_{j=1}^{N_k}\frac{\lambda_{2j}^2}{\left(\bar{\theta}_{1}^{(t)}\lambda_{1j}+\bar{\theta}_{2}^{(t)}\lambda_{2j}\right)^2}&\leq \frac{N_k}{\theta_{min}^2}\\
    0<\sum_{j=1}^{N_k}\frac{\lambda_{1j}\lambda_{2j}}{\left(\bar{\theta}_{1}^{(t)}\lambda_{1j}+\bar{\theta}_{2}^{(t)}\lambda_{2j}\right)^2}&\leq \frac{5+2\alpha_k}{7b_k\theta_{min}^2}\log N_k.
\end{align*}
Therefore, we can show that, with probability at least $1-\frac{3}{N_k^{\alpha_k}}$, 
\begin{align*}
    \left[\nabla L_k^*(\bbtheta^{(t)})\right]_1&=\frac{1}{2N_k}(\bar{\theta}_{1}^{(t)}-\theta_{1k}^*)\sum_{j=1}^{N_k}\frac{\lambda_{1j}^2}{\left(\bar{\theta}_1^{(t)}\lambda_{1j}+\bar{\theta}_2^{(t)}\lambda_{2j}\right)^2} + \frac{1}{2N_k}(\bar{\theta}_2^{(t)}-\theta_{2k}^*)\sum_{j=1}^{N_k}\frac{\lambda_{2j}\lambda_{1j}}{\left(\bar{\theta}_1^{(t)}\lambda_{1j}+\bar{\theta}_2^{(t)}\lambda_{2j}\right)^2}\\
    &\leq\frac{1}{2N_k}(\bar{\theta}_1^{(t)}-\theta_{1k}^*) \frac{4+2\alpha_k}{b_k\theta_{min}^2}\log N_k + \frac{1}{2N_k}\frac{5+2\alpha_k}{7b_k\theta_{min}^2}\log N_k\\
    &\leq\frac{(\theta_{max}-\theta_{min})(33+16\alpha_k)}{14N_kb_k\theta_{min}^2}\log N_k=\frac{(\theta_{max}-\theta_{min})(33+16\alpha_k)}{14b_k\theta_{min}^2}\frac{\log N_k}{N_k},
\end{align*}
and
\begin{align*}
    \left[\nabla L_k^*(\bbtheta^{(t)})\right]_1&=\frac{1}{2N_k}(\bar{\theta}_1^{(t)}-\theta_{1k}^*)\sum_{j=1}^{N_k}\frac{\lambda_{1j}^2}{\left(\bar{\theta}_1^{(t)}\lambda_{1j}+\bar{\theta}_2^{(t)}\lambda_{2j}\right)^2} + \frac{1}{2N_k}(\bar{\theta}_2^{(t)}-\theta_{2k}^*)\sum_{j=1}^N\frac{\lambda_{2j}\lambda_{1j}}{\left(\bar{\theta}_1^{(t)}\lambda_{1j}+\bar{\theta}_2^{(t)}\lambda_{2j}\right)^2}\\
    &\geq \frac{1}{2N_k}\frac{\epsilon_k\log N_k}{8b_k\theta_{max}^2}=\frac{\epsilon_k}{16b_k\theta_{max}^2}\frac{\log N_k}{N_k}>0.
\end{align*}
% By Lemma \ref{lemma:4}, with probability at least $1-C_{\theta,k}(t+1)(\log( M_k^{\epsilon-\frac{1}{2}}))^4\exp\{-c_{\theta,k}M_k^{2\epsilon_k}\}$, we have

Similarly, it can be shown that
\begin{align*}
    &\left[\nabla L_k^*(\bbtheta^{(t)})\right]_2 \\
    &=\frac{1}{2N_k}(\bar{\theta}_1^{(t)}-\theta_{1k}^*)\sum_{j=1}^{N_k}\frac{\lambda_{1j}\lambda_{2j}}{\left(\bar{\theta}_1^{(t)}\lambda_{1j}+\bar{\theta}_2^{(t)}\lambda_{2j}\right)^2} + \frac{1}{2N_k}(\bar{\theta}_2^{(t)}-\theta_{2k}^*)\sum_{j=1}^{N_k}\frac{\lambda_{2j}^2}{\left(\bar{\theta}_1^{(t)}\lambda_{1j}+\bar{\theta}_2^{(t)}\lambda_{2j}\right)^2}\\
    &\leq\frac{1}{2N_k}(\bar{\theta}_1^{(t)}-\theta_{1k}^*)\frac{5+2\alpha_k}{7b_k\theta_{min}^2}\log N_k + \frac{1}{2N_k}(\bar{\theta}_2^{(t)}-\theta_{2k}^*)\frac{N_k}{\theta_{min}^2}\\
    &\leq\frac{(\theta_{max}-\theta_{min})(5+2\alpha_k)}{14b_k\theta_{min}^2}\frac{\log N_k}{N_k} + (\bar{\theta}_2^{(t)}-\theta_{2k}^*)\frac{1}{2\theta_{min}^2},
\end{align*}
and
\begin{align*}
    &\left[\nabla L_k^*(\bbtheta^{(t)})\right]_2 \\
    &=\frac{1}{2N_k}(\bar{\theta}_1^{(t)}-\theta_{1k}^*)\sum_{j=1}^{N_k}\frac{\lambda_{1j}\lambda_{2j}}{\left(\bar{\theta}_1^{(t)}\lambda_{1j}+\bar{\theta}_2^{(t)}\lambda_{2j}\right)^2} + \frac{1}{2N_k}(\bar{\theta}_2^{(t)}-\theta_{2k}^*)\sum_{j=1}^{N_k}\frac{\lambda_{2j}^2}{\left(\bar{\theta}_1^{(t)}\lambda_{1j}+\bar{\theta}_2^{(t)}\lambda_{2j}\right)^2}\\
    &\geq \frac{1}{2N_k}(\bar{\theta}_2^{(t)}-\theta_{2k}^*)\frac{N_k-C_{1k}(\alpha_k,b_k)\log N_k}{4\theta_{max}^2}\\
    &\geq (\bar{\theta}_2^{(t)}-\theta_{2k}^*)\frac{1}{8\theta_{max}^2} - \frac{(\theta_{max}-\theta_{min})C_{1k}(\alpha_k,b_k)}{8\theta_{max}^2}\frac{\log N_k}{N_k}.
\end{align*}
By combining above inequalities, we obtain
\begin{align*}
    &\norm{\nabla L_k^*(\bbtheta^{(t)})}^2_2 \\
    &= \left(\left[\nabla L_k^*(\bbtheta^{(t)})\right]_1^2+\left[\nabla L_k^*(\bbtheta^{(t)})\right]_2^2\right)\\
    &\leq \left\{ \left(\frac{(\theta_{max}-\theta_{min})(33+16\alpha_k)}{14b_k\theta_{min}^2}\frac{\log N_k}{N_k}\right)^2 + \left(\frac{(\theta_{max}-\theta_{min})(5+2\alpha_k)}{14b_k\theta_{min}^2}\frac{\log N_k}{N_k} + (\bar{\theta}_2^{(t)}-\theta_{2k}^*)\frac{1}{2\theta_{min}^2}\right)^2  \right\}.
\end{align*}
Our next goal is therefore to study the behavior of $\bar{\theta}_2^{(t)}-\theta_{2k}^*$ during iteration and provide bound on this parameter iterate.

\paragraph{Part II: Bounding parameter iterates} We consider the full device participation scenario and the partial device participation scenario separately. 

\textbf{Under the full device participation scenario,} following the same procedure in the proof of Theorem \ref{theorem:parameter_full_exp}, we can show that
\begin{align*}
    &\norm{\bar{
    \theta}_2^{(t+1)}-\bar{\theta}_{2k}^*}_2^2
    \leq\frac{2\beta_1^2\left(8(E-1)^2+2\right)G^2}{t+1}\\
    &\qquad + 2\beta_1 \max_kC_{3k}(\alpha_k,b_k)\frac{\log M_k}{M_k}+2\beta_1\max_{0\leq u\leq t}\bigg(\sum_{k=1}^K\sqrt{2}p_k(\theta_{max}-\theta_{min})\norm{g_k(\btheta_k^{(u)};\xi_k^{(u)})-g_k^*(\btheta_k^{(u)})}_2\bigg)\\
    &\leq \frac{2\beta_1^2\left(8(E-1)^2+2\right)G^2}{t+1}+2\beta_1\max_kC_{3k}(\alpha_k,b_k)\frac{\log M_k}{M_k} +2\sqrt{2}\beta_1(\theta_{max}-\theta_{min}) C M_k^{\epsilon_k-\frac{1}{2}},
\end{align*}
with probability at least $1-C_{\theta,k}(t+1)(\log( M_k^{\epsilon-\frac{1}{2}}))^4\exp\{-c_{\theta,k}M_k^{2\epsilon_k}\}$.

\textbf{Under the partial device participation scenario,} following the same procedure in the proof of Theorem \ref{theorem:parameter_partial_exp}, we can show
\begin{align*}
    &\mathbb{E}_{\mathcal{S}_c}\left\{\norm{\bar{\theta}^{(t+1)}-\bar{\theta}_{2k}^*}_2^2\right\}\\
    &\leq\frac{2\beta_1^2\left(\frac{1}{|\mathcal{S}_c|}4E^2+8(E-1)^2+2\right)G^2}{t+1}+2\beta_1\max_kC_{3k}(\alpha_k,b_k)\frac{\log M_k}{M_k} +2\sqrt{2}\beta_1(\theta_{max}-\theta_{min}) C_k M_k^{\epsilon_k-\frac{1}{2}}.
\end{align*}

\paragraph{Part III: Bounding $\left[\nabla L(\bbtheta^{(t)})\right]_i$ for $i=1,2$ and Proving convergence}
Finally, equipped with all aforementioned results, we are going to prove our convergence result.

From Part I, we know
\begin{align*}
    &\norm{\nabla L_k^*(\bbtheta^{(t)})}^2_2 \\
    &\leq \left\{ \left(\frac{(\theta_{max}-\theta_{min})(33+16\alpha_k)}{14b_k\theta_{min}^2}\frac{\log N_k}{N_k}\right)^2 + \left(\frac{(\theta_{max}-\theta_{min})(5+2\alpha_k)}{14b_k\theta_{min}^2}\frac{\log N_k}{N_k} + (\bar{\theta}_2^{(t)}-\theta_{2k}^*)\frac{1}{2\theta_{min}^2}\right)^2  \right\}\\
    &\leq w_{1k}^2\left(\frac{\log N_k}{N_k}\right)^2 + w_{2k}^2\left(\frac{\log N_k}{N_k}\right)^2 + 2w_{2k}\frac{\log N_k}{N_k}(\bar{\theta}_2^{(t)}-\theta_{2k}^*)\frac{1}{2\theta_{min}^2} + \norm{\bar{\theta}_2^{(t)}-\bar{\theta}_{2k}^*}_2^2\frac{1}{4\theta_{min}^4}\\
    &\leq (w_{1k}^2+w_{2k}^2)\left(\frac{\log N_k}{N_k}\right)^2+ 2w_{2k}\frac{\log N_k}{N_k}(\theta_{max}-\theta_{min})\frac{1}{2\theta_{min}^2} \\
    &+ \left(\frac{2\beta_1^2\left(8(E-1)^2+2\right)G^2}{t}+2\beta_1\max_kC_{3k}(\alpha_k,b_k)\frac{\log M_k}{M_k} +2\sqrt{2}\beta_1(\theta_{max}-\theta_{min}) C M_k^{\epsilon_k-\frac{1}{2}}\right)\frac{1}{4\theta_{min}^4},
\end{align*}
where 
\begin{align*}
w_{1k}&=\frac{(\theta_{max}-\theta_{min})(33+16\alpha_k)}{14b_k\theta_{min}^2}, \\
w_{2k}&=\frac{(\theta_{max}-\theta_{min})(5+2\alpha_k)}{14b_k\theta_{min}^2}.
% w_{3k}&=2\beta_1\max_kC_{3k}(\alpha_k,b_k)\frac{(\log M_k)^{-\epsilon_k}}{M_k} +2C\tau\beta_1\max_{0\leq u\leq t}\sum_{k=1}^Kp_k(\theta_{max}-\theta_{min})\frac{1}{M}.
\end{align*}

By Lemma \ref{lemma:4}, with probability at least $1-C_{\theta,k}(t+1)(\log( M_k^{\epsilon-\frac{1}{2}}))^4\exp\{-c_{\theta,k}M_k^{2\epsilon_k}\}$,
\begin{align*}
    &\norm{\nabla L_k(\bbtheta^{(t)})}^2_2\leq \left(C_{\theta,k}M_k^{\epsilon_k-\frac{1}{2}}\right)^2 + \norm{\nabla L_k^*(\bbtheta^{(t)})}^2_2 + 2\norm{\nabla L_k^*(\bbtheta^{(t)})}_2\left(C_{\theta,k}M_k^{\epsilon_k-\frac{1}{2}}\right)\\
    &\leq C^2_{\theta,k}M_k^{2\epsilon_k-1} + (w_{1k}^2+w_{2k}^2)\left(\frac{\log N_k}{N_k}\right)^2+ 2w_{2k}\frac{\log N_k}{N_k}(\theta_{max}-\theta_{min})\frac{1}{2\theta_{min}^2} \\
    &+ \left(\frac{2\beta_1^2\left(8(E-1)^2+2\right)G^2}{t}+2\beta_1\max_kC_{3k}(\alpha_k,b_k)\frac{\log M_k}{M_k} +2\sqrt{2}\beta_1(\theta_{max}-\theta_{min}) C M_k^{\epsilon_k-\frac{1}{2}}\right)\frac{1}{4\theta_{min}^4}\\
    &+ 2\norm{\nabla L_k^*(\bbtheta^{(t)})}_2\left(C_{\theta,k}M_k^{\epsilon_k-\frac{1}{2}}\right).
\end{align*}
Therefore,
\begin{align*}
    &\norm{\nabla L(\bbtheta)}^2_2=\norm{\sum_{k=1}^Kp_k\nabla L_k(\bbtheta;D_k)}_2^2\leq \sum_{k=1}^Kp_k\norm{\nabla L_k(\bbtheta;D_k)}_2^2\leq\max_{k}\norm{\nabla L_k(\bbtheta;D_k)}_2^2\\
    &\leq \max_k\left(\frac{2\beta_1^2\left(8(E-1)^2+2\right)G^2}{4\theta_{min}^4t}+\mathcal{O}\left(\frac{\log M_k}{M_k}+M_k^{\epsilon_k-\frac{1}{2}}+\frac{\log N_k}{N_k}\right)\right).
\end{align*}
Under the partial device participation scenario, we have
\begin{align*}
    &\norm{\nabla L(\bbtheta)}^2_2\leq \max_k\left(\frac{2\beta_1^2\left(\frac{1}{|\mathcal{S}_c|}4E^2+8(E-1)^2+2\right)G^2}{4\theta_{min}^4t}+\mathcal{O}\left(\frac{\log M_k}{M_k}+M_k^{\epsilon_k-\frac{1}{2}}+\frac{\log N_k}{N_k}\right)\right).
\end{align*}

\end{proof}

\subsection{Missing Proof in Theorem \ref{theorem:parameter_matern}}

Following the same strategy in Theorem \ref{theorem:exp_hetero}, we can show that
\begin{align*}
    &\norm{\nabla L_k^*(\bbtheta^{(t)})}^2_2 \\
    &= \left(\left[\nabla L_k^*(\bbtheta^{(t)})\right]_1^2+\left[\nabla L_k^*(\bbtheta^{(t)})\right]_2^2\right)\\
    &=\left(\frac{1}{2N_k}(\bar{\theta}_1^{(t)}-\theta_{1}^*)\sum_{j=1}^{N_k}\frac{\lambda_{1j}^2}{\left(\bar{\theta}_{1}^{(t)}\lambda_{1j}+\bar{\theta}_{2}^{(t)}\lambda_{2j}\right)^2} + \frac{1}{2N_k}(\bar{\theta}_2^{(t)}-\theta_{2}^*)\sum_{j=1}^{N_k}\frac{\lambda_{2j}\lambda_{1j}}{\left(\bar{\theta}_{1}^{(t)}\lambda_{1j}+\bar{\theta}_{2}^{(t)}\lambda_{2j}\right)^2}\right)^2\\
    &\qquad + \left(\frac{1}{2N_k}(\bar{\theta}_1^{(t)}-\theta_{1}^*)\sum_{j=1}^{N_k}\frac{\lambda_{1j}\lambda_{2j}}{\left(\bar{\theta}_{1}^{(t)}\lambda_{1j}+\bar{\theta}_{2}^{(t)}\lambda_{2j}\right)^2} + \frac{1}{2N_k}(\bar{\theta}_2^{(t)}-\theta_{2}^*)\sum_{j=1}^{N_k}\frac{\lambda_{2j}^2}{\left(\bar{\theta}_{1}^{(t)}\lambda_{1j}+\bar{\theta}_{2}^{(t)}\lambda_{2j}\right)^2} \right)^2.
\end{align*}
By Lemma \ref{lemma:2-2}, we have
\begin{align*}
    &\norm{\nabla L_k^*(\bbtheta^{(t)})}^2_2 \\
    &\leq\bigg(\frac{1}{2N_k}(\bar{\theta}_1^{(t)}-\theta_{1}^*) N_k^{\frac{(2+\alpha_k)(4b_k+3)}{4b_k(2b_k-1)}}\left(\frac{1}{\theta_{min}^2}+\frac{C^2_{mat,k}(4b_k+3)}{\theta_{min}^2(8b_k^2-8b_k-3)}\right)\\
    &\quad +\frac{1}{2N_k}(\bar{\theta}_2^{(t)}-\theta_{2}^*)N_k^{\frac{(2+\alpha_k)(4b_k+3)}{4b_k(2b_k-1)}}\left(\frac{1}{\theta_{min}^2}+\frac{C_{mat,k}(4b_k+3)}{\theta_{min}^2(4b_k^2-6b_k-3)}\right)\bigg)^2\\
    &\quad +\bigg(\frac{1}{2N_k}(\bar{\theta}_1^{(t)}-\theta_{1}^*)N_k^{\frac{(2+\alpha_k)(4b_k+3)}{4b_k(2b_k-1)}}\left(\frac{1}{\theta_{min}^2}+\frac{C_{mat,k}(4b_k+3)}{\theta_{min}^2(4b_k^2-6b_k-3)}\right)+ \frac{1}{2N_k}(\bar{\theta}_2^{(t)}-\theta_{2}^*)\frac{N_k}{\theta_{min}^2}\bigg)^2\\
    &\leq\bigg(\frac{1}{2}(\theta_{max}-\theta_{min}) N_k^{\frac{(2+\alpha_k)(4b_k+3)}{4b_k(2b_k-1)}-1}\left(\frac{1}{\theta_{min}^2}+\frac{C^2_{mat,k}(4b_k+3)}{\theta_{min}^2(8b_k^2-8b_k-3)}\right)\\
    &\quad +\frac{1}{2}(\theta_{max}-\theta_{min})N_k^{\frac{(2+\alpha_k)(4b_k+3)}{4b_k(2b_k-1)}-1}\left(\frac{1}{\theta_{min}^2}+\frac{C_{mat,k}(4b_k+3)}{\theta_{min}^2(4b_k^2-6b_k-3)}\right)\bigg)^2\\
    &\quad +\bigg(\frac{1}{2}(\theta_{max}-\theta_{min})N_k^{\frac{(2+\alpha_k)(4b_k+3)}{4b_k(2b_k-1)}-1}\left(\frac{1}{\theta_{min}^2}+\frac{C_{mat,k}(4b_k+3)}{\theta_{min}^2(4b_k^2-6b_k-3)}\right)+ \frac{(\bar{\theta}_2^{(t)}-\theta_{2}^*)}{2\theta_{min}^2}\bigg)^2\\
    &\leq a_{mat,1} N_k^{\frac{2(2+\alpha_k)(4b_k+3)}{4b_k(2b_k-1)}-2}+\left(a_{mat,2}N_k^{\frac{(2+\alpha_k)(4b_k+3)}{4b_k(2b_k-1)}-1}+\frac{(\bar{\theta}_2^{(t)}-\theta_{2}^*)}{2\theta_{min}^2}\right)^2\\
    &\leq a_{mat,1} N_k^{\frac{2(2+\alpha_k)(4b_k+3)}{4b_k(2b_k-1)}-2}+a_{mat,2}^2N_k^{\frac{2(2+\alpha_k)(4b_k+3)}{4b_k(2b_k-1)}-2} + 2a_{mat,2}N_k^{\frac{(2+\alpha_k)(4b_k+3)}{4b_k(2b_k-1)}-1}\frac{(\bar{\theta}_2^{(t)}-\theta_{2}^*)}{2\theta_{min}^2}+\frac{(\bar{\theta}_2^{(t)}-\theta_{2}^*)^2}{4\theta_{min}^4}\\
    &\leq a_{mat,1} N_k^{\frac{2(2+\alpha_k)(4b_k+3)}{4b_k(2b_k-1)}-2}+a_{mat,2}^2N_k^{\frac{2(2+\alpha_k)(4b_k+3)}{4b_k(2b_k-1)}-2} + 2a_{mat,2}N_k^{\frac{(2+\alpha_k)(4b_k+3)}{4b_k(2b_k-1)}-1}\frac{(\bar{\theta}_2^{(t)}-\theta_{2}^*)}{2\theta_{min}^2}\\
    &\quad +\frac{1}{4\theta_{min}^4}\bigg(\frac{2\beta_1^2\left(8(E-1)^2+2\right)G^2}{t}  +  2\beta_1 \max_kC_{4k}M_k^{\frac{(2+\alpha_k)(4b_k+3)}{4b_k(2b_k-1)}-1}+2\beta_1\bigg(\sum_{k=1}^K\sqrt{2}p_k(\theta_{max}-\theta_{min})C_{\btheta} M_k^{\epsilon_k-\frac{1}{2}}\bigg)\bigg)
\end{align*}
where $a_{mat,1}=\left(\frac{1}{2}(\theta_{max}-\theta_{min})
\left(\frac{1}{\theta_{min}^2}+\frac{C^2_{mat,k}(4b_k+3)}{\theta_{min}^2(8b_k^2-8b_k-3)}\right)+\frac{1}{2}(\theta_{max}-\theta_{min})
\left(\frac{1}{\theta_{min}^2}+\frac{C_{mat,k}(4b_k+3)}{\theta_{min}^2(4b_k^2-6b_k-3)}\right)\right)^2$ and $a_{mat,2}=\frac{1}{2}(\theta_{max}-\theta_{min})
\left(\frac{1}{\theta_{min}^2}+\frac{C_{mat,k}(4b_k+3)}{\theta_{min}^2(4b_k^2-6b_k-3)}\right)$

By Lemma \ref{lemma:4} and Lemma \ref{lemma:3-2}, with probability at least $1-\max_k\{C_{\btheta}(t+1)(\log( M_k^{\epsilon_k-\frac{1}{2}}))^4\exp\{-c_{\btheta}M_k^{2\epsilon_k}\}\}$
\begin{align*}
    &\norm{\nabla L_k(\bbtheta^{(t)})}^2_2\\
    &\leq \frac{2\beta_1^2\left(8(E-1)^2+2\right)G^2}{4\theta_{min}^4t}+ \mathcal{O}\left(M_k^{\frac{(2+\alpha_k)(4b_k+3)}{4b_k(2b_k-1)}-1}+\sum_{k=1}^Kp_kM_k^{\epsilon_k-\frac{1}{2}}+N_k^{\frac{(2+\alpha_k)(4b_k+3)}{4b_k(2b_k-1)}-1}\right).
\end{align*}
For the partial device participation scenario, the proof is similar.
% where $f_1(b_k)=-\frac{-2b_k^2-5b_k-3}{2b_k(2b_k-1)}$ and $f_2(b_k)=\frac{4b_k(2b_k-1)}{4b_k+3}$. Here we drop some constant terms for the sake of neatness.

% \subsection{Proof of Theorem \ref{theorem:mat_hetero}}

% The proof is same as Theorem \ref{theorem:parameter_matern}.

\section{Proof of Lemmas}
\label{app:lemma_proof}

\subsection{Proof of Lemma \ref{lemma:2-2}}

Remember that the eigenvalues in each device $k$ are different. For the sake of neatness, we omit the subscript $k$ in the eigenvalues. Let $r_k=j^{\frac{4b_k}{4b_k+3}}$ and $\delta_k=\frac{1}{N_k^{\alpha_k+1}}$, where $0<\alpha_k<\frac{8b_k^2-12b_k-6}{4b_k+3}$, then, by Lemma \ref{lemma:1}, with probability at least $1-\delta_k$, we have
\begin{align*}
    &C(r_k,N_k)<r_k\sqrt{\frac{r_k(r_k+1)}{N_k\delta_k\lambda^*_{r_k}}}=j^{\frac{4b_k}{4b_k+3}}\sqrt{\frac{j^{\frac{4b_k}{4b_k+3}}(j^{\frac{4b_k}{4b_k+3}}+1)}{C_kj^{\frac{-8b^2_k}{4b_k+3}}}}N_k^{\frac{\alpha}{2}}\\
    &=N_k^{\frac{\alpha}{2}}j^{\frac{4b_k^2+6b_k}{4b_k+3}}\sqrt{\frac{j^{\frac{4b_k}{4b_k+3}}(j^{-\frac{4b_k}{4b_k+3}}+1)}{C_k}}\leq N_k^{\frac{\alpha}{2}}j^{\frac{4b_k^2+8b_k}{4b_k+3}}\sqrt{\frac{2}{C_k}}
\end{align*}
and
\begin{align*}
     &H(r_k,N_k)<\frac{C_k}{2b_k-1}r^{-(2b_k-1)} + \sqrt{\frac{2C_k}{2b_k-1}}r^{-(b_k-1/2)}N_k^{\alpha/2}\\
     &\leq\left(\frac{C_k}{2b_k-1}+\sqrt{\frac{2C_k}{2b_k-1}}\right)j^{-\frac{2b_k(2b_k-1)}{4b_k+3}}N_k^{\alpha/2}.
\end{align*}
Therefore, by Lemma \ref{lemma:1}, we obtain
\begin{align*}
    &\frac{\lambda_{j}(\bm{K}_{f,N_k})}{N_k}\leq \lambda_{j}^*+\lambda_{j}^*N_k^{\frac{\alpha}{2}}j^{\frac{4b_k^2+8b_k}{4b_k+3}}\sqrt{\frac{2}{C_k}}+\left(\frac{C_k}{2b_k-1}+\sqrt{\frac{2C_k}{2b_k-1}}\right)j^{-\frac{2b_k(2b_k-1)}{4b_k+3}}N_k^{\alpha/2}\\
    &=C_kj^{-2b_k}+C_kj^{-2b_k}N_k^{\frac{\alpha}{2}}j^{\frac{4b_k^2+8b_k}{4b_k+3}}\sqrt{\frac{2}{C_k}}+\left(\frac{C_k}{2b_k-1}+\sqrt{\frac{2C_k}{2b_k-1}}\right)j^{-\frac{2b_k(2b_k-1)}{4b_k+3}}N_k^{\alpha/2}.
\end{align*}
This implies
\begin{align*}
    &\lambda_{j}(\bm{K}_{f,N_k})\leq C_kj^{-2b_k}\left(N_k+N_k^{1+\frac{\alpha}{2}}j^{\frac{4b_k^2+8b_k}{4b_k+3}}\sqrt{\frac{2}{C_k}}\right)+\left(\frac{C_k}{2b_k-1}+\sqrt{\frac{2C_k}{2b_k-1}}\right)j^{-\frac{2b_k(2b_k-1)}{4b_k+3}}N_k^{1+\alpha/2}\\
    &\leq\left(2\sqrt{2C_k}+\frac{C_k}{2b_k-1}+\sqrt{\frac{2C_k}{2b_k-1}}\right)j^{-\frac{2b_k(2b_k-1)}{4b_k+3}}N_k^{1+\alpha/2},
\end{align*}
where probability at least $1-\frac{1}{N_k^{\alpha_k+1}}$. Let $C_{mat,k}=\left(2\sqrt{2C_k}+\frac{C_k}{2b_k-1}+\sqrt{\frac{2C_k}{2b_k-1}}\right)$. Therefore, we have
\begin{align*}
    &\sum_{j=1}^{N_k}\frac{\lambda_{1j}^2}{\left(\theta_{1k}^{(t)}\lambda_{1j}+\theta_{2k}^{(t)}\lambda_{2j}\right)^2}\leq \frac{L_{mat,k}}{\theta_{min}^2} + \frac{C_{mat,k}^2}{\theta_{min}^2}\sum_{j=L_{mat,k}}^\infty j^{-\frac{4b_k(2b_k-1)}{4b_k+3}}N_k^{2+\alpha}
\end{align*}
for any $0<L_{mat,k}\leq N_k$. Let $L_{mat,k}=N_k^{\frac{(2+\alpha_k)(4b_k+3)}{4b_k(2b_k-1)}}$, then we obtain
\begin{align*}
    \sum_{j=1}^{N_k}\frac{\lambda_{1j}^2}{\left(\theta_{1k}^{(t)}\lambda_{1j}+\theta_{2k}^{(t)}\lambda_{2j}\right)^2}\leq N_k^{\frac{(2+\alpha_k)(4b_k+3)}{4b_k(2b_k-1)}}\left(\frac{1}{\theta_{min}^2}+\frac{C^2_{mat,k}(4b_k+3)}{\theta_{min}^2(8b_k^2-8b_k-3)}\right).
\end{align*}
Similarly, we have
\begin{align*}
    \sum_{j=1}^{N_k}\frac{\lambda_{1j}\lambda_{2j}}{\left(\theta_{1k}^{(t)}\lambda_{1j}+\theta_{2k}^{(t)}\lambda_{2j}\right)^2}\leq N_k^{\frac{(2+\alpha_k)(4b_k+3)}{4b_k(2b_k-1)}}\left(\frac{1}{\theta_{min}^2}+\frac{C_{mat,k}(4b_k+3)}{\theta_{min}^2(4b_k^2-6b_k-3)}\right).
\end{align*}
Additionally, we can show that
\begin{align*}
    \sum_{j=1}^{N_k}\frac{\lambda_{2j}^2}{\left(\theta_{1k}^{(t)}\lambda_{1j}+\theta_{2k}^{(t)}\lambda_{2j}\right)^2}\geq\frac{|\{j:\theta^{(t)}_{1k}\lambda_{1j}+\theta^{(t)}_{2k}\lambda_{2j}\leq2\theta_{max}\}|}{4\theta_{max}^2}.
\end{align*}
The fact that $j:\theta^{(t)}_{1k}\lambda_{1j}+\theta^{(t)}_{2k}\lambda_{2j}\leq2\theta_{max}$ implies
\begin{align*}
    &C_{mat,k}\theta_{max}j^{-\frac{2b_k(2b_k-1)}{4b_k+3}}N_k^{1+\alpha/2}\leq \theta_{max}\\
    &\Rightarrow j\geq C_{mat,k}^{\frac{4b_k+3}{2b_k(2b_k-1)}}N_k^{\frac{(2+\alpha_k)(4b_k+3)}{4b_k(2b_k-1)}}\geq C_{mat,k}N_k^{\frac{(2+\alpha_k)(4b_k+3)}{4b_k(2b_k-1)}}
\end{align*}
since $b_k\geq\frac{\sqrt{21}+3}{4}$. Therefore, 
\begin{align*}
    \frac{N_k-C_{mat,k}N_k^{\frac{(2+\alpha_k)(4b_k+3)}{4b_k(2b_k-1)}}}{4\theta_{max}}\leq\sum_{j=1}^{N_k}\frac{\lambda_{2j}^2}{\left(\theta_{1k}^{(t)}\lambda_{1j}+\theta_{2k}^{(t)}\lambda_{2j}\right)^2}\leq\frac{N_k}{\theta_{min}^2}
\end{align*}
where the upper bound is trivially true.

\subsection{Proof of Lemma \ref{lemma:3}}
\begin{proof}
For device $k$, denote by $\btheta_k^{(t)}=(\theta_{1k}^{(t)},\theta_{2k}^{(t)})$ the model parameter at iteration $t$. Let $\lambda^{(t)}_{1jk}$ be the $j^{th}$ largest eigenvalue of $\bm{K}_{f,\xi_k^{(t)}}$ and $\lambda^{(t)}_{2jk}=1$ be the $j^{th}$ largest eigenvalue of $\bm{I}_M$. By definition, 
\begin{align*}
    &\left[g_k^*(\btheta_k^{(t)})\right]_1 \\
    &= \frac{1}{2s_1(M)}\Tr\left[\bm{K}_{\xi_k^{(t)}}(\btheta_k^{(t)})^{-1}\left(\bm{I}_M-\bm{K}_{\xi_k^{(t)}}(\btheta_k^*)\bm{K}_{\xi_k^{(t)}}(\btheta_k^{(t)})^{-1}\right)\frac{\partial \bm{K}_{\xi_k^{(t)}}(\btheta_k^{(t)})}{\partial\theta_{1k}^{(t)}} \right]\\
    &=\frac{1}{2s_1(M)}(\theta_{1k}^{(t)}-\theta_{1k}^*)\sum_{j=1}^M\frac{\lambda_{1jk}^{(t)2}}{(\theta_{1k}^{(t)}\lambda_{1jk}^{(t)}+\theta_{2k}^{(t)}\lambda_{2jk}^{(t)})^2} + \frac{1}{2s_1(M)}(\theta_{2k}^{(t)}-\theta_{2k}^*)\sum_{j=1}^M\frac{\lambda_{2jk}^{(t)}\lambda_{1jk}^{(t)}}{(\theta_{1k}^{(t)}\lambda_{1jk}^{(t)}+\theta_{2k}^{(t)}\lambda_{2jk}^{(t)})^2}
\end{align*}
and
\begin{align*}
    &\left[g_k^*(\btheta_k^{(t)})\right]_2 \\
    &= \frac{1}{2M}\Tr\left[\bm{K}_{\xi_k^{(t)}}(\btheta_k^{(t)})^{-1}\left(\bm{I}_M-\bm{K}_{\xi_k^{(t)}}(\btheta_k^*)\bm{K}_{\xi_k^{(t)}}(\btheta_k^{(t)})^{-1}\right)\frac{\partial \bm{K}_{\xi_k^{(t)}}(\btheta_k^{(t)})}{\partial\theta_{2k}^{(t)}} \right]\\
    &=\frac{1}{2M}(\theta_{1k}^{(t)}-\theta_{1k}^*)\sum_{j=1}^M\frac{\lambda_{1jk}^{(t)}}{(\theta_{1k}^{(t)}\lambda_{1jk}^{(t)}+\theta_{2k}^{(t)}\lambda_{2jk}^{(t)})^2} + \frac{1}{2s_1(M)}(\theta_{2k}^{(t)}-\theta_{2k}^*)\sum_{j=1}^M\frac{\lambda_{2jk}^{(t)}}{(\theta_{1k}^{(t)}\lambda_{1jk}^{(t)}+\theta_{2k}^{(t)}\lambda_{2jk}^{(t)})^2}.
\end{align*}
Based on those two expressions, we can obtain
\begin{align*}
    &\langle \btheta_k^{(t)}-\btheta^*, g_k^*(\btheta_k^{(t)}) \rangle \\
    &=(\btheta_k^{(t)}-\btheta^*)^\intercal \begin{bmatrix} A_{11} & A_{12} \\ A_{21} & A_{22} \end{bmatrix} (\btheta_k^{(t)}-\btheta^*)
\end{align*}
where $A_{11}, A_{12}, A_{21}, A_{22}$ will be clarified shortly. Let $\epsilon_k=\frac{1}{2}$, by Lemma \ref{lemma:2}, with probability at least $1-\frac{2}{M^{\alpha_k}}$, 
\begin{align*}
    A_{11}&\coloneqq\frac{1}{2\tau\log M}\sum_{j=1}^M\frac{\lambda_{1jk}^{(t)2}}{(\theta_{1k}^{(t)}\lambda_{1jk}^{(t)}+\theta_{2k}^{(t)}\lambda_{2jk}^{(t)})^2} \geq \frac{1}{2\tau\log M}\frac{\epsilon_k\log M}{8b_k\theta_{max}^2}=\frac{1}{32\tau b_k\theta_{max}^2},\\
    A_{12}&\coloneqq\frac{1}{2\tau\log M}\sum_{j=1}^M\frac{\lambda_{1jk}^{(t)}}{(\theta_{1k}^{(t)}\lambda_{1jk}^{(t)}+\theta_{2k}^{(t)}\lambda_{2jk}^{(t)})^2} \leq \frac{1}{2\tau\log M}\frac{5+2\alpha_k}{7b_k\theta_{min}^2}\log M\leq\frac{1}{2\tau b_k\theta_{min}^2},\\
    A_{21}&\coloneqq\frac{1}{2M}\sum_{j=1}^M\frac{\lambda_{1jk}^{(t)}}{(\theta_{1k}^{(t)}\lambda_{1jk}^{(t)}+\theta_{2k}^{(t)}\lambda_{2jk}^{(t)})^2}\leq\frac{1}{2M}\frac{5+2\alpha_k}{7b_k\theta_{min}^2}\log M=\frac{5+2\alpha_k}{14b_k\theta_{min}^2}\frac{\log M}{M}\leq\frac{1}{2b_k\theta_{min}^2}\frac{\log M}{M},\\
    A_{12}&\coloneqq\frac{1}{2M}\sum_{j=1}^M\frac{1}{(\theta_{1k}^{(t)}\lambda_{1jk}^{(t)}+\theta_{2k}^{(t)}\lambda_{2jk}^{(t)})^2}\geq\frac{1}{2M} \frac{M-C_{1k}(\alpha_k,b_k)\log M}{4\theta_{max}^2}.
\end{align*}
Therefore,
\begin{align*}
    &\langle \btheta_k^{(t)}-\btheta^*, g_k^*(\btheta_k^{(t)}) \rangle\\
    &\geq \left(\frac{1}{64\tau b_k\theta_{max}^2}-\frac{\log M}{64\theta_{max}^2b_kM}\right)(\theta_{1k}^{(t)}-\theta_1^*)^2\\
    &\qquad +\left(\frac{1}{8\theta_{max}^2}-\frac{4\theta_{max}^2}{\tau b_k\theta_{min}^4}-\frac{C_{1k}(\alpha_k,b_k)\log M}{8\theta_{max}^2 M} + \frac{4\theta_{max}^2\log M}{b\theta_{min}^4 M}\right) (\theta_{2k}^{(t)}-\theta_1^*)^2 \\
    &=\frac{1}{64\tau b_k\theta_{max}^2}(\theta_{1k}^{(t)}-\theta_1^*)^2+\left(\frac{1}{8\theta_{max}^2}-\frac{4\theta_{max}^2}{\tau b_k\theta_{min}^4}\right)(\theta_{2k}^{(t)}-\theta_1^*)^2\\
    &\qquad - \frac{\log M}{64\theta_{max}^2b_kM}\theta_{max}^2 -\frac{C_{1k}(\alpha_k,b_k)\log M}{8\theta_{max}^2 M}\theta_{max}^2 + \frac{4\theta_{max}^2\log M}{b_k\theta_{min}^4 M}\theta_{min}^2\\
    &\geq\frac{\gamma_k}{2}\norm{\btheta_k^{(t)}-\btheta_k^*}_2^2 - C_{3k}(\alpha_k,b_k)\frac{\log M}{M},
\end{align*}
where $\gamma_k=\min\left\{\frac{1}{32\tau b_k\theta_{max}^2},\frac{1}{4\theta_{max}^2}-\frac{8\theta_{max}^2}{\tau b_k\theta_{min}^4}\right\}>0$ and $C_{3k}(\alpha_k,b_k)=\frac{1}{64b_k}+\frac{C_{1k}(\alpha_k,b_k)}{8}- \frac{4\theta_{max}^2}{b\theta_{min}^2}$.
\end{proof}

\subsection{Proof of Lemma \ref{lemma:3-2}}
By definition, we can show that
\begin{align*}
    &\left[g_k^*(\btheta_k^{(t)})\right]_2 \\
    &= \frac{1}{2M}\Tr\left[\bm{K}_{\xi_k^{(t)}}(\btheta_k^{(t)})^{-1}\left(\bm{I}_M-\bm{K}_{\xi_k^{(t)}}(\btheta_k^*)\bm{K}_{\xi_k^{(t)}}(\btheta_k^{(t)})^{-1}\right)\frac{\partial \bm{K}_{\xi_k^{(t)}}(\btheta_k^{(t)})}{\partial\theta_{2k}^{(t)}} \right]\\
    &=\frac{1}{2M}(\theta_{1k}^{(t)}-\theta_{1k}^*)\sum_{j=1}^M\frac{\lambda_{1jk}^{(t)}}{(\theta_{1k}^{(t)}\lambda_{1jk}^{(t)}+\theta_{2k}^{(t)}\lambda_{2jk}^{(t)})^2} + \frac{1}{2M}(\theta_{2k}^{(t)}-\theta_{2k}^*)\sum_{j=1}^M\frac{\lambda_{2jk}^{(t)}}{(\theta_{1k}^{(t)}\lambda_{1jk}^{(t)}+\theta_{2k}^{(t)}\lambda_{2jk}^{(t)})^2}.
\end{align*}
Therefore,
\begin{align*}
    &\left[g_k^*(\btheta_k^{(t)})\right]_2(\theta_{2k}^{(t)}-\theta_{2k}^*)\\
    &=\frac{1}{2M}(\theta_{1k}^{(t)}-\theta_{1k}^*)(\theta_{2k}^{(t)}-\theta_{2k}^*)\sum_{j=1}^M\frac{\lambda_{1jk}^{(t)}}{(\theta_{1k}^{(t)}\lambda_{1jk}^{(t)}+\theta_{2k}^{(t)}\lambda_{2jk}^{(t)})^2} + \frac{1}{2M}(\theta_{2k}^{(t)}-\theta_{2k}^*)^2\sum_{j=1}^M\frac{\lambda_{2jk}^{(t)}}{(\theta_{1k}^{(t)}\lambda_{1jk}^{(t)}+\theta_{2k}^{(t)}\lambda_{2jk}^{(t)})^2}\\
    &\geq \frac{1}{2M}(\theta_{2k}^{(t)}-\theta_{2k}^*)^2\sum_{j=1}^M\frac{1}{(\theta_{1k}^{(t)}\lambda_{1jk}^{(t)}+\theta_{2k}^{(t)}\lambda_{2jk}^{(t)})^2}-\frac{1}{2M}(\theta_{max}-\theta_{min})^2\sum_{j=1}^M\frac{\lambda_{1jk}^{(t)}}{(\theta_{1k}^{(t)}\lambda_{1jk}^{(t)}+\theta_{2k}^{(t)}\lambda_{2jk}^{(t)})^2} \\
    &\geq  \frac{1}{2M_k}(\theta_{2k}^{(t)}-\theta_{2k}^*)^2 \frac{M_k-C_{mat,k}M_k^{\frac{(2+\alpha_k)(4b_k+3)}{4b_k(2b_k-1)}}}{4\theta_{max}}-\frac{1}{2M_k}(\theta_{max}-\theta_{min})^2M_k^{\frac{(2+\alpha_k)(4b_k+3)}{4b_k(2b_k-1)}}\left(\frac{1}{\theta_{min}^2}+\frac{C_{mat,k}(4b_k+3)}{\theta_{min}^2(4b_k^2-6b_k-3)}\right)
\end{align*}
with probability at least $1-\frac{1}{M_k^{1+\alpha_k}}$. Therefore,
\begin{align*}
    \left[g_k^*(\btheta_k^{(t)})\right]_2(\theta_{2k}^{(t)}-\theta_{2k}^*)\geq\frac{\gamma_k}{2}(\theta_{2k}^{(t)}-\theta_{2k}^*)^2-(\theta_{max}-\theta_{min})^2M_k^{\frac{(2+\alpha_k)(4b_k+3)}{4b_k(2b_k-1)}-1}\left(\frac{1}{2\theta_{min}^2}+\frac{C_{mat,k}(4b_k+3)}{2\theta_{min}^2(4b_k^2-6b_k-3)}\right),
\end{align*}
where we slightly abuse the notation and define $\gamma_k\coloneqq\frac{1}{2M_k}\frac{M_k-C_{mat,k}M_k^{\frac{(2+\alpha_k)(4b_k+3)}{4b_k(2b_k-1)}}}{4\theta_{max}}$. Here note that this $\gamma_k$ is different from the $\gamma_k$ in the Lemmas/Theorems involved with RBF kernels.

\bibliography{mybib}
\bibliographystyle{apalike}

\end{document}